\documentclass{article} 
\usepackage[OT1]{fontenc}
\usepackage{iclr2027_conference,times}

\usepackage{amsmath,amsfonts,bm}

\def\eqref#1{equation~\ref{#1}}

\def\1{\bm{1}}

\DeclareMathAlphabet{\mathsfit}{\encodingdefault}{\sfdefault}{m}{sl}
\SetMathAlphabet{\mathsfit}{bold}{\encodingdefault}{\sfdefault}{bx}{n}

\usepackage{graphicx}
\usepackage{capt-of}
\usepackage{float}
\usepackage{tabularx}
\usepackage{booktabs}
\usepackage{multirow}
\usepackage{colortbl}
\usepackage{wrapfig}
\usepackage{needspace}
\usepackage{etoc}
\usepackage{algorithm}
\usepackage{algpseudocode}

\usepackage[hidelinks]{hyperref}
\usepackage{url}

\title{ICL-R: Reconstructing the Computational \\Effects of Demonstrations for Implicit \\Multimodal In-Context Learning}

\author{%
  Shi Qiu\textsuperscript{1}\thanks{Equal contribution.}\quad
  Ziwei Yan\textsuperscript{1}\footnotemark[1]\quad
  Yunkai Dang\textsuperscript{1}\thanks{Project lead.}\quad
  Hongguang Zhang\textsuperscript{2}\quad
  Wenbin Li\textsuperscript{1}\thanks{Corresponding author:
     \nolinkurl{liwenbin@nju.edu.cn} ,
    \nolinkurl{qius19876@gmail.com}.}\\[3pt]
  \textsuperscript{1}School of Intelligence Science and Technology, Nanjing University\\
  \textsuperscript{2}National University of Defense Technology
}

\hypersetup{pdfauthor={Shi Qiu, Ziwei Yan, Yunkai Dang, Hongguang Zhang, Wenbin Li}}

\iclrpreprintcopy
\begin{document}

\maketitle

\begin{abstract}
Multimodal in-context learning (ICL) enables multimodal large language models (MLLMs) to adapt to downstream tasks through interleaved image and text demonstrations.
However, retaining these demonstrations not only increases the sequence length and latency at inference time but also makes predictions sensitive to how demonstrations are selected and presented.
Although existing implicit ICL methods approximate demonstration effects through representation shifts or attention routing, they still suffer from coarse or partial reconstruction of these effects.
To address these limitations, we first decompose the change in attention output caused by demonstrations into the attention routing shift, the value shift, their interaction, and information contributed directly by demonstration tokens.
Based on this decomposition, we propose \emph{\textbf{I}n-\textbf{C}ontext \textbf{L}earning via \textbf{R}econstruction (\textbf{ICL-R})}, a lightweight framework for jointly reconstructing these computational effects.
Aiming at incomplete reconstruction under demonstration removal, we propose \emph{mass-restoring joint attention-value calibration} which jointly calibrates attention routing and value representations while rescaling the attention output.
To further extend supervision beyond isolated state matching, we introduce a \emph{Riemannian state transition objective} that aligns both incoming states and the transitions induced by attention.
Extensive experiments across multiple MLLMs and benchmarks show that ICL-R improves performance while maintaining an inference cost close to that of standard zero-shot inference.
\end{abstract}
\section{Introduction}\label{sec:introduction}
Multimodal Large Language Models (MLLMs) \citep{alayrac2022flamingo,li2023blip2,liu2023visual} provide a general interface for tasks including visual question answering \citep{10203158,goyal2017making}, image captioning \citep{chen2015microsoft}, and visual reasoning \citep{xu2026visulogic}. 
A key approach to adapting these models to downstream tasks is multimodal in-context learning (ICL), which provides an MLLM with interleaved image and text demonstrations of inputs followed by a target query.
The main advantage of this mechanism lies in its ability to support rapid and flexible adaptation without any updates to model parameters \citep{zong2025vlicl}.
However, such explicit ICL is costly with each demonstration substantially adding visual and textual tokens that increase memory use and inference latency \citep{ma2026and}.
The performance of ICL also strongly depends on demonstration selection and format, which determine the task-specific cues and how the outputs are presented \citep{li2024configure,qin2024factors}.
Even with the same demonstrations, changing their order can lead to different predictions \citep{zhang2023good,baldassini2024makes}.
Therefore, demonstrations may provide limited gains or distract the model from the target query \citep{gupta2026mapd}. 
To address these limitations, implicit multimodal ICL captures demonstration effects within the model, allowing it to answer queries without prefixing demonstrations to the input.

Early studies on task and function vectors \citep{hendel2023taskvectors,todd2024functionvectors} show that tasks conveyed through ICL can be encoded as compact vectors within language models. 
These vectors are extracted from internal activations and injected into hidden states to elicit the corresponding task behavior on new queries.
ICV \citep{liu2024icv} develops this perspective into a steering approach by extracting a shared direction from differences between demonstration input and output representations.
Following this idea, LIVE \citep{peng2024live} and MimIC \citep{jiang2025mimic} learn additive representation shifts.
M$^2$IV \citep{li2025m2iv} similarly approximates demonstration effects through separate learnable shifts to the outputs of attention and feedforward modules at each layer.
HiFICL \citep{li2026hificl} introduces virtual key-value pairs to approximate demonstration effects.
However, these designs do not explicitly distinguish interventions on attention scores from changes to the information carried by value vectors, which may limit their ability to adjust the relative attention scores among query tokens.
In contrast, attention routing methods such as ICR \citep{li2026icr} and Hyper-ICL \citep{talemi2026hypericl} focus on adjusting the attention weights assigned to query tokens, but their reconstruction of demonstrations remains incomplete.
Therefore, the remaining challenge is to identify and reconstruct the complementary computational effects induced by demonstrations.

In this paper, we first compare attention outputs with and without demonstrations, decomposing the difference into changes in attention routing and value representations, their interaction, and the direct contribution of demonstration tokens.
This decomposition reveals that demonstrations reshape information selection and content propagation in a coupled manner while contributing additional information directly.
Motivated by both the limitations of existing methods and this empirical decomposition, we propose \emph{\textbf{I}n-\textbf{C}ontext \textbf{L}earning via \textbf{R}econstruction (\textbf{ICL-R})} , a query-only framework that reconstructs the computational effects of demonstrations from a frozen teacher to a student with lightweight modules. 
Specifically, we propose joint attention and value calibration to reconstruct the coupled changes in information selection and content propagation, with query-adaptive corrections accounting for variation across inputs.
The accompanying missing mass estimation compensates for the attention normalization mismatch caused by removing demonstrations, rescaling the query contribution.
Our Riemannian state transition objective extends supervision beyond isolated state matching by aligning incoming states on the unit hypersphere and the direction and magnitude of attention-induced transitions in the corresponding tangent space.
We evaluate ICL-R across different model architectures on six datasets covering visual question answering, image captioning, and multimodal perception and reasoning.
Additionally, we analyze the complementary roles of attention and value calibration and evaluate the contributions of the individual components in ICL-R.
We further investigate the data efficiency, inference efficiency, robustness to varying demonstration counts, sensitivity to rank configurations, and object hallucination.

In summary, our main contributions in this work are as follows:
\begin{itemize}
    \setlength{\topsep}{2pt}
    \setlength{\partopsep}{0pt}
    \setlength{\itemsep}{1pt}
    \setlength{\parskip}{0pt}
    \setlength{\parsep}{0pt}
    \item We formulate implicit ICL as the problem of the reconstruction of demonstration effects under demonstration removal and establish an exact mathematical decomposition of these effects in attention output. 
    This view reveals that demonstration effects arise from more than a single representation shift, providing the basis for our explicit reconstruction.
    \item We propose ICL-R, a lightweight framework that reconstructs information selection and content propagation through joint attention and value calibration with missing mass estimation correcting the output scale change caused by removing demonstrations.
    \item We introduce a Riemannian objective that aligns the states entering attention on the unit hypersphere and the direction and magnitude of their attention-induced transitions in the corresponding tangent spaces to more accurately capture the effects of demonstrations.
\end{itemize}

\section{Related Work}\label{sec:related-work}
\textbf{Multimodal In-Context Learning.} Multimodal Large Language Models provide general capabilities for vision and language understanding \citep{alayrac2022flamingo,li2023blip2}. 
Building on these capabilities, multimodal in-context learning adapts models to new tasks by inputting interleaved image and text demonstrations and a target query without parameter updates \citep{tsimpoukelli2021multimodal}. 
However, visual demonstrations substantially increase input length and inference cost \citep{jiang2025mimic,li2025m2iv,chen2024fastv}, while performance also depends on demonstration selection and configuration \citep{li2024configure,sun2025exploring}.
Selecting useful examples is particularly challenging because similarity in one modality may overlook information required in the other, allowing similar questions to induce answer copying that conflicts with the query image \citep{li2024configure}.
To improve demonstration quality, prior studies investigate prompt retrieval in visual ICL \citep{zhang2023good}, sequence configuration for visual question answering \citep{li2024configure}, and multimodal context retrieval and refinement \citep{fu2026contextnav}.
Although these strategies improve the use of demonstrations, they continue to process explicit examples during inference and therefore retain the associated computational burden and memory overhead.

\textbf{Implicit In-Context Learning.} 
To address the inference overhead, implicit ICL seeks to preserve the benefits of demonstrations through representation engineering \citep{snell2022learning,li2025implicit}.
Early studies like task and function vector \citep{hendel2023taskvectors,todd2024functionvectors,liu2024icv} extract task representations from the activations and inject them at empirically selected layers to guide responses to new queries.
Building on this idea of vector-based intervention, LIVE \citep{peng2024live} learns an in-context vector for each layer, while MimIC \citep{jiang2025mimic} introduces separate shifts after each attention head and adjusts their strength according to the query. 
M$^2$IV \citep{li2025m2iv} further applies separate learnable shifts to the outputs of attention and feedforward modules at each layer.
Moving beyond shift vectors, HiFICL \citep{li2026hificl} learns virtual key and value pairs, allowing each query to dynamically combine the original attention output with learned context values. 
Although effective, these designs conflate changes in attention distribution with changes in the information carried by value vectors \citep{talemi2026hypericl}.
Attention routing methods instead directly adjust the attention weights assigned to tokens, providing direct calibration over how information is selected during aggregation.
ICR \citep{li2026icr} learns to combine reusable attention directions, while Hyper-ICL \citep{talemi2026hypericl} applies low-rank attention-logit corrections with query-dependent strength. 
However, they capture only part of the computation induced by demonstrations.
To deal with these limitations, our ICL-R explicitly separates and reconstructs the effects of demonstrations in attention routing and value content with missing mass correction accounting for the change in attention normalization after demonstration removal.

\textbf{Geometric Intermediate Representation Distillation.} Intermediate representation distillation transfers knowledge by supervising internal features beyond task predictions. 
Early work such as FitNets \citep{romero2014fitnets} introduced feature regression to guide student learning with teacher representations. 
Subsequent work extended supervision to relationships between layers, aiming to transfer the teacher's solution process \citep{yim2017gift}. 
Within implicit multimodal ICL, MimIC \citep{jiang2025mimic} combines task supervision with squared $\ell_2$ alignment between teacher and student hidden states. 
HiFICL \citep{li2026hificl} relies on cross-entropy supervision alone, while Hyper-ICL \citep{talemi2026hypericl} uses Lorentz geodesic distances for intermediate representation alignment.
However, task supervision constrains output predictions, while the intermediate alignment objectives above compare absolute teacher and student hidden states at individual layers.
To account for both the state transitions before and after the attention, we draw on Riemannian geometry, which provides geodesic distances for comparing states and logarithmic maps for describing local displacements in tangent spaces \citep{Pennec2006,Lin2021}. 
We represent normalized hidden states on the unit hypersphere and jointly supervise pre-attention states and attention-induced transitions in their corresponding tangent spaces. 
This objective extends pointwise hidden-state matching by jointly aligning pre-attention states and transitions represented as tangent vectors.

\begingroup
\raggedbottom
\fontsize{10pt}{11.5pt}\selectfont
\setlength{\parskip}{3.5pt}
\setlength{\abovedisplayskip}{5.5pt}
\setlength{\belowdisplayskip}{5.5pt}
\setlength{\abovedisplayshortskip}{3pt}
\setlength{\belowdisplayshortskip}{4pt}
\setlength{\jot}{2.5pt}
\setlength{\textfloatsep}{10pt}
\setlength{\intextsep}{8pt}
\setlength{\abovecaptionskip}{5pt}
\setlength{\belowcaptionskip}{0pt}
\makeatletter
\def
\subsection{\@startsection{subsection}{2}{\z@}
{-1.6ex}{0.7ex}{\normalsize\sc\raggedright}}
\def\paragraph{\@startsection{paragraph}{4}{\z@}
{1.2ex}{-1em}{\normalsize\bf}}
\makeatother

\section{Method}\label{sec:method}
In this section, we first decompose demonstration effects to guide their modeling in Section~\ref{sec:problem-formulation}.
We then present mass-restoring joint attention--value calibration, integrating routing and value corrections with missing mass estimation, in Section~\ref{sec:joint-calibration}.
Finally, we introduce the Riemannian state transition loss for training the query-only student in Section~\ref{sec:riemannian-loss}.

\subsection{Motivation and Problem Formulation}\label{sec:problem-formulation}
Following prior work on transferring demonstration effects to query-only models \citep{jiang2025mimic,talemi2026hypericl}, we consider a training episode containing a set of demonstrations $\mathcal{D}$ and a target query $X$. 
The teacher receives the concatenated context $[\mathcal{D};X]$, while the student receives only $X$. 
For one self-attention head with projected dimension $d_h$, let $H_{T,X}$ and $H_{T,D}$ denote the teacher hidden states for query and demonstration tokens.
The teacher projections are:
\begin{equation}
  \label{eq:teacher-projections}
  Q_{T,X}=H_{T,X}W_Q,\qquad
  K_T=[H_{T,D};H_{T,X}]W_K,\qquad
  V_T=[H_{T,D};H_{T,X}]W_V.
\end{equation}
We partition the resulting attention matrix and output into demonstration and query components to distinguish information retrieved from demonstrations from that aggregated over query tokens:
\begin{equation}
  \label{eq:teacher-attention-output}
    A_T
    =\operatorname{softmax}
    \left(\frac{Q_{T,X}K_T^{\top}}{\sqrt{d_h}}\right)
    =[A_{T,D},A_{T,X}],
    \qquad
    O_T=A_{T,D}V_{T,D}+A_{T,X}V_{T,X},
\end{equation}
where $A_{T,D}\in\mathbb{R}^{n_X\times n_D}$ captures attention from query tokens to demonstrations and $A_{T,X}\in\mathbb{R}^{n_X\times n_X}$ captures attention among query tokens, with $n_X$ and $n_D$ denoting the numbers of query and demonstration tokens.
For the frozen query-only model, let $Q_0=H_0W_Q$, $K_0=H_0W_K$, and $V_0=H_0W_V$ denote the query, key, and value projections of the query-only hidden states $H_0$ respectively. 
The attention matrix and corresponding output are computed as follows:
\begin{equation}
  \label{eq:base-attention-output}
  \begin{aligned}
    A_0&=\operatorname{softmax}
    \left(\frac{Q_0K_0^{\top}}{\sqrt{d_h}}\right),
    &O_0&=A_0V_0.
  \end{aligned}
\end{equation}

To expose the effect of demonstrations, define the query-side changes $\Delta A_T=A_{T,X}-A_0$ and $\Delta V_T=V_{T,X}-V_0$. 
Substituting them into Eq.~(\ref{eq:teacher-attention-output}) and subtracting the base output in Eq.~(\ref{eq:base-attention-output}) yields:
\begin{equation}
  \label{eq:demonstration-effect-decomposition}
  \begin{aligned}
    O_T-O_0
    =\;&\underbrace{\Delta A_TV_0}_{\text{routing shift}}
    +\underbrace{A_0\Delta V_T}_{\text{value shift}}
    +\underbrace{\Delta A_T\Delta V_T}_{\text{interaction}}
    +\underbrace{A_{T,D}V_{T,D}}_{\text{demonstration information}}.
  \end{aligned}
\end{equation}
The four terms in Eq.~(\ref{eq:demonstration-effect-decomposition}) respectively capture changes in routing, value content, their interaction, and information retrieved directly from demonstrations.
On this basis, our goal is to learn a student with lightweight modules whose output $\widetilde{O}$ approximates $O_T$, as summarized in Figure~\ref{fig:icl-r-overview}.
\begin{figure}[t]
  \centering
  \setlength{\abovecaptionskip}{-8pt}
  \includegraphics[width=1\linewidth]{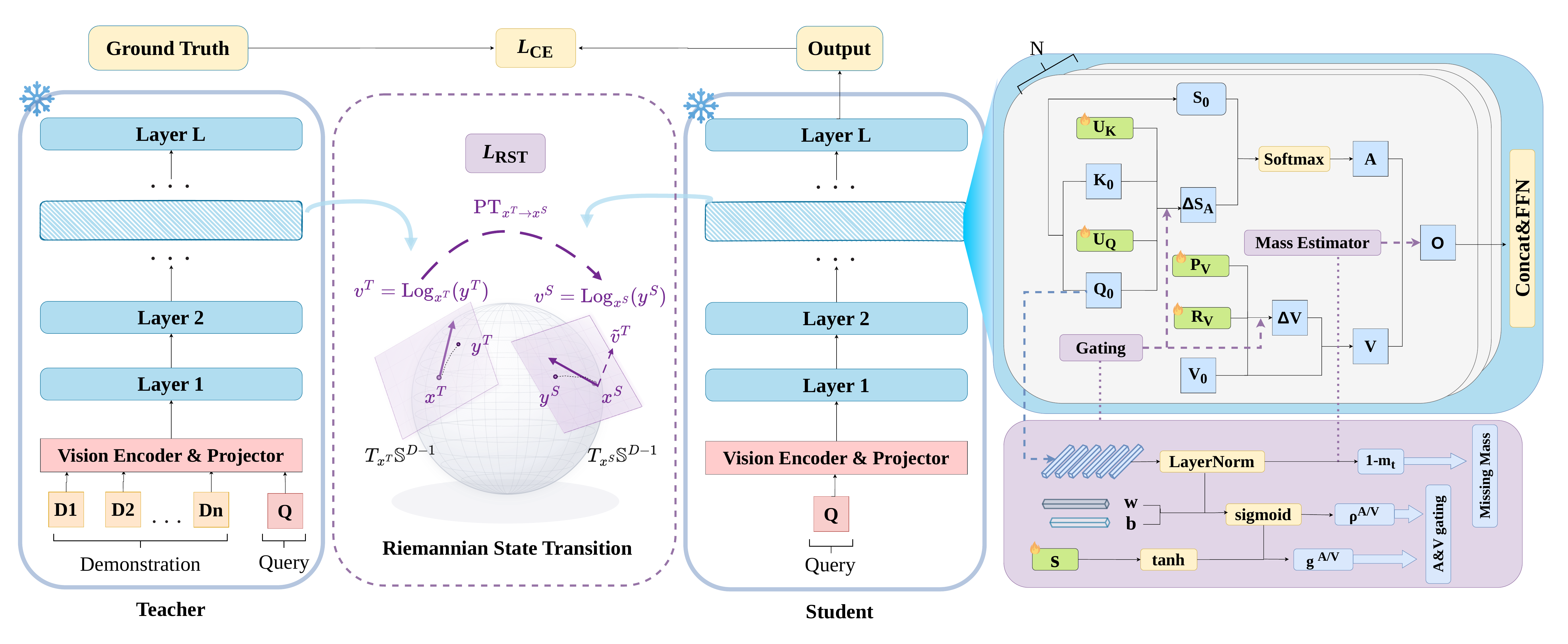}
  \vspace{-10pt}
  \caption{\textbf{Overview of ICL-R.} Our method transfers the effects of demonstrations from a frozen teacher to a query-only student through  mass-restoring joint attention-value calibration and Riemannian state transition alignment. The student receives only the query without demonstrations.}
  \label{fig:icl-r-overview}
\end{figure}

\subsection{Mass-Restoring Joint Attention-Value Calibration}\label{sec:joint-calibration}

The decomposition in Eq.~(\ref{eq:demonstration-effect-decomposition}) motivates reconstructing demonstration effects by jointly calibrating attention and value content.
This requires calibrating not only where information is selected and what content is propagated, but also how much attention mass is assigned to query content.
Specifically, for a query token indexed by $t$, the teacher distributes its attention over both demonstrations and query tokens, with $m_t^T$ denoting the attention mass assigned to demonstrations. Consequently, the total mass assigned to query tokens is reduced as expressed below:
\nopagebreak[4]
\begin{equation}
  \label{eq:teacher-missing-mass}
  m_t^T=\sum_{s\in\mathcal{D}}(A_{T,D})_{t,s},
  \qquad
  \sum_{s\in X}(A_{T,X})_{t,s}=1-m_t^T.
\end{equation}
Removing demonstrations forces a query-only softmax to assign unit mass to query tokens, changing the scale of their contribution.
We incorporate mass correction into the attention weights that aggregate calibrated values.
Let $q_t$ be the student query projection, $\bar q_t=\operatorname{LN}(q_t)$ its normalized form, $S_0=Q_0K_0^{\top}/\sqrt{d_h}$ the base logits before calibration at the current layer and $\Delta S^A$ adjusts these logits to redistribute attention among query tokens.
The resulting calibrated output is:
\begin{equation}
  \label{eq:calibrated-attention-output}
  \begin{aligned}
    \widetilde O_t
    &=\sum_s\widehat A_{t,s}\widetilde V_{t,s},
    &\qquad \widehat A_{t,s}
    &=\left[1-s_m\,\sigma\!\left(
    \frac{\bar q_t^{\top}w_m}{\sqrt{d_h}}+b_m\right)\right]
    \bigl[\operatorname{softmax}(S_0+\Delta S^A)\bigr]_{t,s}.
  \end{aligned}
\end{equation}

Here, $\widehat A$ combines calibrated routing with missing mass estimation, and $\widetilde V_{t,s}=V_{0,s}+\Delta V_{t,s}$ denotes the calibrated value content.
The trainable $w_m\in\mathbb{R}^{d_h}$ and $b_m\in\mathbb{R}$ parameterize a learned proxy for the demonstration attention mass, while $s_m$ sets its maximum value.
Within this unified output, we parameterize the routing and value corrections jointly through query-adaptive low-rank updates.
The scalar gates $g_t^A$ and $g_t^V$ control the overall strengths of the two corrections for query token $t$, while $\rho_t^A$ and $\rho_t^V$ modulate the contributions of their individual low-rank directions:
\begin{equation}
  \label{eq:joint-attention-value-calibration}
  \Delta S^A_{t,s}
  =g_t^A
  \left[\rho_t^A\odot\left(Q_0U_Q\right)_{t,:}\right]
  \left(K_0U_K\right)_{s,:}^{\top},
  \qquad
  \Delta V_{t,s}
  =g_t^V\left[\left(\rho_t^V\odot V_{0,s}P_V\right)R_V\right].
\end{equation}

Here, $U_Q,U_K\in\mathbb{R}^{d_h\times r_A}$ map the uncalibrated queries and keys into a low-rank attention subspace, while $P_V\in\mathbb{R}^{d_h\times r_V}$ and $R_V\in\mathbb{R}^{r_V\times d_h}$ project values into and out of a low-rank value subspace.
The score correction changes the relative allocation within $\widehat A$, while the value correction changes the content aggregated under these mass-aware weights.
The ranks satisfy $r_A,r_V\ll d_h$, keeping the updates lightweight. 
Both low-rank updates use the same query-adaptive design.
For component $c\in\{A,V\}$, let $u_t^c=\bar q_t^{\top}w_c+b_c$ be the scalar score derived from the current query token. 
We use this score to define gates for correction strength and the contribution of each low-rank direction:
\begin{equation}
  \label{eq:token-conditioned-gates}
  \begin{aligned}
    g_t^c
    &=2\tanh(s_c)\,\sigma(u_t^c),
    &\qquad
    \rho_{t,j}^c
    &=2\sigma\!\left(u_t^c\gamma_j^c+\eta_j^c\right),
    \qquad j=1,\ldots,r_c.
  \end{aligned}
\end{equation}
where $w_c\in\mathbb{R}^{d_h}$, $b_c,s_c\in\mathbb{R}$, and $\gamma^c,\eta^c\in\mathbb{R}^{r_c}$ are trainable parameters.
The scalar gate $g_t^c\in(-2,2)$ sets the signed correction strength, while $\rho_t^c\in(0,2)^{r_c}$ selects the active low-rank directions.

\subsection{Riemannian State Transition Loss}\label{sec:riemannian-loss}
Existing implicit ICL methods use cross-entropy supervision either alone \citep{li2026hificl} or together with intermediate representation alignment based on Euclidean or hyperbolic distances \citep{jiang2025mimic,talemi2026hypericl}.
Cross-entropy loss supervises only the final prediction, while directly matching hidden states with a Euclidean distance constrains only their endpoints. 
Neither objective directly constrains the transition from the pre-attention state to the post-attention state.
We therefore model self-attention as a state transition in the tangent space on the hypersphere and distill both its starting point and its demonstration-induced motion. 
For each aligned query token indexed by $i=(b,\ell,t)$, let $h_i^a\in\mathbb{R}^{D}$ be the residual state before self-attention and $o_i^a\in\mathbb{R}^{D}$ its attention update where $a\in\{T,S\}$ denotes the teacher or student and $D$ is the hidden dimension of the language model.
We map the states before and after attention onto the unit hypersphere:

\begin{equation}
  \label{eq:spherical-state-projection}
  x_i^a=\pi(h_i^a),\qquad
  y_i^a=\pi(h_i^a+o_i^a),\qquad
  \pi(z)=\frac{z}{\lVert z\rVert_2}.
\end{equation}
Here, $x_i^a$ represents the normalized state entering self-attention, while $y_i^a$ represents the normalized state after applying the attention update $o_i^a$. 
Both lie on $\mathbb{S}^{D-1}$, and the transition $x_i^a\rightarrow y_i^a$ captures the effect of self-attention on that token. 
We encode this transition as a tangent vector through the spherical logarithmic map. 
For two points $x,y\in\mathbb{S}^{D-1}$ with $y\neq\pm x$, we define:
\begin{equation}
  \label{eq:spherical-logarithmic-map}
  \operatorname{Log}_x(y)
  =\frac{\theta}{\lVert u\rVert_2}u.
\end{equation}
Here, $u=y-\langle x,y\rangle x$ and $\theta=\operatorname{atan2}(\lVert u\rVert_2,\langle x,y\rangle)$.
Using the logarithmic map in Eq.~(\ref{eq:spherical-logarithmic-map}), we represent each attention-induced transition as a tangent vector $v_i^a=\operatorname{Log}_{x_i^a}(y_i^a)\in T_{x_i^a}\mathbb{S}^{D-1}$.
This tangent vector represents the state change induced by self-attention. 
Its direction specifies where the state moves, while its norm $\lVert v_i^a\rVert_2=\theta$ quantifies the magnitude of the change. 
Since $v_i^T$ and $v_i^S$ lie in different tangent spaces, they cannot be compared directly. 
We therefore parallel-transport the teacher transition from the tangent space at $x_i^T$ to that at $x_i^S$ along the shortest spherical geodesic:
\begin{equation}
  \label{eq:spherical-parallel-transport}
  \widetilde v_i^T
  =v_i^T-
  \frac{\langle v_i^T,x_i^S\rangle}
  {1+\langle x_i^T,x_i^S\rangle}
  (x_i^T+x_i^S).
\end{equation}
We then align both the pre-attention state and the attention-induced transition. 
Let $\mathcal I$ be the set of aligned batch, layer, and query-token indices. 
The point and fiber losses are:
\begin{equation}
  \label{eq:rst-point-fiber-losses}
  \begin{aligned}
    \mathcal L_{\mathrm{point}}
    &=\frac{4}{|\mathcal I|}\sum_{i\in\mathcal I}
    \arcsin^2\!\left(\frac{\lVert x_i^S-x_i^T\rVert_2}{2}\right),
    &\qquad
    \mathcal L_{\mathrm{fiber}}
    &=\frac{1}{|\mathcal I|}\sum_{i\in\mathcal I}
    \lVert v_i^S-\widetilde v_i^T\rVert_2^2.
  \end{aligned}
\end{equation}
The point term $\mathcal L_{\mathrm{point}}$ in Eq.~(\ref{eq:rst-point-fiber-losses}) uses the squared spherical geodesic distance to align the states entering self-attention. 
The fiber term $\mathcal L_{\mathrm{fiber}}$ compares the student transition with the transported teacher transition from Eq.~(\ref{eq:spherical-parallel-transport}). 
The complete training objective is:
\begin{equation}
  \label{eq:complete-training-objective}
  \mathcal L_{\mathrm{RST}}
  =\mathcal L_{\mathrm{point}}+\mathcal L_{\mathrm{fiber}},
  \qquad
  \mathcal L_{\mathrm{total}}
  =\mathcal L_{\mathrm{CE}}+\lambda\mathcal L_{\mathrm{RST}}.
\end{equation}
Here, $\lambda \geq 0$ is a hyperparameter that balances task supervision and geometric alignment.
The complete algorithmic workflow is summarized in Appendix~\ref{app:algorithmic-procedure}.

\endgroup

\begingroup
\raggedbottom
\section{Experiments}\label{sec:experiments}

\noindent\textbf{Implementation details.}\label{sec:implementation}
To assess the effectiveness and generalization of our method, we evaluate it on two fully autoregressive models, Idefics2-8B-Base~\citep{laurencon2024matters} and LLaVA-Interleave-7B~\citep{li2024llavanextinterleave} across six multimodal benchmarks.
Specifically, VQAv2 \citep{goyal2017making} and OK-VQA \citep{marino2019okvqa} evaluate general and knowledge-based visual question answering. 
COCO Caption \citep{chen2015microsoft} and Flickr30K \citep{young2014image} evaluate image captioning, while MME \citep{fu2025mme} and SEED-Bench \citep{li2024seedbench} assess broader multimodal perception and reasoning.
We report answer accuracy for VQAv2, OK-VQA, MME, and SEED, and CIDEr \citep{vedantam2015cider} for COCO Caption and Flickr30K.
During training, the frozen teacher receives demonstrations together with the query, while the student receives only the query.
We keep the vision encoder, projector, and language model frozen, updating only our modules.
For each benchmark, we adopt eight sampled demonstrations per episode, set both calibration ranks to 8 and optimize our modules with AdamW \citep{loshchilov2019decoupled}.
Evaluation subsets, sampling procedures, and additional implementation settings are detailed in Appendix~\ref{app:implementation-details}.

\noindent\textbf{Baselines and main results.}\label{sec:baselines}\label{sec:Main-experiment}
We compare ICL-R with zero-shot inference, 8-shot ICL and representative trainable baselines.
Specifically, LoRA \citep{hu2022lora} adapts the attention projection matrices through low-rank updates. LIVE \citep{peng2024live} and MimIC \citep{jiang2025mimic} approximate demonstration effects through additive shifts while HiFICL~\citep{li2026hificl} replaces demonstrations with virtual key-value pairs.
Table~\ref{tab:main-results} shows that 8-shot ICL consistently improves over zero-shot inference with Idefics2. With LLaVA-Interleave, it improves performance on VQAv2, OK-VQA, and COCO, but remains below zero-shot inference on Flickr30K, MME, and SEED, highlighting that the benefits of explicit demonstrations vary across tasks and backbones.
LoRA and LIVE fall below vanilla ICL on OK-VQA with Idefics2, while LoRA underperforms zero-shot inference on COCO with LLaVA-Interleave.
In contrast, our method achieves the best results on every benchmark with Idefics2, leading both the average accuracy and average CIDEr scores.
On LLaVA-Interleave, our ICL-R leads on VQAv2, OK-VQA, COCO and SEED and achieves the highest average accuracy and CIDEr, while ranking second on MME and Flickr30K.
Overall, ICL-R delivers strong and consistent performance across diverse multimodal tasks while avoiding the inference-time overhead of retaining explicit demonstrations.
Additional results on other model backbones and task-specific analyses are provided in Appendices~\ref{app:other-mllms} and~\ref{app:error-correction}.

\begin{table}[h]
\setlength{\abovecaptionskip}{-5pt}
  \centering
  \addtolength{\belowcaptionskip}{6pt}
  \caption{Results on six benchmarks with Average denoting the mean of accuracy or of CIDEr. Bold and underlined numbers denote the best and second-best results.}
  \label{tab:main-results}
  \scriptsize
  \setlength{\tabcolsep}{1.2pt}
  \renewcommand{\arraystretch}{1.08}
  \begin{tabular*}{\linewidth}{@{\extracolsep{\fill}}l@{\hspace{2pt}}l@{\hspace{3pt}}c*{8}{>{\centering\arraybackslash}p{0.065\linewidth}}@{}}
    \toprule
    \multirow{2}{*}{Model} & \multirow{2}{*}{Method} & \multirow{2}{*}{Trainable}
    & \multicolumn{5}{c}{Accuracy $\uparrow$} & \multicolumn{3}{c}{CIDEr $\uparrow$} \\
    \cmidrule(lr){4-8}\cmidrule(lr){9-11}
    & & & VQAv2 & OKVQA & MME & SEED & Average & COCO & Flickr30K & Average \\
    \midrule
    \multirow{7}{*}{\rotatebox[origin=c]{90}{\textbf{Idefics2}}}
    & Zero-shot & $\times$ & 55.39 & 43.08 & 74.82 & 44.77 & 54.52 & 0.6828 & 0.5327 & 0.6078 \\
    & 8-shot ICL & $\times$ & 68.25 & 57.68 & 78.17 & 48.34 & 63.11 & 1.1923 & \underline{0.8457} & 1.0190 \\
    \cmidrule(lr){2-11}
    & LoRA~\citep{hu2022lora} & $\checkmark$ & 69.14 & 55.05 & 79.26 & \underline{63.14} & 66.65 & 1.2665 & 0.8054 & 1.0360 \\
    & LIVE~\citep{peng2024live} & $\checkmark$ & 67.60 & 54.86 & \underline{85.52} & 61.71 & 67.42 & 1.2764 & 0.8280 & \underline{1.0522} \\
    & MimIC~\citep{jiang2025mimic} & $\checkmark$ & 69.29 & 58.74 & 82.61 & 60.80 & \underline{67.86} & 1.2827 & 0.7879 & 1.0353 \\
    & HiFICL~\citep{li2026hificl} & $\checkmark$ & \underline{72.08} & \underline{59.56} & 77.07 & 46.12 & 63.71 & \underline{1.2951} & 0.7831 & 1.0391 \\
    & \textbf{ICL-R (Ours)} & $\checkmark$ & \textbf{73.38} & \textbf{62.49} & \textbf{86.83} & \textbf{63.95} & \textbf{71.66} & \textbf{1.3757} & \textbf{0.8982} & \textbf{1.1370} \\
    \midrule
    \multirow{7}{*}{\rotatebox[origin=c]{90}{\textbf{LLaVA-Interleave}}}
    & Zero-shot & $\times$ & 13.02 & 5.10 & 81.88 & 70.07 & 42.52 & 1.1516 & 0.7879 & 0.9698 \\
    & 8-shot ICL & $\times$ & 68.19 & 43.84 & 79.48 & 62.67 & 63.55 & 1.2085 & 0.7411 & 0.9748 \\
    \cmidrule(lr){2-11}
    & LoRA~\citep{hu2022lora} & $\checkmark$ & 70.12 & 48.19 & \textbf{85.30} & 72.29 & 68.98 & 1.0665 & 0.7529 & 0.9097 \\
    & LIVE~\citep{peng2024live} & $\checkmark$ & 74.17 & 51.77 & 84.35 & \underline{72.55} & 70.71 & 1.2743 & 0.8815 & 1.0779 \\
    & MimIC~\citep{jiang2025mimic} & $\checkmark$ & 74.40 & 52.29 & 84.43 & 71.26 & 70.60 & 1.3169 & \textbf{0.9137} & \underline{1.1153} \\
    & HiFICL~\citep{li2026hificl} & $\checkmark$ & \underline{74.66} & \underline{54.19} & 84.64 & 71.03 & \underline{71.13} & \underline{1.3315} & 0.7064 & 1.0190 \\
    & \textbf{ICL-R (Ours)} & $\checkmark$ & \textbf{76.59} & \textbf{57.38} & \underline{84.72} & \textbf{72.75} & \textbf{72.86} & \textbf{1.3386} & \underline{0.8926} & \textbf{1.1156} \\
    \bottomrule
  \end{tabular*}
\end{table}

\noindent\textbf{Attention value effect analysis.}\label{sec:av-effect}
To determine whether attention and value calibration recover complementary components of the effects of demonstrations, we conduct a head-wise recovery analysis, with the results shown in Figure~\ref{fig:av-heatmap}. 
All three metrics calculate cosine similarity between changes relative to the frozen model without demonstrations.
$M_A$ compares the attention redistribution produced by attention calibration with the demonstration-induced redistribution in the teacher.
$M_V$ compares the attention-output changes attributable to value calibration with the teacher's query-side value shifts aggregated under the frozen model's attention weights.
$M_{AV}$ compares the total attention-output changes produced by joint calibration with the teacher's full demonstration-induced changes.
The left panel of Figure~\ref{fig:av-heatmap} shows highly head-specific recovery, with its directional alignment varying substantially across heads. 
The middle panel shows weaker recovery at individual heads but a broader distribution across layers. 
The right panel shows more consistent positive alignment across layers and heads, indicating that the two mechanisms capture complementary aspects of the teacher's computation. 
Together, these results support jointly reconstructing information selection and content propagation to better recover demonstration effects within attention.
Appendix~\ref{app:layer-wise-recovery} provides an additional layer-wise and question-level analysis.

\begingroup
\setlength{\intextsep}{3pt}
\setlength{\abovecaptionskip}{3pt}
\setlength{\belowcaptionskip}{0pt}
\begin{figure}[H]
  \centering
  \includegraphics[width=0.9\linewidth]{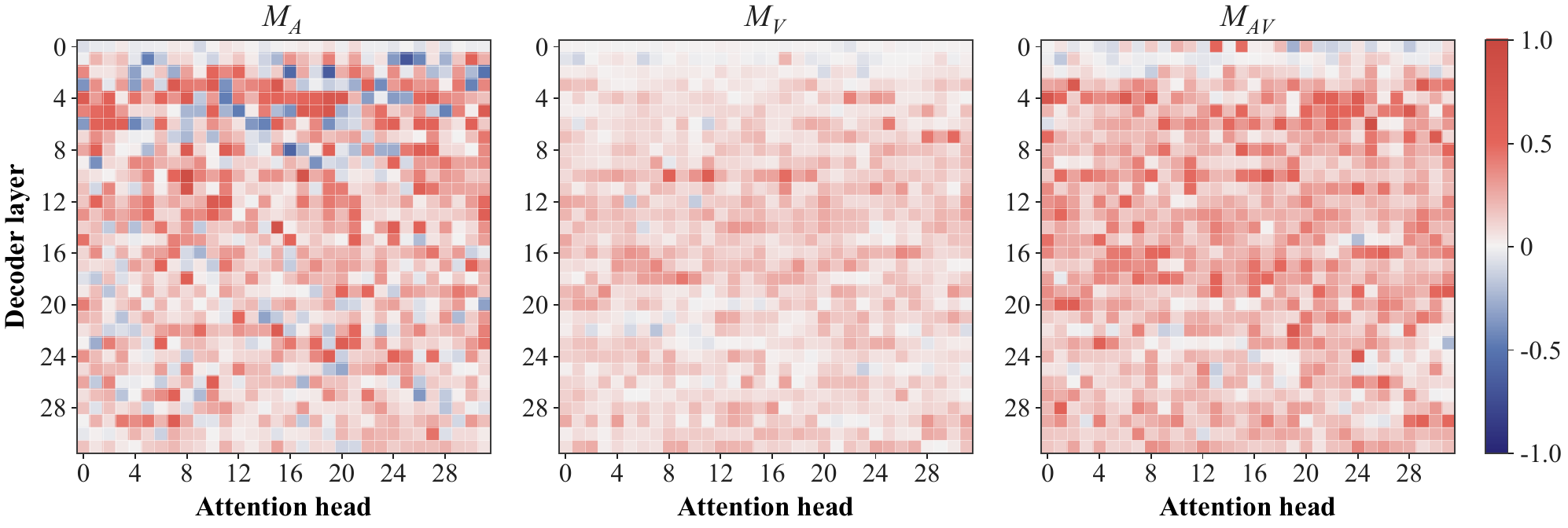}
  \vspace{-4pt}
  \caption{Head-wise cosine similarities between the student and teacher for attention routing, value content, and joint output recovery. Joint calibration produces more consistent positive alignment, demonstrating the complementary roles of attention and value calibration.}
  \label{fig:av-heatmap}
\end{figure}
\endgroup
\vspace{-3pt}

\begingroup
\setlength{\abovecaptionskip}{3pt}
\setlength{\belowcaptionskip}{3pt}
\setlength{\intextsep}{3pt}
\setlength{\columnsep}{0.03\linewidth}
\begin{wrapfigure}{r}{0.42\linewidth}
  \vspace{0pt}
  \centering
  \setlength{\abovecaptionskip}{0pt}
  \setlength{\belowcaptionskip}{0pt}
  \includegraphics[width=\linewidth]{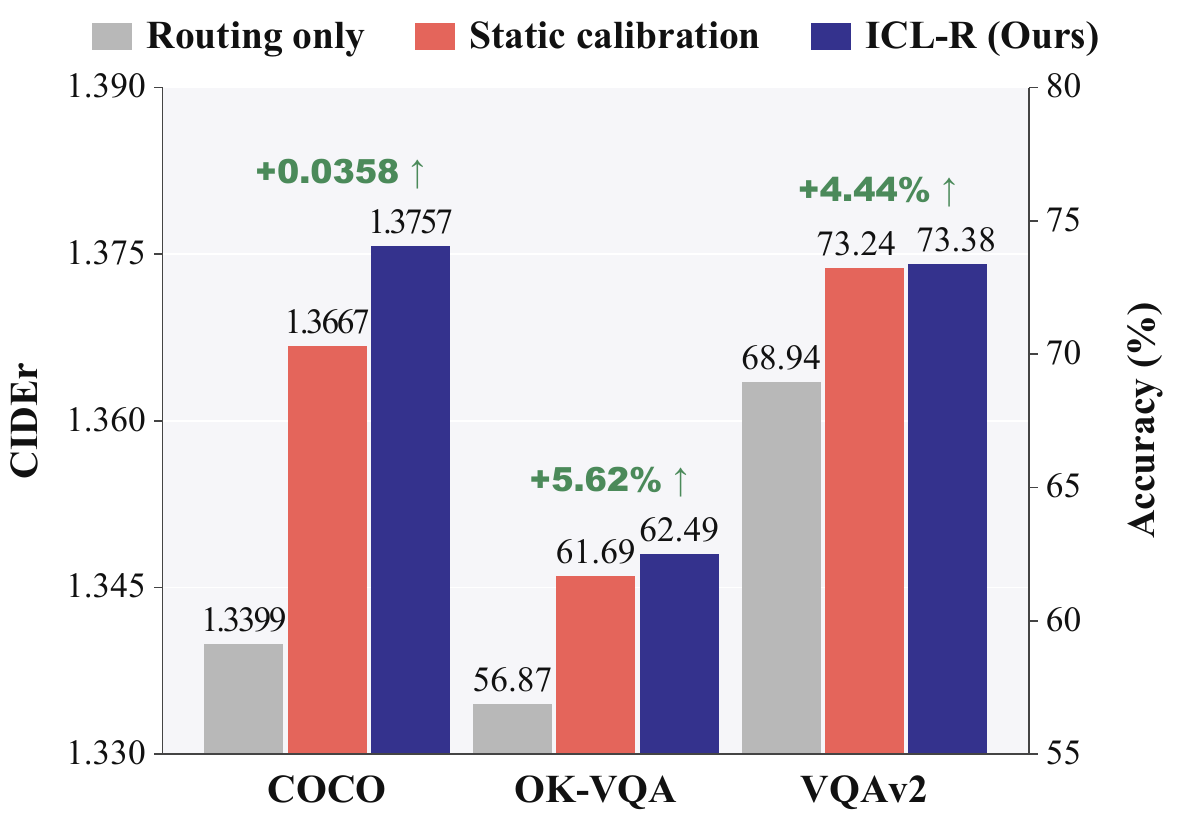}
  \vspace{-8pt}
  \caption{Ablation of value calibration and query-adaptive modulation in ICL-R on Idefics2 across three benchmarks.}
  \label{fig:component-ablation-1}
\end{wrapfigure}
\noindent\textbf{Core component ablations.}\label{sec:ablations}
To assess the contributions of our key designs, we compare our method with routing-only and static calibration variants in Figure~\ref{fig:component-ablation-1}.
Routing only removes value calibration and allocates the full rank budget to attention routing, keeping the total rank equal to ICL-R for a fair comparison.
Static calibration ablates query-adaptive modulation while retaining joint attention value calibration.
Compared with routing-only calibration, ICL-R improves accuracy by 5.62 percentage points on OK-VQA from 56.87 to 62.49 and 4.44 points on VQAv2 from 68.94 to 73.38, while increasing COCO Caption CIDEr by 0.0358 from 1.3399 to 1.3757.
Our method consistently outperforms the routing-only variant, indicating that increasing routing capacity alone cannot compensate for the absence of value calibration.
Its advantage over static calibration further suggests that query-adaptive modulation provides useful flexibility beyond fixed correction.

\par\medskip
\Needspace{8\baselineskip}
\begingroup
\setlength{\intextsep}{4pt}
\setlength{\columnsep}{0.02\linewidth}
\begin{wraptable}{r}{0.47\linewidth}
  \vspace{0pt}
  \centering
  \caption{Accuracy on VQAv2 and OK-VQA with
  Idefics-9B under the 32-shot setting. Bold marks the best result for
  each benchmark.}
  \label{tab:32shot-objective-ablation}
  \scriptsize
  \setlength{\tabcolsep}{0.5pt}
  \renewcommand{\arraystretch}{1.15}
  \begin{tabularx}{\linewidth}{@{}>{\raggedright\arraybackslash}p{0.22\linewidth}*{4}{>{\centering\arraybackslash}X}@{}}
    \toprule
     & CE-only & Query-only & RST-only & ICL-R \\
    \midrule
    VQAv2 & 44.25 & 52.69 & 55.90 & \textbf{57.10} \\
    OK-VQA & 43.98 & 45.78 & 47.21 & \textbf{50.63} \\
    \bottomrule
  \end{tabularx}
\end{wraptable}

To further probe whether demonstrations provide useful distillation supervision beyond labels and query-only teacher states, we compare four variants using the same Idefics-9B teacher and student under the 32-shot setting: CE-only, CE with query-only teacher targets (Query-only), demonstration-conditioned RST without CE (RST-only), and full ICL-R combining CE and our RST loss. 
CE-only and Query-only provide references for learning from ground-truth answers and intermediate alignment without demonstrations, while RST-only tests whether demonstration-conditioned targets can provide effective supervision on their own. 
As Table~\ref{tab:32shot-objective-ablation} shows, although Query-only already improves over CE-only, ICL-R achieves additional gains of 4.41 and 4.85 percentage points on VQAv2 and OK-VQA, supporting the value of demonstration conditioning beyond query-only alignment. 
The strong RST-only performance on OK-VQA(47.21) and VQAv2 (55.90) indicates that demonstration-conditioned intermediate states retain task-relevant information that can guide student without a direct ground-truth CE term. 
Thus, these results support effective distillation of demonstration.
\par
\endgroup
\vspace{-2pt}

\par\medskip
To assess transition-based supervision in RST, we compare five alternatives matching hidden-state endpoints or attention updates in different geometries. 
We follow the notation in Sec.~\ref{sec:riemannian-loss}, with $z_i^a=h_i^a+o_i^a$ denoting the unnormalized post-attention state. 
The spherical distance $d_{\mathbb S}$ is the geodesic distance on the unit sphere. 
For hyperbolic endpoint matching, $\Phi=\operatorname{Exp}o\circ\operatorname{LN}$ applies layer normalization followed by the Lorentz exponential map at the origin, with curvature $\kappa=-0.1$ and $d{\mathbb H}$ denotes the corresponding hyperbolic distance.
Table~\ref{tab:component-ablation-2} shows that RST leads on OK-VQA (62.49) and COCO CIDEr (1.3757), improving over paired spherical endpoint matching (61.64 and 1.3707). The closest alternatives are spherical transitions without parallel transport on OK-VQA (62.03) and state--update $\ell_2$ on COCO (1.3742). These results favor jointly supervising the starting state and attention-induced transition, with additional gains from parallel transport.
\begin{table}[H]
  \centering
  \caption{Comparison of intermediate-representation losses on
  OK-VQA and COCO Caption with Idefics2.
  Each alternative replaces only the RST term in ICL-R,
  retaining the same student architecture and CE supervision.
  Bold marks the best result in each column.}
  \label{tab:component-ablation-2}
  \footnotesize
  \setlength{\tabcolsep}{2pt}
  \renewcommand{\arraystretch}{1.35}
  \begin{tabularx}{\linewidth}{@{}>{\raggedright\arraybackslash}p{0.34\linewidth}X>{\centering\arraybackslash}p{0.09\linewidth}>{\centering\arraybackslash}p{0.09\linewidth}@{}}
    \toprule
    Loss & Objective & OK-VQA & COCO \\
    \midrule

    Hyperbolic endpoint
    & \mbox{\scriptsize $\frac{1}{2}\bigl[d_{\mathbb{H}}^2(\Phi(h_i^S),\Phi(h_i^T))+d_{\mathbb{H}}^2(\Phi(z_i^S),\Phi(z_i^T))\bigr]$}
    & 59.62 & 1.2377 \\
    \addlinespace[2pt]

    Endpoint $\ell_2$
    & \mbox{\scriptsize $\frac{1}{2}\bigl[\|h_i^S-h_i^T\|_2^2+\|z_i^S-z_i^T\|_2^2\bigr]$}
    & 61.91 & 1.3578 \\
    \addlinespace[2pt]

    Spherical endpoint $\ell_2$
    & \mbox{\scriptsize $\frac{1}{2}\bigl[\|x_i^S-x_i^T\|_2^2+\|y_i^S-y_i^T\|_2^2\bigr]$}
    & 61.64 & 1.3707 \\
    \addlinespace[2pt]

    State--update $\ell_2$
    & \mbox{\scriptsize $\frac{1}{2}\bigl[\|h_i^S-h_i^T\|_2^2+\|o_i^S-o_i^T\|_2^2\bigr]$}
    & 61.86 & 1.3742 \\
    \addlinespace[2pt]

    Spherical transition (no PT)
    & \mbox{\scriptsize $d_{\mathbb{S}}^2(x_i^S,x_i^T)+\|v_i^S-v_i^T\|_2^2$}
    & 62.03 & 1.3708 \\
    \addlinespace[2pt]

    \textbf{ICL-R (RST)}
    & \mbox{\scriptsize $d_{\mathbb{S}}^2(x_i^S,x_i^T)+\|v_i^S-\widetilde v_i^T\|_2^2$}
    & \textbf{62.49} & \textbf{1.3757} \\

    \bottomrule
  \end{tabularx}
\end{table}
\par
\endgroup
\vspace{-2pt}

\noindent\textbf{Data efficiency analysis.}\label{sec:training-size-analysis}
Having established the effectiveness of its individual components, we next assess the data efficiency and scaling behavior of our method by varying the training set size from 100 to 1,000 instances and comparing it with LoRA, MimIC, and HiFICL in Figure~\ref{fig:training-size}. 
Our method achieves the strongest performance across most training sizes on both VQAv2 and COCO settings with Idefics2 and LLaVA-Interleave. 
In particular, its performance on LLaVA-Interleave remains high even with only 100 examples, whereas the competing methods require substantially more data to approach the same level. On COCO with Idefics2, ICL-R is less effective in the extreme 100-example regime but improves rapidly as the training set grows and outperforms all baselines from 300 examples onward. 
Overall, these results show that our method can reconstruct useful demonstration effects from limited supervision while continuing to benefit from larger training sets.
This pattern holds across tasks and model backbones, showing that ICL-R remains effective across data regimes and that its data efficiency is not limited to a particular benchmark or model architecture.

\begin{figure}[!t]
  \centering
  \setlength{\abovecaptionskip}{4pt}
  \includegraphics[width=\linewidth]{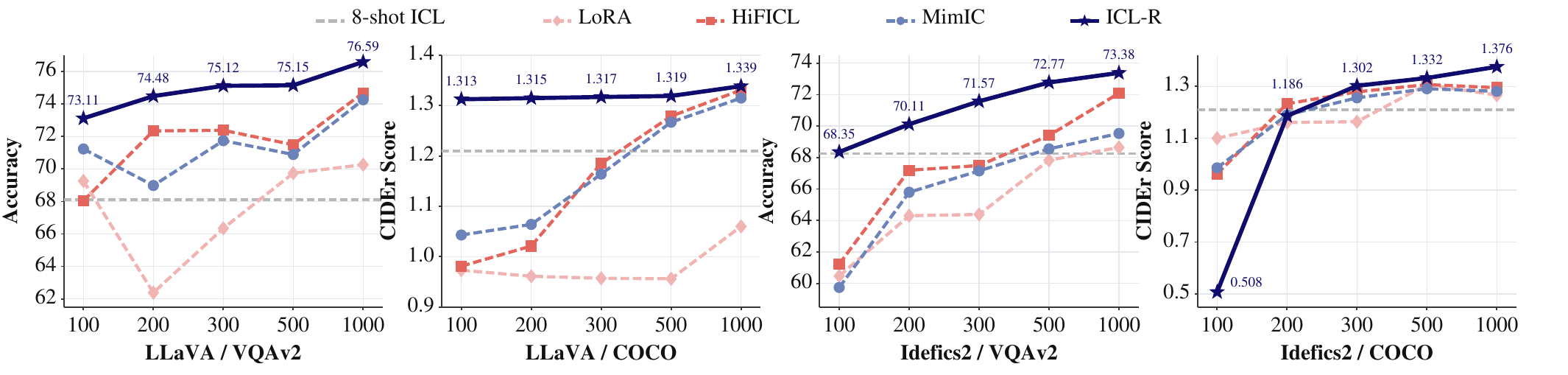}
  \caption{Performance under different training set sizes on VQAv2 and COCO Caption. ICL-R generally provides stronger data efficiency and more consistent scaling than the compared methods.}
  \label{fig:training-size}
\end{figure}

\Needspace{0.30\textheight}
\begingroup
\setlength{\intextsep}{4pt}
\setlength{\abovecaptionskip}{3pt}
\setlength{\belowcaptionskip}{3pt}
\phantomsection\label{sec:parameter-analysis}
\begin{wraptable}{r}{0.43\linewidth}
  \vspace{0pt}
  \centering
  \caption{Rank allocation between attention and value calibration under a fixed total rank with $\Delta$ denoting the change in percentage points relative to the 8/8 allocation. }
  \label{tab:parameter-quantity}
  \scriptsize
  \setlength{\tabcolsep}{1.5pt}
  \renewcommand{\arraystretch}{1.18}
  \begin{tabular*}{\linewidth}{@{\extracolsep{\fill}}ccccc@{}}
    \toprule
    $r_A/r_V$ & OK-VQA & $\Delta$ & VQAv2 & $\Delta$ \\
    \midrule
    2/14  & 56.76 & $-0.62$ & 76.538          & $-0.047$ \\
    4/12  & 57.40          & $+0.02 $ & \textbf{76.644} & $+0.059$ \\
    8/8   & 57.38          & --       & 76.585          & --       \\
    12/4  & 56.18          & $-1.20$ & 76.359          & $-0.226$ \\
    14/2  & \textbf{57.50}         & $+0.12$ & 76.367          & $-0.218$ \\
    \bottomrule
  \end{tabular*}
\end{wraptable}
\noindent\textbf{Attention-value rank allocation.}\label{sec:rank-allocation}
To investigate how capacity allocation between attention and value calibration affects performance, we vary their respective ranks while fixing the total rank at $r_A+r_V=16$ to keep the total number of trainable parameters unchanged.
Table~\ref{tab:parameter-quantity} reports results on OK-VQA and VQAv2 with LLaVA-Interleave and the reported deltas are measured relative to the balanced 8/8 allocation. 
The best allocation differs across tasks, with 14/2 achieving the highest accuracy on OK-VQA and 4/12 performing best on VQAv2, while gains over 8/8 remain small. 
Under the fixed parameter budget, these results suggest that the preferred capacity allocation depends on the task, with the balanced configuration providing competitive performance across both benchmarks.
\par
\endgroup
\vspace{-2pt}

\noindent\textbf{Total rank sensitivity and gating behavior. }\label{sec:total-rank}
To study calibration capacity, we vary rank configurations of 1/1, 2/2, 4/4, and 8/8 on Idefics2 with OK-VQA and COCO in Figure~\ref{fig:total-rank}.
Increasing rank generally improves performance: OK-VQA and COCO continue to improve with relatively weak gains from 4/4 to 8/8.
To determine whether these gains merely reflect larger corrections, we analyze the learned static output scales.
Each branch yields a \(32\times32\) matrix of \(s_A\) or \(s_V\), with one entry per layer--head pair.
The center heatmaps show normalized task-averaged \(\tanh(s_A)\) and \(\tanh(s_V)\) at rank 4/4, revealing distinct layer--head structures, while the right panels show their absolute-value distributions across tasks and rank settings.
Median magnitudes remain stable or decrease slightly with rank, indicating that additional rank enriches the low-rank representation and improves calibration capacity without increasing static gate scales.
\begin{figure}[!t]
  \centering
  \setlength{\abovecaptionskip}{4pt}
  \includegraphics[width=\linewidth]{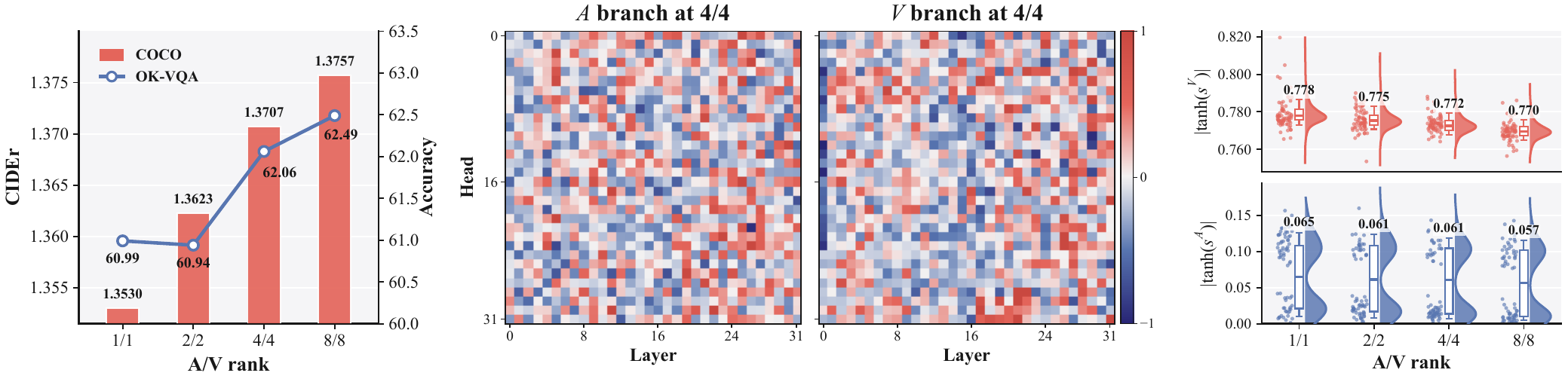}
  \caption{Rank sensitivity and learned calibration scales on Idefics2. Left: COCO CIDEr and OK-VQA accuracy across ranks. Center: normalized attention and value scales at rank 4/4. Right: distributions of their absolute magnitudes across both tasks and rank configurations.}
  \label{fig:total-rank}
\end{figure}

\noindent\textbf{Effect of the number of demonstrations.}\label{sec:number-of-demos}
We further examine the robustness of our method to the amount of demonstration context used during training. 
We vary the number of demonstrations from 4 to 48 and report the results in Figure~\ref{fig:shot-scaling-efficiency} (left). 
On VQAv2, explicit ICL degrades sharply beyond 16 demonstrations, whereas our method varies by only 0.125 points across all settings and exceeds explicit ICL by more than 22 points at 32 demonstrations. 
On COCO Caption, explicit ICL also fluctuates substantially and declines at 48 demonstrations. 
In contrast, ICL-R remains within a narrow range of 0.011 CIDEr, with its best result obtained using eight demonstrations. 
These small variations suggest that ICL-R remains robust to the growing number of redundant visual tokens introduced by additional demonstrations.
Thus, ICL-R preserves the benefits of demonstrations across context sizes, even when longer contexts degrade explicit ICL.

\noindent\textbf{Inference efficiency.}\label{sec:inference-efficiency}
We evaluate inference efficiency on VQAv2 with Idefics2-8B-Base using an NVIDIA RTX A6000 GPU, comparing the FLOPs and runtime of ICL-R with zero-shot inference and explicit ICL at different demonstration counts in Figure~\ref{fig:shot-scaling-efficiency} (right).
ICL-R remains close to zero-shot inference in computational cost, with approximately 98\% fewer FLOPs and 82\% lower runtime than 32-shot ICL.
The runtime reduction confirms that the savings in arithmetic operations translate into a substantial practical speedup in our evaluation setting.
Overall, these results show that ICL-R combines stable performance across demonstration counts with inference cost close to that of zero-shot inference, offering an effective balance between accuracy and efficiency.

\begingroup
\setlength{\intextsep}{0pt}
\setlength{\abovecaptionskip}{1pt}
\setlength{\belowcaptionskip}{-5pt}
\begin{figure}[!t]
  \vspace{-4pt}
  \centering
  \includegraphics[width=\linewidth]{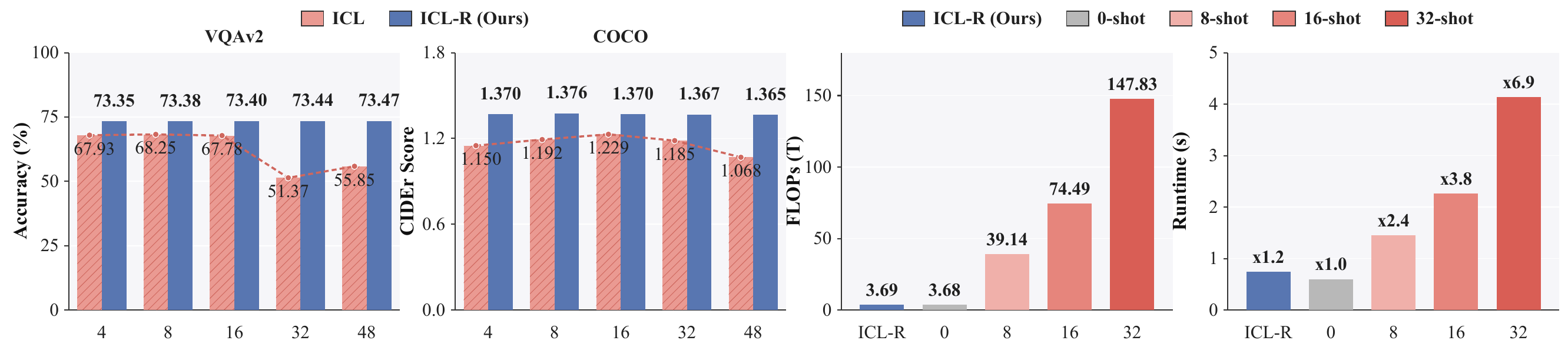}
  \vspace{-10pt}
  \caption{Performance and inference efficiency comparisons. Left: VQAv2 accuracy and COCO CIDEr for ICL-R and explicit ICL with varying demonstration counts. Right: FLOPs and runtime of ICL-R, zero-shot inference, and explicit ICL with different context sizes.}
  \label{fig:shot-scaling-efficiency}
  \vspace{-4pt}
\end{figure}
\endgroup
\vspace{-2pt}
\noindent\textbf{Hallucination analysis.}\label{sec:hallucination-analysis}
To assess whether ICL-R reduces object hallucination while preserving coverage, we compare CHAIR \citep{rohrbach2019objecthallucinationimagecaptioning} scores and object recall with zero-shot inference, explicit ICL, and trainable baselines in Table~\ref{tab:hallucination-analysis}.
ICL-R achieves the lowest instance-level CHAIR of 2.0 and the highest object recall of 46.0 among all compared methods.
Compared with 8-shot ICL, it reduces the instance-level score from 2.3 to 2.0 while maintaining the same sentence-level score of 3.1 and increasing recall from 40.3 to 46.0.
These results demonstrate that ICL-R provides a favorable balance between reducing object hallucination and preserving object coverage.

\begingroup
\setlength{\intextsep}{4pt}
\setlength{\abovecaptionskip}{3pt}
\setlength{\belowcaptionskip}{3pt}
\begin{table}[h]
  \centering
  \caption{Hallucination performance measured by CHAIR and object recall on Idefics2. Our method achieves the lowest CHAIR$_{\mathrm{I}}$ and highest object recall among compared approaches.}
  \label{tab:hallucination-analysis}
  \footnotesize
  \setlength{\tabcolsep}{1.5pt}
  \renewcommand{\arraystretch}{1.08}
  \begin{tabular*}{\linewidth}{@{\extracolsep{\fill}}lcccccccccc@{}}
    \toprule
    & 0-shot & 4-shot & 8-shot & 16-shot & 32-shot & LoRA
    & LIVE & MimIC & HiFICL & \textbf{ICL-R(Ours)} \\
    \midrule
    CHAIR$_{\mathrm{S}}$ $\downarrow$ & \textbf{2.8} & 3.4 & 3.1 & 3.2 &3.9 &3.3  & 4.4  & 4.0  & 3.2  & 3.1  \\
    CHAIR$_{\mathrm{I}}$ $\downarrow$ & 3.3 & 2.5 & 2.3 & 2.4 &2.4 &2.4  & 3.1  & 2.9  & 2.2  & \textbf{2.0}  \\
    Recall $\uparrow$                  & 27.2 & 41.3 & 40.3 & 41.3& 42.1 & 44.3 & 44.8 & 45.4 & 45.7 & \textbf{46.0} \\
    \bottomrule
  \end{tabular*}
\end{table}
\endgroup
\vspace{-2pt}

\endgroup

\section{Conclusion}\label{sec:conclusion}
In this work, we introduced ICL-R, a lightweight framework for reconstructing demonstration-induced computation in frozen MLLMs. The framework jointly models attention routing and value transformations, while using internal state transitions to provide supervision beyond endpoint matching. By preserving the benefits of few-shot prompting while reducing context length and inference latency, ICL-R offers a practical route to efficient multimodal adaptation. 
Future work will assess its generality across a broader range of models and datasets and investigate how the framework can be extended to video understanding to reconstruct demonstration effects on temporal reasoning.
\subsection*{AI Use Statement}
Generative AI tools were used to assist with language polishing, LaTeX formatting, figure-layout code, and
the organization across the paper. The research ideas, method design, experimental data, and quantitative results were produced and verified by the authors. All AI-assisted content was manually reviewed and revised, and the authors take full responsibility for the final text, claims, equations, citations, and reported results.

\subsection*{Reproducibility Statement}
We specify the complete formulation, model components, and training objective in Section~\ref{sec:method} and report the evaluated backbones, datasets, metrics, trainable modules and optimization setup in Section~\ref{sec:implementation}. Appendix~\ref{app:complete-derivation} gives the complete derivation of the formulation, and Appendix~\ref{app:algorithmic-procedure} shows the general procedure of the algorithm. Appendix~\ref{app:fused-sdpa} describes the memory-efficient fused SDPA implementation. Overall, these details support independent reproduction of our work.

\bibliography{iclr2027_conference}
\bibliographystyle{iclr2027_conference}

\appendix

\clearpage
\etocdepthtag.toc{appendix}
\begingroup
\etocsetnexttocdepth{subsubsection}
\etocsettagdepth{appendix}{subsubsection}
\etocsettocstyle{\section*{Appendix Contents}\setcounter{tocdepth}{-2}}{}
\tableofcontents
\endgroup
\clearpage

\section{Appendix}\label{sec:appendix}
\raggedbottom
\subsection{Complete Derivation of the Formulation}\label{app:complete-derivation}

We provide the complete derivation of our decomposition in Section~\ref{sec:problem-formulation}. The derivation is exact for each self-attention head. Let $n_X$ and $n_D$ denote the numbers of query and demonstration tokens, respectively, as in the main text. Then
$A_{T,D}\in\mathbb{R}^{n_X\times n_D}$,
$A_{T,X},A_0\in\mathbb{R}^{n_X\times n_X}$,
$V_{T,D}\in\mathbb{R}^{n_D\times d_h}$, and
$V_{T,X},V_0\in\mathbb{R}^{n_X\times d_h}$.

\noindent\textbf{Step 1: Partition the teacher output.}
For the query-aligned attention rows, the teacher attends to both demonstration and query tokens, whereas the query-only model attends only to query tokens. Their outputs are:
\begin{equation}
\label{eq:app-teacher-output}
O_T=A_{T,D}V_{T,D}+A_{T,X}V_{T,X},
\qquad
O_0=A_0V_0.
\end{equation}

\noindent\textbf{Step 2: Express the query-side changes.}
Using the definitions from Section~\ref{sec:problem-formulation}:
\begin{equation}
\label{eq:app-query-side-changes}
\begin{aligned}
\Delta A_T=A_{T,X}-A_0
&\quad\Longleftrightarrow\quad
A_{T,X}=A_0+\Delta A_T,\\
\Delta V_T=V_{T,X}-V_0
&\quad\Longleftrightarrow\quad
V_{T,X}=V_0+\Delta V_T.
\end{aligned}
\end{equation}

\noindent\textbf{Step 3: Expand the output difference.}
Substituting these identities into the teacher output gives:
\begin{equation}
\label{eq:app-output-expansion}
\begin{aligned}
O_T-O_0
&=A_{T,D}V_{T,D}
 +(A_0+\Delta A_T)(V_0+\Delta V_T)-A_0V_0\\
&=A_{T,D}V_{T,D}
 +A_0V_0+A_0\Delta V_T
 +\Delta A_TV_0+\Delta A_T\Delta V_T-A_0V_0.
\end{aligned}
\end{equation}
After canceling the two $A_0V_0$ terms and arranging the remaining terms by their source, we obtain:
\begin{equation}
\label{eq:app-output-decomposition}
\begin{aligned}
O_T-O_0
=\;&\underbrace{\Delta A_TV_0}_{\text{routing shift}}
+\underbrace{A_0\Delta V_T}_{\text{value shift}}
+\underbrace{\Delta A_T\Delta V_T}_{\text{interaction}}
+\underbrace{A_{T,D}V_{T,D}}_{\text{demonstration information}}.
\end{aligned}
\end{equation}
The first term changes attention routing while holding the base values fixed. The second changes value content under the base routing. The third is the coupled effect that appears only when routing and values change together. The final term is information read directly from demonstration tokens. Because the query-only student cannot form this final term explicitly, its effect must be absorbed into the calibrated query-side output.

\noindent\textbf{Step 4: Derive the missing-mass factorization.}
For query token $t$, define the teacher attention mass assigned to demonstrations as:
\begin{equation}
\label{eq:app-demonstration-mass}
m_t^T=\sum_{s\in\mathcal D}(A_{T,D})_{t,s},
\qquad
\sum_{s\in X}(A_{T,X})_{t,s}=1-m_t^T.
\end{equation}
For $m_t^T\in(0,1)$, normalize the teacher attention within the query block as
\begin{equation}
\label{eq:app-normalized-query-attention}
(\overline A_{T,X})_{t,s}
=\frac{(A_{T,X})_{t,s}}{1-m_t^T},
\qquad s\in X,
\end{equation}
and define the attention-weighted demonstration information conditioned on attending to demonstrations as:
\begin{equation}
\label{eq:app-conditional-demonstration-content}
E_t^T
=\frac{1}{m_t^T}
\sum_{s\in\mathcal D}
(A_{T,D})_{t,s}(V_{T,D})_s.
\end{equation}
The teacher output for token $t$ can then be rewritten exactly as:
\begin{equation}
\label{eq:app-missing-mass-factorization}
\begin{aligned}
O_{T,t}
&=\sum_{s\in X}(A_{T,X})_{t,s}(V_{T,X})_s
 +\sum_{s\in\mathcal D}(A_{T,D})_{t,s}(V_{T,D})_s\\
&=(1-m_t^T)
\sum_{s\in X}(\overline A_{T,X})_{t,s}(V_{T,X})_s
 +m_t^TE_t^T.
\end{aligned}
\end{equation}
The first term is the normalized query-side aggregation multiplied by its true total mass, while the second is the conditional content obtained from demonstration tokens. 
A query-only softmax instead forces its attention mass over query tokens to sum to one and therefore omits the coefficient $1-m_t^T$. 
The missing-mass estimator introduces a learned coefficient motivated by this factor, and the joint attention and value calibration reconstructs the routing and content needed to approximate the full teacher output. The limiting cases $m_t^T=0$ and $m_t^T=1$ follow directly by removing the absent component. For multi-head attention, the same factorization applies independently to each head, and the output projection preserves the equality by linearity.

\subsection{Implementation Details}\label{app:implementation-details}
\noindent\textbf{Training and evaluation protocol.}
For VQAv2, OK-VQA, COCO Caption, and Flickr30K, we randomly sample 1,000
instances from the corresponding training split to form the support/training
set.  Because MME and SEED-Bench do not provide an equivalent training split
for our setting, we instead construct disjoint support and evaluation
partitions from their released data.  Each training episode
contains eight demonstrations sampled from the support set and an independent
query.  The frozen teacher processes the demonstrations together with the
query, whereas the student receives only the query.  At inference, all
trainable methods receive the query alone; only vanilla ICL retains eight
demonstrations in the input context.

All parameters of the vision encoder, multimodal projector, and language model
remain frozen.  Only our proposed modules are optimized using the task loss and our Riemannian
State Transition (RST) loss.  By default, the ranks of the attention and value
calibration modules are both set to 8.  We use 16-bit mixed precision
(\texttt{16-mixed}) and AdamW~\citep{loshchilov2019decoupled}, with cosine
learning-rate decay and linear warmup over the first 10\% of the optimization
steps.

We use 10,000 examples from the VQAv2 validation set and the complete validation
splits of OK-VQA (5,046 examples), COCO Caption under the Karpathy split \citep{karpathy2015deep} (5,000
examples), and Flickr30K under the Karpathy split (1,014 examples).  For MME,
we deterministically use the first 1,000 question--answer pairs (500
images) as support and remaining 1,374 pairs (687 images) for evaluation;
the two partitions have no image overlap.  For the image subset of SEED-Bench,
we use the first 1,000 image questions as support and a fixed sample of 10,000
questions from the remaining image questions for evaluation.  The exact
partitions and metrics are summarized in Table~\ref{tab:app_data_splits}.

\begin{table}[H]
    \centering
    \small
    \setlength{\tabcolsep}{5.5pt}
    \begin{tabular}{l r r l}
        \toprule
        Dataset & $K$ &  $N_{\mathrm{eval}}$ & Metric \\
        \midrule
        VQAv2        &  1,000 &  10,000 & VQA Acc. \\
        OK-VQA       &  1,000 &  5,046  & VQA Acc. \\
        COCO Caption &  1,000 &  5,000  & CIDEr \\
        MME          &  1,000 &  1,374  & Accuracy \\
        SEED-Bench   &  1,000 &  10,000 & Exact match \\
        Flickr30K    &  1,000 &  1,014  & CIDEr \\
        \bottomrule
    \end{tabular}
    \caption{Training and evaluation partitions. $K$ is the size of
    the training set, and $N_{\mathrm{eval}}$ is the number of evaluated
    queries.}
    \label{tab:app_data_splits}
\end{table}

For VQAv2 and OK-VQA, the first annotated answer is used as the training
target, while evaluation uses all human-annotated answers and the standard VQA
accuracy.  For COCO Caption and Flickr30K, the first reference caption is used
as the training target, while evaluation computes CIDEr against all five
references.  MME is evaluated using question-level yes/no accuracy and SEED-Bench is evaluated by exact matching of the predicted option in
\{A, B, C, D\}.

\noindent\textbf{Benchmarks.}
VQAv2~\citep{goyal2017making} evaluates general visual question answering on
MS COCO images, whereas OK-VQA~\citep{marino2019okvqa} additionally requires
external knowledge beyond the visible image.  COCO Caption~\citep{chen2015microsoft}
and Flickr30K~\citep{young2014image} evaluate open-ended image description.
MME~\citep{fu2025mme} covers a broad collection of perception and cognition
abilities using paired yes/no questions, and SEED-Bench~\citep{li2024seedbench}
evaluates multimodal perception and reasoning in a multiple-choice format.
We report VQA accuracy for VQAv2 and OK-VQA, question-level accuracy for MME,
exact match for SEED-Bench, and CIDEr~\citep{vedantam2015cider} for COCO
Caption and Flickr30K.

\noindent\textbf{Numerical treatment of RST.}
We evaluate the geometric operations in float32 with a numerical tolerance of $\varepsilon = 10^{-6}$.
State normalization is implemented as $\pi_\varepsilon(z) = z / \max(\|z\|_2, \varepsilon)$.
For the spherical logarithmic map, we compute $c = \operatorname{clip}(\langle x,y\rangle,-1+\varepsilon,1)$, form $u = y-cx$, and obtain the angle using $\theta = \operatorname{atan2}(\|u\|_2,c)$.
When $\|u\|_2 \leq \varepsilon$, the scaling factor is set to one; otherwise, it is $\theta/\|u\|_2$.
Near coincident points, this implements the first-order approximation $\operatorname{Log}_x(y) \approx u$, recovering the zero transition at coincidence.
For parallel transport, we lower-bound $1+\langle x^T,x^S\rangle$ by $\varepsilon$.
At exactly antipodal pairs, the spherical logarithmic map and shortest-geodesic parallel transport are not uniquely defined. When the clipping or denominator bound is activated near these singularities, the resulting quantities serve as numerical surrogates.
The half-chord argument in the point-distance calculation is capped at one before $\arcsin$.

\subsection{Algorithmic Procedure of ICL-R}\label{app:algorithmic-procedure}

To clarify how the proposed components are integrated into a complete workflow, we provide pseudocode for the training and inference procedures below.
Algorithm~\ref{alg:icl-r-training} summarizes the overall training and inference procedure of ICL-R.
During training, the frozen teacher and query-only student are evaluated under teacher forcing, after which the student is optimized by combining task supervision with Riemannian State Transition (RST) alignment while the backbone remains fixed; at inference, the trained student generates answers from the query alone.
Algorithm~\ref{alg:icl-r-calibration} further details the student attention computation, including query-adaptive gating, low-rank corrections to attention logits and values, missing-mass estimation, and aggregation with the resulting calibrated attention weights.

\begin{algorithm}[H]
\caption{Training and inference procedure of ICL-R.}
\label{alg:icl-r-training}
\begin{algorithmic}[1]
\Require Frozen backbone parameters $\theta$; training episodes containing demonstrations $\mathcal D$, query $X$, and target answer $Y$; loss weight $\lambda$
\Ensure Trained calibration parameters $\phi$
\State Freeze $\theta$ and initialize $\phi$, including the attention and value calibration modules and the missing-mass estimator
\ForAll{training episodes $(\mathcal D,X,Y)$}
    \State Run the frozen teacher on $[\mathcal D;X]$ with teacher forcing using $Y$; record $h_i^T$ and $o_i^T$
    \State Run the student on $X$ with teacher forcing using $Y$ and \Call{CalibratedAttention}{$Q_0,K_0,V_0;\phi$}; record $h_i^S$ and $o_i^S$
    \ForAll{aligned token--layer indices $i\in\mathcal I$}
        \For{$a\in\{T,S\}$}
            \State $x_i^a\gets\pi(h_i^a)$ and $y_i^a\gets\pi(h_i^a+o_i^a)$
            \State $v_i^a\gets\operatorname{Log}_{x_i^a}(y_i^a)$
        \EndFor
        \State $\widetilde v_i^T\gets\operatorname{PT}_{x_i^T\rightarrow x_i^S}(v_i^T)$
    \EndFor
    \State $\mathcal L_{\mathrm{RST}}\gets\mathcal L_{\mathrm{point}}+\mathcal L_{\mathrm{fiber}}$
    \State $\mathcal L_{\mathrm{total}}\gets\mathcal L_{\mathrm{CE}}(Y)+\lambda\mathcal L_{\mathrm{RST}}$
    \State Update only $\phi$ to minimize $\mathcal L_{\mathrm{total}}$; keep $\theta$ frozen
\EndFor
\Statex
\State \textbf{Inference:} generate the answer with the trained student using only query $X$
\end{algorithmic}
\end{algorithm}

\begin{algorithm}[H]
\caption{Jointly calibrated attention used by the ICL-R student.}
\label{alg:icl-r-calibration}
\begin{algorithmic}[1]
\Function{CalibratedAttention}{$Q_0,K_0,V_0;\phi$}
    \ForAll{query tokens $t$}
        \State Compute $\bar q_t=\operatorname{LN}(q_t)$ and predict $g_t^A,\rho_t^A,g_t^V,\rho_t^V$ from the current query
        \State $\Delta S^A_{t,s}\gets g_t^A\left[\rho_t^A\odot(Q_0U_Q)_{t,:}\right](K_0U_K)_{s,:}^{\top}$ for all allowed keys $s$
        \State $\Delta V_{t,s}\gets g_t^V\left[\left(\rho_t^V\odot V_{0,s}P_V\right)R_V\right]$ for all allowed keys $s$
        \State Predict $m_t\gets s_m\sigma\!\left(\bar q_t^{\top}w_m/\sqrt{d_h}+b_m\right)$
        \State $\widehat A_{t,:}\gets(1-m_t)\operatorname{softmax}\!\left((Q_0K_0^{\top}/\sqrt{d_h})_{t,:}+\Delta S^A_{t,:}\right)$
        \State $\widetilde O_t\gets\sum_s\widehat A_{t,s}\left(V_{0,s}+\Delta V_{t,s}\right)$
    \EndFor
    \State \Return $\widetilde O$
\EndFunction
\end{algorithmic}
\end{algorithm}

\subsection{Decomposing Demonstration-Induced Attention-Output Shifts}\label{app:decomposing-attention-output-shifts}

To understand what internal behavior ICL-R needs to recover beyond task-level predictions, we analyze how in-context demonstrations change the attention outputs of the teacher model. Specifically, we compare the attention output produced with demonstrations \(O_T\) against that of the query-only model \(O_0\) on 64 teacher-forced VQAv2 samples with eight demonstrations. After aligning the query tokens in the two input sequences, we exactly decompose the resulting output shift into routing, value, interaction, and direct-demonstration terms. We then investigate three complementary questions: how large each term is across layers and heads, whether each term aligns with or counteracts the total output shift, and how interactions between the terms relate to the attention probability mass assigned to demonstrations. This analysis provides a quantitative characterization of the internal effects that must be approximated when demonstrations are removed at inference time.

\subsubsection{Relative Magnitudes of the Four Decomposition Terms}\label{app:decomposition-magnitudes}
We first examine whether demonstration-induced attention-output shifts are dominated by a single mechanism such as attention rerouting. Let \(\Delta A=A_{T,X}-A_0\) and \(\Delta V=V_{T,X}-V_0\), where \(X\) denotes the aligned query tokens and \(D\) denotes the demonstration-only tokens. The total shift admits the exact decomposition:
\begin{equation}
\label{eq:app-decomposition-relative-magnitude}
O_T-O_0
=
\underbrace{\Delta A V_0}_{T_R}
+
\underbrace{A_0\Delta V}_{T_V}
+
\underbrace{\Delta A\Delta V}_{T_I}
+
\underbrace{A_{T,D}V_{T,D}}_{T_D},
\qquad
\rho_k=\frac{\lVert T_k\rVert_F}{\sum_j\lVert T_j\rVert_F}.
\end{equation}
Here, \(T_R\), \(T_V\), \(T_I\), and \(T_D\) denote the routing, value, interaction, and direct-demonstration terms, respectively, and \(\rho_k\) measures the normalized relative magnitude of each term for each layer and attention head. As shown in Figure~\ref{fig:decomposition-magnitude-heatmaps}, routing has the largest average relative magnitude at \(29.2\%\), followed by value at \(25.7\%\), direct demonstration retrieval at \(25.1\%\), and interaction at \(20.0\%\). The decomposition is numerically exact, with a mean relative reconstruction residual of \(2.09\times10^{-7}\) and a maximum residual of \(9.81\times10^{-7}\). Although routing is the largest individual term, it accounts for less than one third of the total normalized magnitude, while the other three terms remain substantial and exhibit considerable layer- and head-level heterogeneity. These results show that the effect of demonstrations cannot be reduced to attention reweighting alone: demonstrations simultaneously alter information routing, value representations, their interaction, and the direct evidence available to the model.

\begingroup
\setlength{\abovecaptionskip}{2pt}
\setlength{\belowcaptionskip}{0pt}
\begin{figure}[H]
    \centering
    \includegraphics[width=0.9\linewidth]{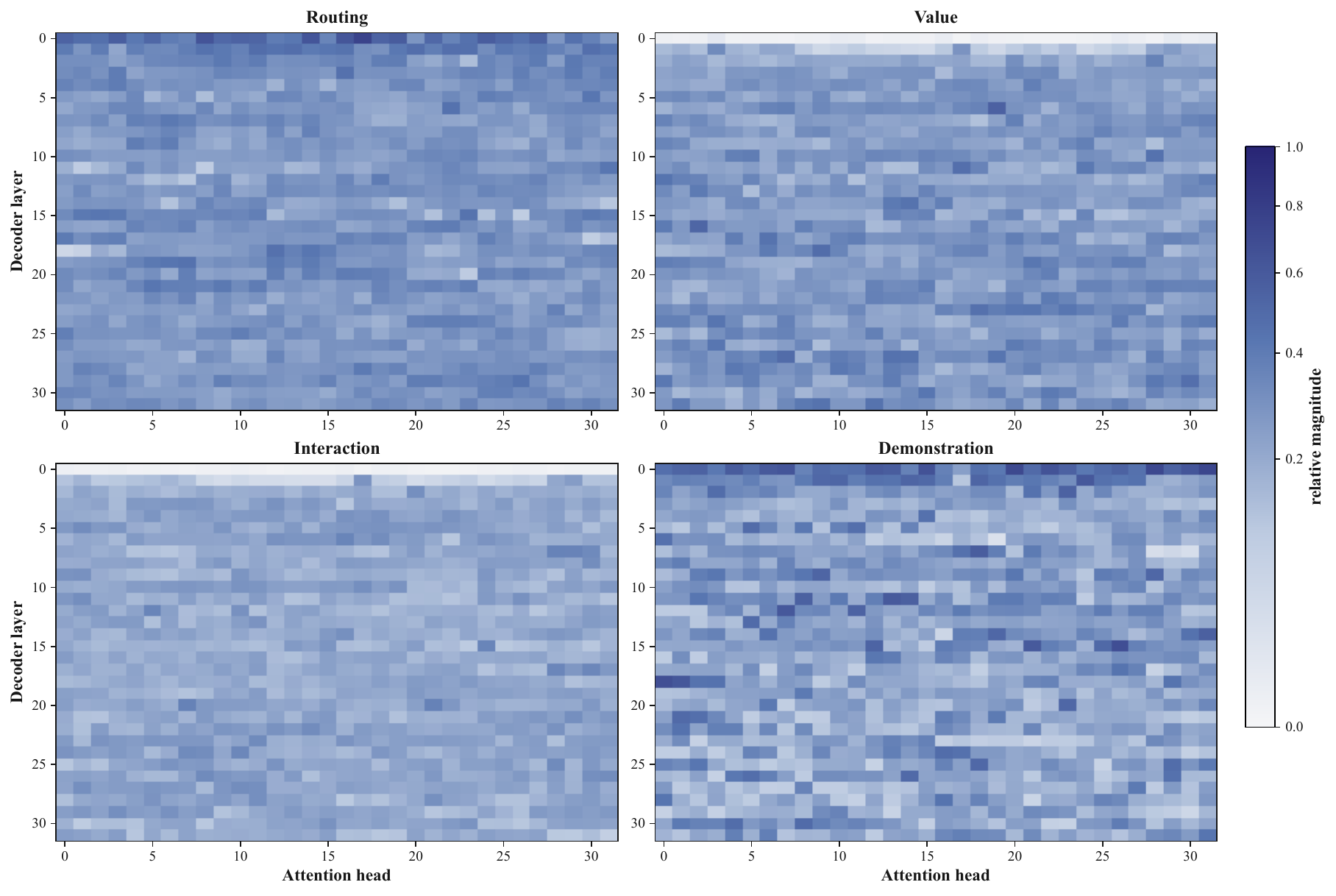}
    \vspace{-3pt}
    \caption{Relative magnitudes of the four terms in the demonstration-induced attention-output decomposition across model layers and attention heads.}
    \label{fig:decomposition-magnitude-heatmaps}
\end{figure}
\endgroup

\subsubsection{Layer-wise Dynamics and Alignment with the Total Shift}\label{app:decomposition-layerwise-dynamics}
We next study how the four mechanisms evolve with model depth and distinguish term magnitude from directional relevance. In addition to the layer-wise average relative magnitude \(\rho_{k,l}\), we calculate the cosine similarity between each term and the total attention-output shift,
\begin{equation}
\label{eq:app-layerwise-term-alignment}
c_{k,l}
=
\mathbb{E}_h
\left[
\frac{
\langle T_{k,l,h},\,O_{T,l,h}-O_{0,l,h}\rangle
}{
\lVert T_{k,l,h}\rVert_F
\lVert O_{T,l,h}-O_{0,l,h}\rVert_F
}
\right].
\end{equation}
A positive cosine indicates that a term generally moves the output in the direction of the total teacher-induced shift, whereas a negative cosine indicates a compensatory or cancelling effect. Figure~\ref{fig:decomposition-layerwise-curves} reveals a clear depth-dependent transition. From early to late layers, the relative magnitude of routing decreases from \(31.9\%\) to \(27.8\%\), while that of value increases from \(21.8\%\) to \(28.4\%\). The direct-demonstration term decreases from \(28.2\%\) to \(22.1\%\), whereas the interaction term increases from \(18.1\%\) to \(21.7\%\). Directionally, routing is most strongly aligned with the total shift, with an average cosine of \(0.574\), followed by the direct-demonstration and value terms at \(0.477\) and \(0.437\) respectively. In contrast, the interaction term has a substantial magnitude but a negative average cosine of \(-0.158\). These results suggest that early layers primarily modify attention routing and retrieve demonstration information, whereas deeper layers increasingly transform the representations being aggregated. They also show that a large term is not necessarily a direct positive contributor for the interaction term acts predominantly as a compensatory correction to the other components.

\begingroup
\setlength{\abovecaptionskip}{2pt}
\setlength{\belowcaptionskip}{0pt}
\begin{figure}[H]
    \centering
    \includegraphics[width=0.9\linewidth]{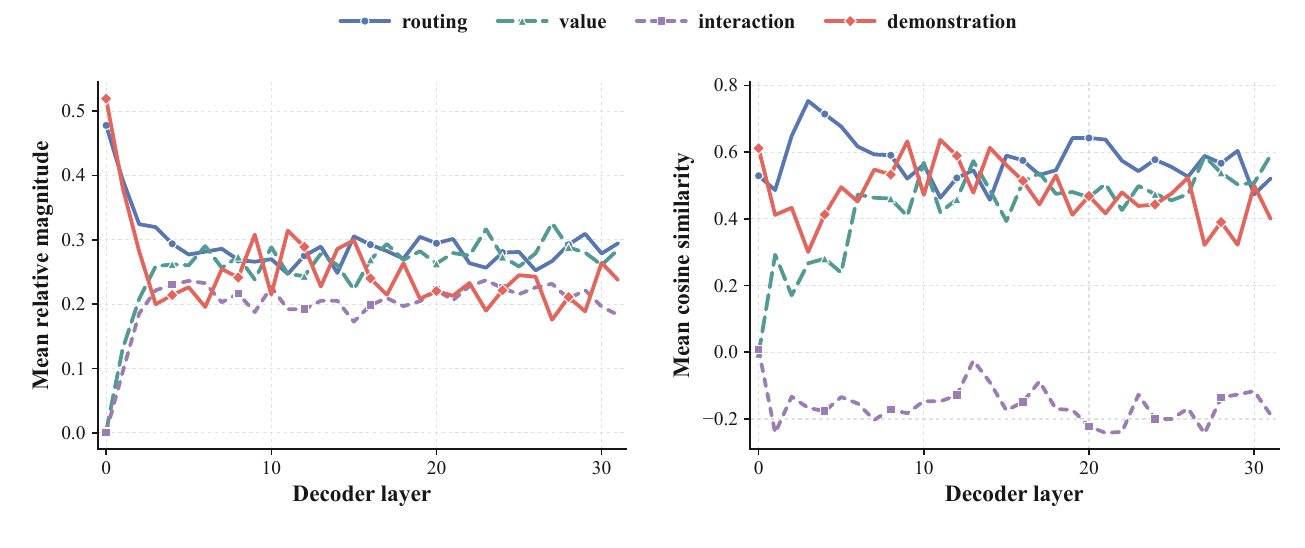}
    \vspace{-3pt}
    \caption{Layer-wise dynamics of the four decomposition terms and their alignment with the total demonstration-induced attention-output shift.}
    \label{fig:decomposition-layerwise-curves}
\end{figure}
\endgroup

\subsubsection{Term Coupling and Missing Attention Mass}\label{app:decomposition-coupling-mass}

Finally, we investigate whether the four decomposition terms operate independently and quantify the attention normalization mass that disappears when demonstrations are removed. We compute the pairwise cosine similarity and the teacher attention mass assigned to demonstration tokens as:
\begin{equation}
\label{eq:app-term-coupling-missing-mass}
C_{ij}
=
\frac{\langle T_i,T_j\rangle}
{\lVert T_i\rVert_F\lVert T_j\rVert_F},
\qquad
m^T_{l,h,i}
=
\sum_{j\in D}A^T_{l,h,i,j}.
\end{equation}
Negative values of \(C_{ij}\) indicate systematic cancellation. We average \(m^T_{l,h,i}\) over aligned query tokens and samples for each layer and head. Figure~\ref{fig:decomposition-coupling-mass} shows that the strongest coupling occurs between the value and interaction terms, whose cosine similarity is \(-0.729\). Routing is also negatively aligned with interaction and direct demonstration retrieval, with pairwise cosines of \(-0.250\) and \(-0.198\) respectively, while routing and value exhibit only a weak positive alignment of \(0.175\). This confirms that the four terms are strongly coupled and explains why their normalized magnitudes cannot be treated as independent additive contributions. Meanwhile, the teacher assigns an average attention mass of \(0.299\) to demonstrations. The missing mass is particularly large in early layers, decreasing from \(0.535\) in Layer 0 and \(0.473\) in Layer 1 to approximately \(0.250\) by Layer 7, with substantial variation across attention heads. Thus, removing demonstrations causes not only the loss of their value evidence but a considerable, layer-dependent change in attention normalization. This observation provides the mechanistic motivation for explicitly accounting for missing attention mass.
\begingroup
\setlength{\abovecaptionskip}{2pt}
\setlength{\belowcaptionskip}{0pt}
\begin{figure}[H]
    \centering
    \begin{minipage}[b]{0.44\linewidth}
        \centering
        \includegraphics[width=\linewidth]{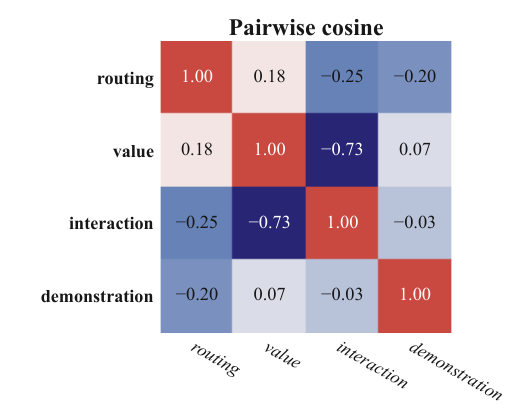}
    \end{minipage}\hfill
    \begin{minipage}[b]{0.54\linewidth}
        \centering
        \includegraphics[width=\linewidth]{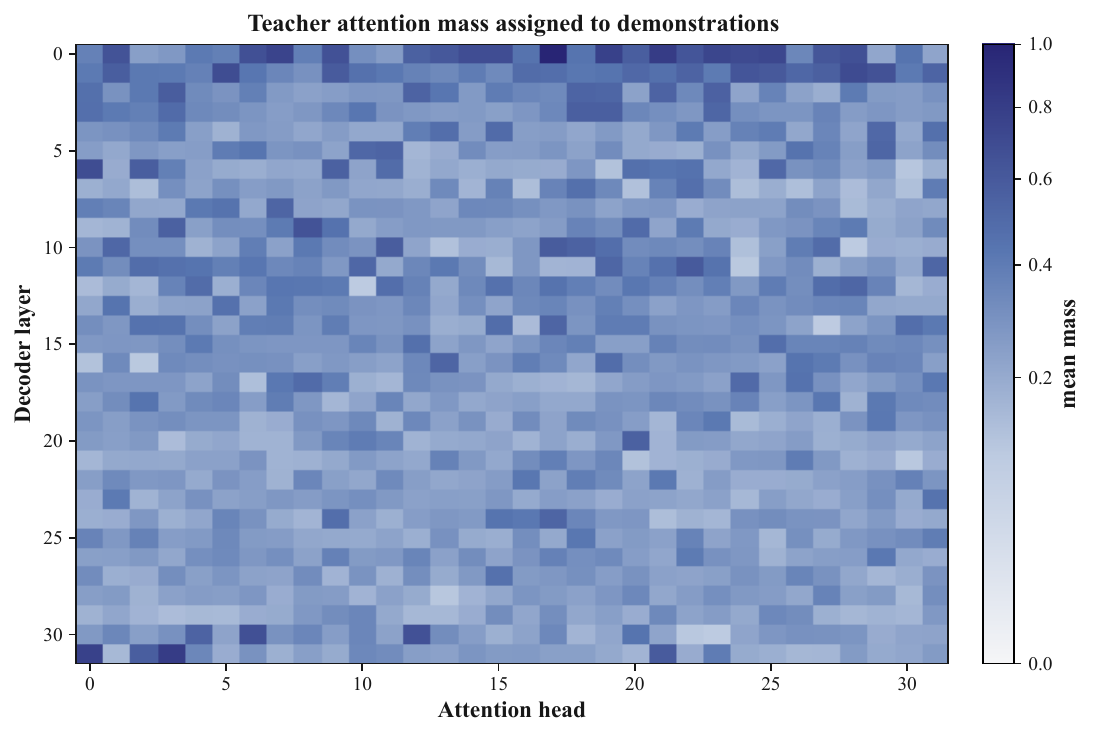}
    \end{minipage}
    \vspace{-3pt}
    \caption{Term coupling and missing attention mass. Left: pairwise relationships among the four decomposition terms. Right: attention mass assigned to demonstration tokens.}
    \label{fig:decomposition-coupling-mass}
\end{figure}
\endgroup

\subsection{Layer-Wise and Question-level Recovery Analysis}\label{app:layer-wise-recovery}

\begingroup
\setlength{\abovedisplayskip}{2pt}
\setlength{\belowdisplayskip}{2pt}
\setlength{\abovedisplayshortskip}{1pt}
\setlength{\belowdisplayshortskip}{1pt}
To assess whether joint calibration improves recovery across model depth and individual questions, we compare attention-only calibration, value-only calibration, and joint calibration against the teacher output shift.
For each variant, we subtract the output of the frozen query-only model, denoted by $O_{\mathrm B}^{(\ell)}$, to obtain $\Delta O_X^{(\ell)}=O_X^{(\ell)}-O_{\mathrm B}^{(\ell)}$.
We measure recovery at layer $\ell$ using the cosine similarity between the vectorized student and teacher output residuals:
\begin{equation}
\label{eq:app-layerwise-recovery}
R_X^{(\ell)}=\cos\!\left(
\operatorname{vec}(\Delta O_X^{(\ell)}),
\operatorname{vec}(\Delta O_{\mathrm T}^{(\ell)})
\right),
\qquad X\in\{\mathrm A,\mathrm V,\mathrm{AV}\}.
\end{equation}
Higher values indicate closer directional alignment with the teacher output shift. We also average recovery across layers for each question and compare the resulting distributions.
The left panel of Figure~\ref{fig:av-recovery-curve} shows that recovery rises after the initial layers, remains relatively high through the middle layers, and generally weakens in later layers.
The relative performance of A-only and V-only varies with depth, while joint calibration achieves higher recovery at many early and middle layers without dominating every layer.
In the right panel, joint calibration shifts the distribution toward higher recovery, achieving a mean of $0.2000$, compared with $0.1810$ for A-only and $0.1756$ for V-only.
Together, these results support the complementary roles of attention routing and value content that their joint calibration improves overall alignment with the teacher output shift.
\endgroup

\begingroup
\setlength{\abovecaptionskip}{1pt}
\setlength{\belowcaptionskip}{0pt}
\begin{figure}[H]
    \centering
        \centering
        \includegraphics[width=0.9\linewidth]{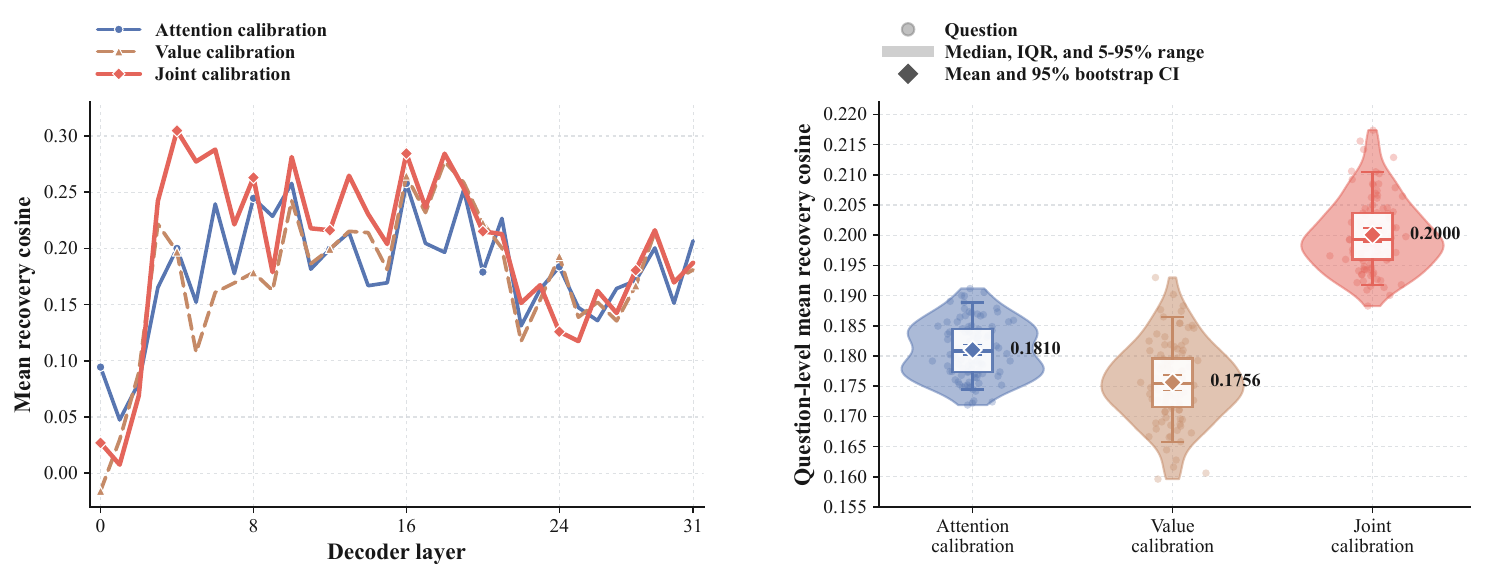}
    \vspace{-4pt}
    \caption{Recovery of the teacher induced output shift by attention only, value only, and joint attention and value calibration. Left: layer wise recovery trajectories. Right: distributions of question level mean recovery. Joint calibration provides the strongest and most stable recovery.}
    \label{fig:av-recovery-curve}
\end{figure}
\endgroup

\subsection{Additional Gating Ablations}\label{app:query-adaptive-gating-ablation}

To assess the contributions of query-adaptive gating and missing-mass correction, we ablate the gating components in Eq.~\ref{eq:token-conditioned-gates} and the mass estimator in Eq.~\ref{eq:calibrated-attention-output}.
We compare the full design with variants that separately remove the $\tanh$ scaling, sigmoid gating, channel gating, or missing-mass estimation, evaluating accuracy on OK-VQA and CIDEr on COCO Caption.
As shown in Table~\ref{tab:query-adaptive-gating-ablation}, the full design achieves the best results on both benchmarks, while removing the $\tanh$ scaling causes the largest performance drop.
Removing the sigmoid or channel gating also lowers both scores, indicating that query-dependent correction strength and low-rank direction selection are beneficial.
Without missing-mass estimation, OK-VQA accuracy decreases from 62.49 to 61.86 and COCO CIDEr decreases from 1.3757 to 1.3705, supporting the utility of the learned query-side scaling factor.

\begin{table}[H]
    \centering
    \caption{Ablation of query-adaptive gating components and missing-mass estimation. Bold numbers denote the best scores.}
    \label{tab:query-adaptive-gating-ablation}
    \small
    \setlength{\tabcolsep}{10pt}
    \renewcommand{\arraystretch}{1.08}
    \begin{tabular}{@{}lcc@{}}
        \toprule
        Variant & OK-VQA & COCO CIDEr \\
        \midrule
        ICL-R &  \textbf{62.49} & \textbf{1.3757} \\
        w/o $\tanh$ scaling  & 59.94  & 1.3589  \\
        w/o sigmoid gating &  61.72  & 1.3671  \\
        w/o channel gating & 61.86 & 1.3709 \\
        w/o missing mass & 61.86 & 1.3705 \\
        \bottomrule
    \end{tabular}
\end{table}

\subsection{Geometric Analysis of RST}
\subsubsection{Complementarity of Point and Fiber Supervision}
We investigate whether the point and fiber terms provide complementary supervision for operator alignment. 
Using captured teacher states from VQAv2 and OKVQA as anchors, we construct three controlled settings that preserve the endpoint, preserve the base point, or preserve the transition under parallel transport. 
We compare the responses of endpoint matching, the individual point and fiber terms, their RST sum, and normalized state pair matching. 
As shown in Figure~\ref{fig:rst-endpoint-ambiguity}, endpoint matching remains insensitive to changes in the base and transition when the endpoint is fixed. 
Likewise, the point term cannot detect transition changes at a fixed base, whereas the fiber term remains near zero when the base moves but the transported transition remains equivalent. 
RST responds to the discrepancies left undetected by either component alone. 
These results support the complementary roles of point and fiber supervision: the former aligns the state from which an operator acts, while the latter aligns the update performed from that state.

\begingroup
\setlength{\abovecaptionskip}{2pt}
\setlength{\belowcaptionskip}{0pt}
\begin{figure}[H]
    \centering
    \includegraphics[width=0.9\linewidth]{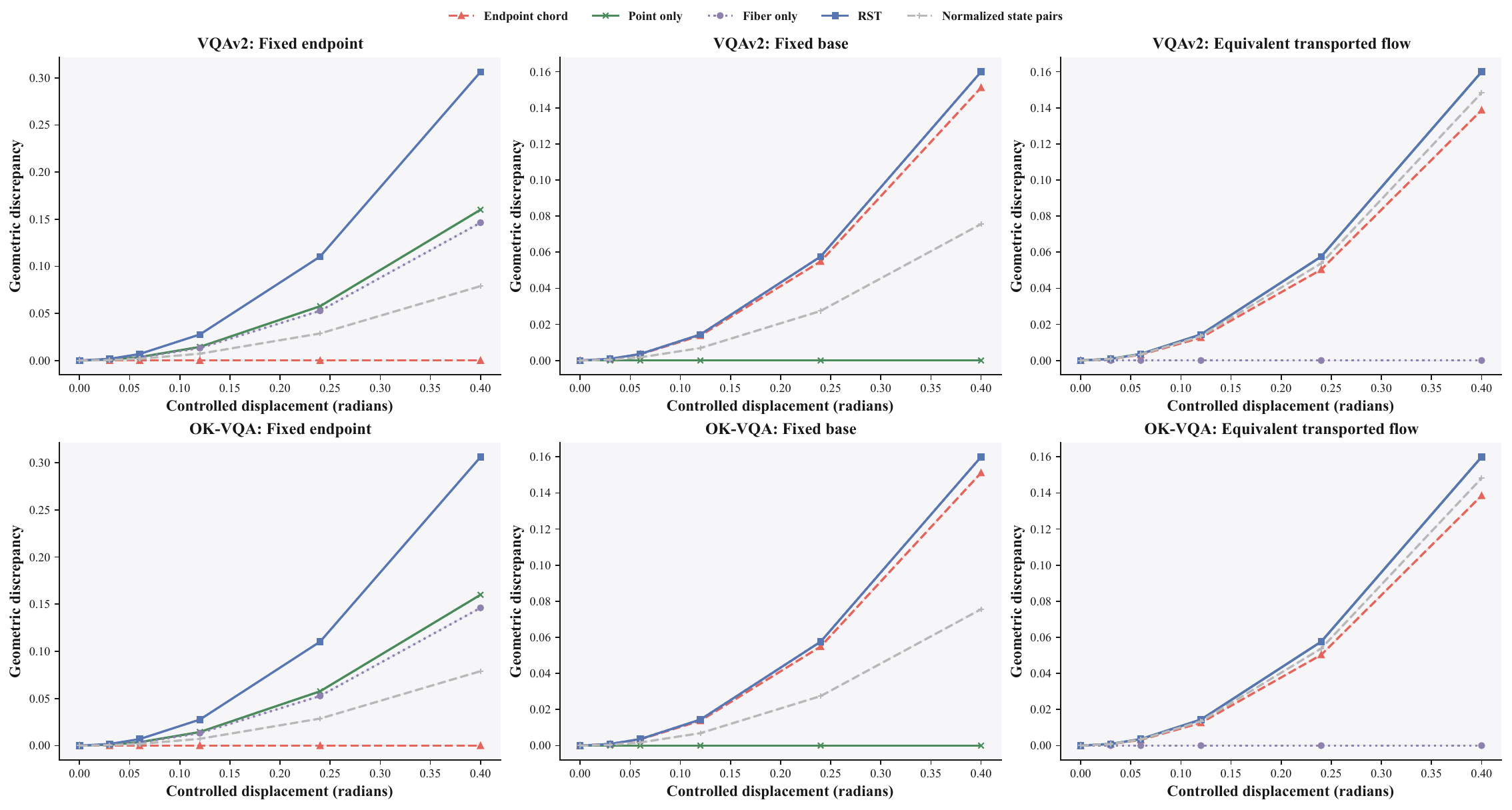}
    \caption{Responses of geometric objectives to controlled displacements anchored at teacher states from VQAv2 (top) and OK-VQA (bottom). Columns show a fixed endpoint, a fixed base point, and an equivalent transition under parallel transport. Curves compare endpoint matching, point and fiber supervision, their RST sum, and normalized state pair matching.}
    \label{fig:rst-endpoint-ambiguity}
\end{figure}
\endgroup

\subsubsection{Geometric Alignment on Unseen Queries}
We next examine whether the geometry encouraged by RST is reflected in representations of unseen queries. 
We compare L2 and RST on 128 distinct images from each of VQAv2, OKVQA, MME and SEED using identical query inputs and shared teacher states. 
We measure point and transported fiber distances across all decoder layers and eight selected query prompt positions, including boundary and task markers. 
Figure~\ref{fig:rst-paired-geometry} shows that RST shifts both distance distributions toward lower discrepancy, with improvements in their central regions and upper tails. 
Figure~\ref{fig:rst-token-layer-geometry} localizes these improvements across layers and query positions, revealing broadly improved alignment alongside some localized regressions. 
Together, these observations provide evidence that the RST achieves better geometric fidelity on unseen queries in both their states and transitions.

\begingroup
\setlength{\abovecaptionskip}{2pt}
\setlength{\belowcaptionskip}{0pt}
\begin{figure}[H]
    \centering
    \includegraphics[width=0.9\linewidth]{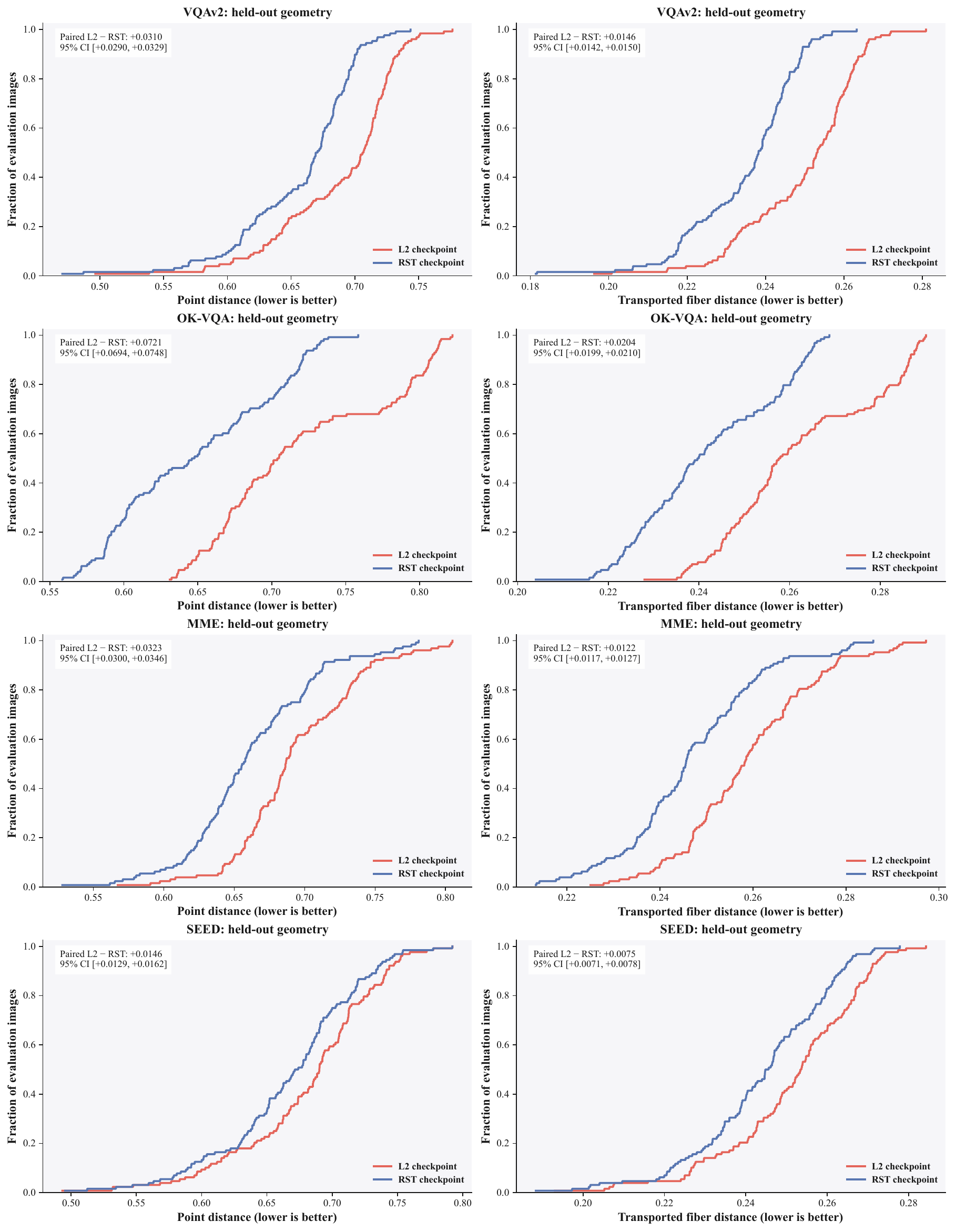}
    \caption{Cumulative distributions of point distance (left) and transported fiber distance (right) for L2 and RST checkpoints on unseen VQAv2, OK-VQA, MME and SEED images. Annotations report paired mean differences (L2 minus RST) and 95\% confidence intervals over images. Lower distances indicate closer geometric alignment.}
    \label{fig:rst-paired-geometry}
\end{figure}
\endgroup

\begingroup
\setlength{\abovecaptionskip}{2pt}
\setlength{\belowcaptionskip}{0pt}
\begin{figure}[H]
    \centering
    \includegraphics[width=\linewidth]{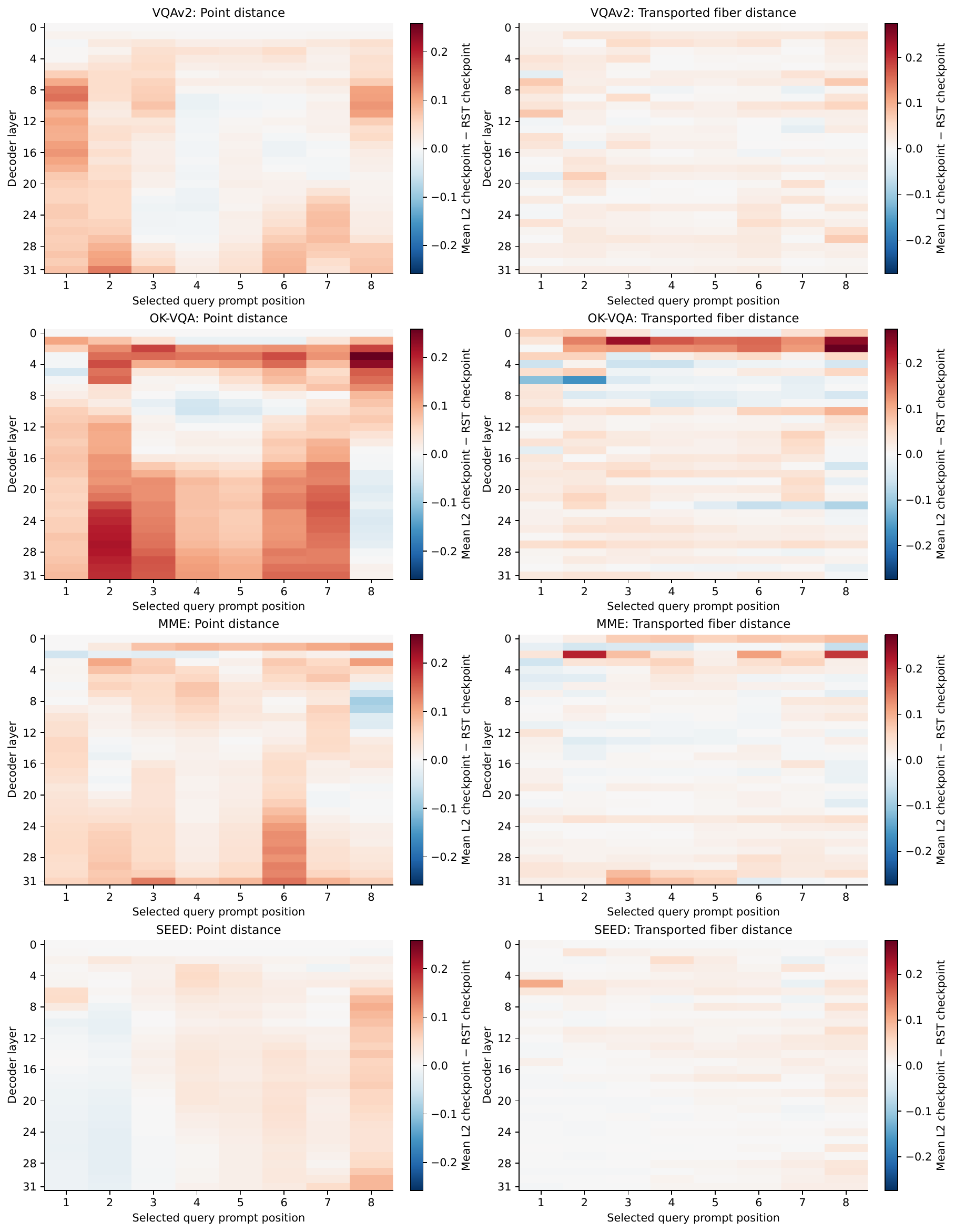}
    \caption{Mean geometric distance differences (L2 minus RST) across decoder layers and eight selected query prompt positions on VQAv2, OK-VQA, MME and SEED. The left and right columns show point and transported fiber distances, respectively. Positive values indicate closer alignment under RST, while negative values indicate localized regressions.}
    \label{fig:rst-token-layer-geometry}
\end{figure}
\endgroup

\subsubsection{Geometric Consistency of Parallel Transport}
To examine whether parallel transport prevents spurious discrepancies between geometrically equivalent transitions, we conduct controlled tests using states collected from existing Idefics2 checkpoints on 128 evaluation images from each of VQAv2 and OK-VQA. At four decoder depths, we use the final query prompt token and move the teacher base point along either the observed direction toward the student base or the teacher transition direction. An independent plane rotation constructs the corresponding equivalent transition at each target point. We measure the discrepancy relative to this reference, normalized by the teacher transition norm, and additionally evaluate tangency residuals, norm preservation, and agreement with the rotation reference. Figure~\ref{fig:parallel-transport-consistency} shows that direct subtraction introduces discrepancies that increase with base displacement and vary substantially across directions. Parallel transport remains close to the reference in both settings, while the numerical consistency checks exhibit small residuals. These results support parallel transport as a geometrically consistent component of the RST fiber term, allowing equivalent transitions at different base points to be compared without introducing an additional penalty solely from their base displacement.

\begingroup
\setlength{\abovecaptionskip}{2pt}
\setlength{\belowcaptionskip}{0pt}
\begin{figure}[H]
    \centering
    \includegraphics[width=\linewidth]{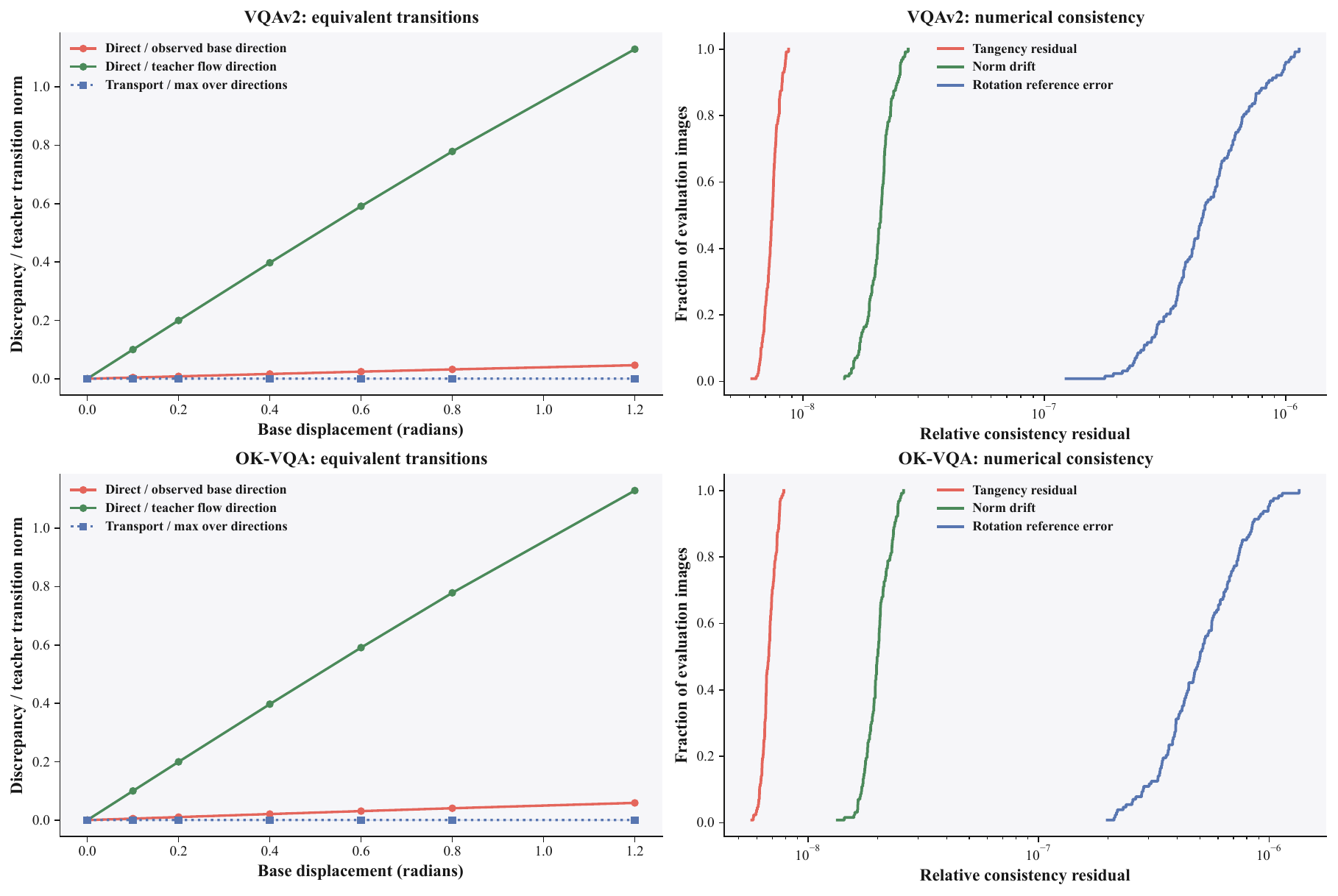}
    \caption{Geometric consistency of parallel transport on VQAv2 (top) and OK-VQA (bottom). Left: normalized discrepancies between equivalent transitions as the base point moves along the observed base direction or the teacher transition direction; the transport curve reports the maximum over both directions. Right: cumulative distributions of relative tangency residuals, norm drift, and errors against the independent rotation reference.}
    \label{fig:parallel-transport-consistency}
\end{figure}
\endgroup

To determine whether this geometric correction preserves sensitivity to actual transition differences, we further evaluate the recovery of controlled perturbations. Using the same datasets, images, and four decoder depths, we displace the teacher base along the observed direction toward the student base and add a perturbation to the equivalent rotated transition. The perturbation is orthogonal to the plane of base displacement, and its magnitude is specified relative to the teacher transition norm. We vary both base displacement and perturbation strength, then measure the absolute deviation between the estimated normalized squared discrepancy and its known value. Figure~\ref{fig:parallel-transport-perturbation-recovery} shows that direct subtraction incurs an additional error that grows with base displacement, whereas parallel transport accurately recovers the injected discrepancy throughout the tested grid. The near-zero values in the transport panels therefore indicate accurate measurement, including under nonzero perturbations. Together, these results show that the RST fiber comparison can exclude discrepancies caused by changing base points while retaining sensitivity to genuine differences between transitions.

\begingroup
\setlength{\abovecaptionskip}{2pt}
\setlength{\belowcaptionskip}{0pt}
\begin{figure}[H]
    \centering
    \includegraphics[width=\linewidth]{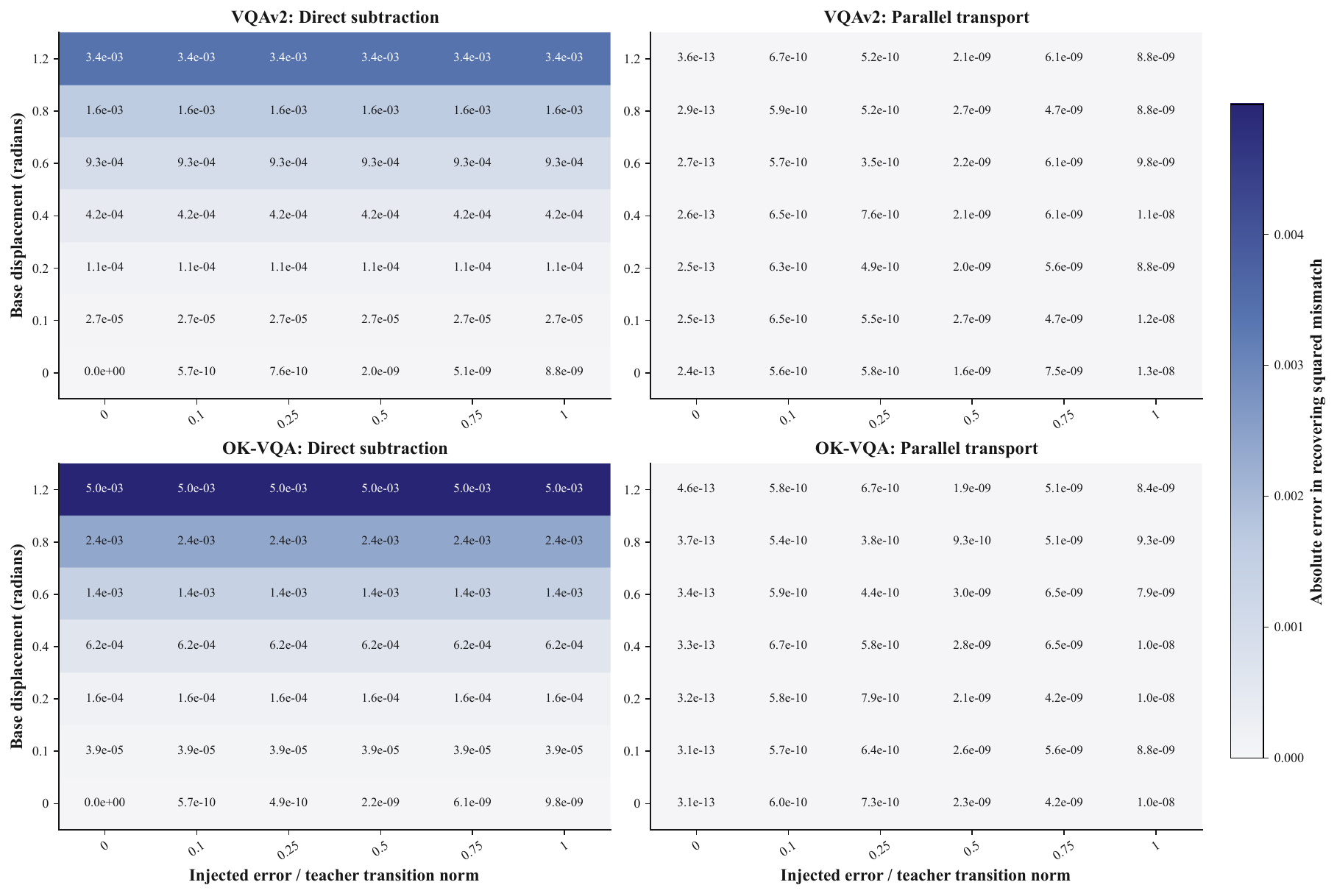}
    \caption{Recovery of controlled transition perturbations on VQAv2 (top) and OK-VQA (bottom), comparing direct subtraction (left) with parallel transport (right). Heatmaps report the absolute error in recovering the known normalized squared discrepancy across base displacements and perturbation strengths. Near-zero errors indicate accurate recovery of the injected discrepancy, not the absence of a transition difference.}
    \label{fig:parallel-transport-perturbation-recovery}
\end{figure}
\endgroup

\subsection{Generalization to Other MLLMs}\label{app:other-mllms}

To examine the applicability of ICL-R beyond the backbones used in the main experiments, we extend our evaluation to Idefics1-9B and LLaVA-v1.6(Mistral-7B).
The evaluation covers VQAv2 and OK-VQA for visual question answering and COCO for image captioning.
We compare ICL-R with zero-shot inference, explicit ICL, MimIC, and HiFICL, as summarized in Table~\ref{tab:generalization-other-mllms}.
For both backbones, all trainable methods use 8 demonstrations per training episode.
On Idefics1-9B, ICL-R achieves the best performance on all three benchmarks, outperforming both explicit ICL and the implicit ICL baselines.
On LLaVA-v1.6, it leads on VQAv2 and COCO and remains competitive on OK-VQA, trailing MimIC and HiFICL by only 0.42 and 0.25 points, respectively.

\begin{table}[H]
    \centering
    \setlength{\belowcaptionskip}{6pt}
    \caption{Generalization results on additional MLLM backbones across visual question answering and image captioning benchmarks. Bold numbers denote the best result.}
    \label{tab:generalization-other-mllms}
    \small
    \setlength{\tabcolsep}{10pt}
    \renewcommand{\arraystretch}{1.08}
    \begin{tabular}{@{}llccc@{}}
        \toprule
        Model & Method & VQAv2 & OK-VQA & COCO \\
        \midrule
        \multirow{5}{*}{Idefics1-9B}
        & Zero-shot   & 42.77             & 29.08             & 0.6111 \\
        & 8-shot ICL & 54.75             & 47.30             & 0.9866 \\
        & MimIC       & 57.85 & 49.94   & 1.0895 \\
        & HiFICL      & 53.20    & 49.20 & 0.8283 \\
        & ICL-R       & \textbf{57.91}    &  \textbf{52.84}   & \textbf{1.1761}\\
        \midrule
        \multirow{5}{*}{LLaVA-v1.6(Mistral-7B)}
        & Zero-shot  & 70.00             & 26.21             & 0.7157 \\
        & 8-shot ICL & 68.00             & 20.64             & 0.6678 \\
        & MimIC      & 71.24 & \textbf{63.34} & 1.2857 \\
        & HiFICL     & 74.71    & 63.17    & 1.3192 \\
        & ICL-R      & \textbf{75.36}    &   62.92                & \textbf{1.3203}\\ 
        \bottomrule
    \end{tabular}
\end{table}

\subsection{Dataset Transfer}\label{app:dataset-transfer}

To assess whether the demonstration effects learned by ICL-R generalize beyond the training dataset, we evaluate transfer between benchmarks within the same task family on Idefics2-8B-Base.
We consider two settings: training the calibration modules on COCO Caption and evaluating on Flickr30K for image captioning, and training on VQAv2 and evaluating on OK-VQA for visual question answering.
The learned modules are applied to the target benchmark without further training, and we compare against zero-shot inference and 8-shot ICL using CIDEr for captioning and answer accuracy for question answering.
For 8-shot ICL, the demonstrations are sampled from the source benchmark.
As shown in Table~\ref{tab:dataset-transfer}, ICL-R outperforms both baselines in both transfer settings, indicating that its gains extend to benchmarks not used to train the calibration modules.
These results suggest that joint attention and value calibration captures demonstration effects that can transfer across related datasets, supporting its applicability to new benchmarks within the same task family.

\begin{table}[H]
    \centering
    \caption{Cross-dataset transfer performance. Train and Test indicate the source and target datasets, respectively. Bold numbers denote the best scores.}
    \label{tab:dataset-transfer}
    \small
    \setlength{\tabcolsep}{10pt}
    \renewcommand{\arraystretch}{1.08}
    \begin{tabular}{@{}llccc@{}}
        \toprule
        Train & Test & Zero-shot & 8-shot ICL & ICL-R \\
        \midrule
        COCO & Flickr30K & 0.5327 & 0.8364 & \textbf{0.8802} \\
        VQAv2 & OK-VQA & 43.08 & 51.43 & \textbf{54.49} \\
        \bottomrule
    \end{tabular}
\end{table}

\subsection{Trainable Parameter Analysis}\label{app:parameter-analysis}

To assess the parameter cost of learning demonstration effects, we compare the number of trainable parameters across methods on Idefics2 and LLaVA-Interleave.
Table~\ref{tab:trainable-parameters} reports parameter counts in millions and their ratios relative to ICL-R; zero-shot inference and explicit 8-shot ICL require no trainable parameters.
ICL-R uses 4.5M trainable parameters on both backbones, compared with 17.6M and 19.7M for LoRA, respectively.
Its parameter count is also lower than that of MimIC on LLaVA-Interleave, but higher than those of LIVE and HiFICL on both backbones and MimIC on Idefics2.
The larger parameter count of MimIC on LLaVA-Interleave stems from its dense query-dependent linear mapping $f(q)$, whose parameter requirements grow with the hidden dimension, bringing the total to 17.0M on this backbone.
Thus, ICL-R does not minimize parameter count among all implicit ICL methods. Instead, its joint attention and value calibration provides the strong task performance reported in Table~\ref{tab:main-results} with substantially fewer trainable parameters than LoRA and without updating the backbone.

\begin{table}[H]
    \centering
    \caption{Trainable parameters in millions (M). Parentheses give the ratio to ICL-R on the same backbone; a dash denotes no trainable parameters.}
    \label{tab:trainable-parameters}
    \scriptsize
    \setlength{\tabcolsep}{2pt}
    \renewcommand{\arraystretch}{1.15}
    \begin{tabular*}{\linewidth}{@{\extracolsep{\fill}}lccccccc@{}}
        \toprule
        Model & Zero-shot & 8-shot ICL & LoRA & LIVE & MimIC & HiFICL & \textbf{ICL-R} \\
        \midrule
        Idefics2 & -- & -- & 17.6 ($\times$3.91) & 0.13 ($\times$0.03) & 0.26 ($\times$0.06) & 2.2 ($\times$0.49) & 4.5 ($\times$1.00) \\
        LLaVA-Interleave & -- & -- & 19.7 ($\times$4.38) & 0.13 ($\times$0.03) & 17.0 ($\times$3.78) & 2.2 ($\times$0.49) & 4.5 ($\times$1.00) \\
        \bottomrule
    \end{tabular*}
\end{table}

\subsection{Single-Pass Fused SDPA Implementation}\label{app:fused-sdpa}

We express the joint attention and value calibration in Eq.~\ref{eq:joint-attention-value-calibration} using a single scaled dot product attention (SDPA) call followed by value decoding and mass scaling. The construction does not require explicitly forming the query-dependent value tensor, and a compatible memory-efficient SDPA backend also avoids materializing the full attention score matrix.
We first derive the equivalence with attention dropout disabled. Batch and head indices are omitted in the per-token equations, and $Q_{0,t}$, $K_{0,s}$, and $V_{0,s}$ denote row vectors. The query used for gating in Section~\ref{sec:joint-calibration} is the corresponding column vector $q_t=Q_{0,t}^{\top}$; the channel gates $\rho_t^A$ and $\rho_t^V$ are represented as row vectors below.
Let $s_0=d_h^{-1/2}$ be the original attention scale, and define
\begin{equation}
\label{eq:fused-augmented-query}
\widehat Q_t^A
=\frac{g_t^A}{s_0}
\left[\rho_t^A\odot\left(Q_{0,t}U_Q\right)\right].
\end{equation}
We concatenate the base and calibration features as
\begin{equation}
\label{eq:fused-sdpa-inputs}
Q^+=Q_0\mathbin\Vert\widehat Q^A,
\qquad
K^+=K_0\mathbin\Vert(K_0U_K),
\qquad
V^+=V_0\mathbin\Vert Z,
\quad Z=V_0P_V,
\end{equation}
where $\Vert$ denotes concatenation along the feature dimension within each head. For batch size $N$, $H$ heads, and query and key lengths $T_q$ and $T_k$, the augmented tensors satisfy
\begin{equation}
\label{eq:fused-sdpa-shapes}
\begin{aligned}
Q^+&\in\mathbb{R}^{N\times H\times T_q\times(d_h+r_A)},\\
K^+&\in\mathbb{R}^{N\times H\times T_k\times(d_h+r_A)},\\
V^+&\in\mathbb{R}^{N\times H\times T_k\times(d_h+r_V)}.
\end{aligned}
\end{equation}
The SDPA call explicitly uses the original scale $s_0$, not the default scale $(d_h+r_A)^{-1/2}$ associated with the augmented query and key dimension. No additional $r_A^{-1/2}$ factor is introduced, consistent with Eq.~\ref{eq:joint-attention-value-calibration}. The augmented score then satisfies
\begin{equation}
\label{eq:fused-score-equivalence}
\begin{aligned}
\left[s_0Q^+(K^+)^\top\right]_{t,s}
&=s_0Q_{0,t}K_{0,s}^{\top}
+s_0\widehat Q_t^A(K_{0,s}U_K)^\top\\
&=s_0Q_{0,t}K_{0,s}^{\top}
+g_t^A
\left[\rho_t^A\odot\left(Q_{0,t}U_Q\right)\right]
\left(K_{0,s}U_K\right)^\top\\
&=S_{0,t,s}+\Delta S^A_{t,s},
\end{aligned}
\end{equation}
which is exactly the calibrated pre-mask score in Eq.~\ref{eq:joint-attention-value-calibration}. We therefore compute
\begin{equation}
\label{eq:single-fused-sdpa}
Y^+=\operatorname{SDPA}(Q^+,K^+,V^+;M,s_0)
=O^P\mathbin\Vert Z^P,
\end{equation}
where $M$ is the additive form of the original causal and padding mask, with zero at allowed positions and $-\infty$ at masked positions. Each evaluated query row is assumed to have at least one allowed key. The attention matrix implicitly produced by this call is $\widetilde A=\operatorname{softmax}(S_0+\Delta S^A+M)$, and the two output blocks satisfy
\begin{equation}
\label{eq:fused-value-aggregation}
O_t^P=\sum_s\widetilde A_{t,s}V_{0,s},
\qquad
Z_t^P=\sum_s\widetilde A_{t,s}(V_{0,s}P_V).
\end{equation}
To match the mass-aware weights in Eq.~\ref{eq:calibrated-attention-output}, define
\begin{equation}
\label{eq:fused-mass-definition}
m_t=s_m\,\sigma\!\left(\frac{\bar q_t^{\top}w_m}{\sqrt{d_h}}+b_m\right),
\qquad
\widehat A_{t,s}=(1-m_t)\widetilde A_{t,s},
\end{equation}
where $\bar q_t=\operatorname{LN}(q_t)$ and the same attention mask is understood in the method formulation.
The value correction can be decoded after attention because its query-dependent gates are constant with respect to the key index:
\begin{equation}
\label{eq:fused-output-decoding}
\begin{aligned}
O_t^N
&=O_t^P+g_t^V\left[\left(\rho_t^V\odot Z_t^P\right)R_V\right]\\
&=\sum_s\widetilde A_{t,s}
\left(V_{0,s}+g_t^V
\left[\left(\rho_t^V\odot V_{0,s}P_V\right)R_V\right]\right),\\
\widetilde O_t&=(1-m_t)O_t^N.
\end{aligned}
\end{equation}
Here, $O_t^N=\sum_s\widetilde A_{t,s}\widetilde V_{t,s}$ denotes the normalized aggregation before applying the total-mass factor. Consequently, $\widetilde O_t=\sum_s\widehat A_{t,s}\widetilde V_{t,s}$, establishing algebraic equivalence to Eq.~\ref{eq:calibrated-attention-output} under the same mask and with attention dropout disabled.
If attention dropout is enabled, the same rearrangement holds for a fixed dropout realization when both formulations use identical dropped attention weights; these weights no longer necessarily sum to one.
The augmented feature construction, value decoding, and mass scaling occur outside the SDPA call. Softmax, masking, optional dropout, and value aggregation are handled by SDPA. A memory-efficient fused backend can be used when the augmented dimensions, data type, and hardware are supported; a single SDPA call alone does not guarantee fused execution.

\subsection{Peak GPU memory usage}

\begin{figure}[H]
    \centering
    \IfFileExists{figures/gpumemory.pdf}{
        \includegraphics[width=0.92\linewidth]{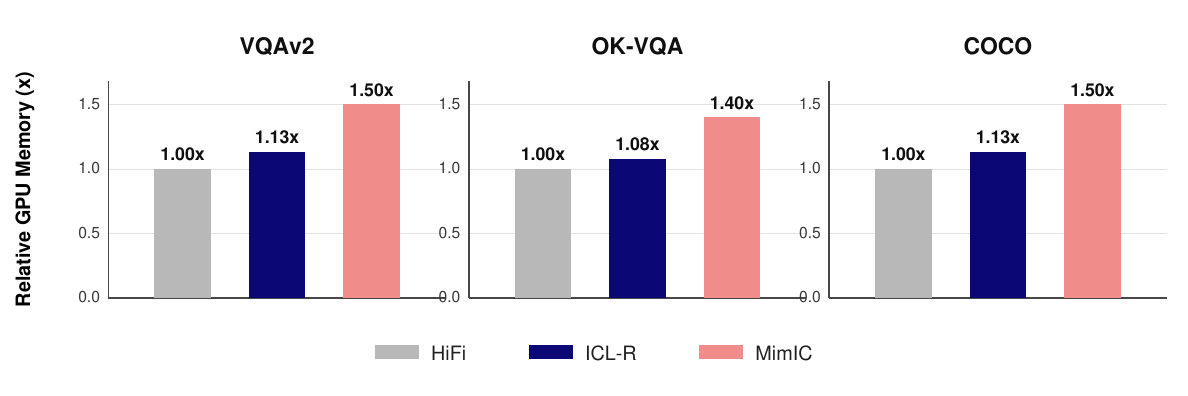}
    }{
        \fbox{\parbox[c][0.20\textheight][c]{0.88\linewidth}{
            \centering
            Placeholder for the peak GPU memory comparison between ICL-R and HiFICL on VQAv2, OK-VQA, and COCO Caption.
        }}
    }
    \caption{Relative peak GPU memory usage of HiFICL, ICL-R, and MimIC on VQAv2, OK-VQA, and COCO Caption, normalized to HiFICL on each benchmark.}
    \label{fig:gpu-memory-comparison}
\end{figure}

To assess the memory overhead of joint attention and value calibration, we compare the peak GPU memory usage of ICL-R, HiFICL, and MimIC on VQAv2, OK-VQA, and COCO Caption.
Figure~\ref{fig:gpu-memory-comparison} reports relative memory usage, normalized to HiFICL on each benchmark.
ICL-R requires 1.13$\times$, 1.08$\times$, and 1.13$\times$ the memory of HiFICL, respectively, whereas MimIC requires 1.50$\times$, 1.40$\times$, and 1.50$\times$.
Thus, ICL-R incurs an 8\%--13\% overhead over HiFICL while using less memory than MimIC on all three benchmarks.
The fused SDPA implementation described above avoids explicitly materializing calibrated attention logits and query-dependent value tensors, helping limit the intermediate storage required by joint calibration.
Overall, these results indicate that joint attention and value calibration can be implemented with a modest memory overhead relative to HiFICL, although ICL-R does not achieve the lowest peak memory usage among the compared methods.

\subsection{Error Correction Across Categories and Backbones}\label{app:error-correction}
To identify which capabilities benefit from ICL-R and how individual prediction changes contribute to its accuracy gains, we analyze paired predictions on MME and SEED-Bench with Idefics2 and LLaVA-Interleave.
For each backbone, we compare ICL-R with zero-shot inference, 8-shot ICL, LoRA, LIVE, MimIC, and HiFICL on the same evaluation examples using a common answer parser.
We examine category accuracy differences, the balance between corrected and newly introduced errors, and the uncertainty of the resulting overall gains.

\noindent\textbf{Category accuracy differences.}
To determine whether the improvements span multiple capabilities, we compute the accuracy difference between ICL-R and each baseline in percentage points for all MME categories and SEED-Bench categories.
Figure~\ref{fig:error-correction-categories} reports these differences, with the number of questions shown for each category.
On Idefics2, ICL-R matches or exceeds 8-shot ICL in every evaluated category, improving all categories except MME position, where performance is unchanged.
Gains are broadly distributed, with particularly large improvements on SEED-Bench instance location and counting, reaching 18.2 and 23.7 percentage points over 8-shot ICL, respectively.
This pattern is consistent with the benefits of joint attention and value calibration. Specifically, adjusting attention can help identify relevant objects, while calibrating values can improve the object information aggregated for localization and counting.
However, on SEED-Bench text understanding, ICL-R underperforms all four trainable baselines.
A possible explanation is that value calibration introduces interference into representations used for text recognition, altering fine-grained textual information even when attention selects the relevant image regions.
On LLaVA-Interleave, ICL-R also improves broadly over explicit ICL, with positive gains in every evaluated category, although its advantages over trainable baselines are less uniform.
MME position is a notable weakness. Despite improving over 8-shot ICL, ICL-R falls below zero-shot inference, LoRA, and LIVE.
One possible explanation is that the joint corrections favor salient object content but sometimes disturb the relative spatial information needed to distinguish object positions, limiting their benefit for precise positional judgments.
These observations suggest that joint calibration benefits object selection and aggregation, while potentially introducing interference in text recognition and spatial reasoning.

\begingroup
\setlength{\abovecaptionskip}{2pt}
\setlength{\belowcaptionskip}{0pt}
\begin{figure}[H]
    \centering
    \includegraphics[width=\linewidth]{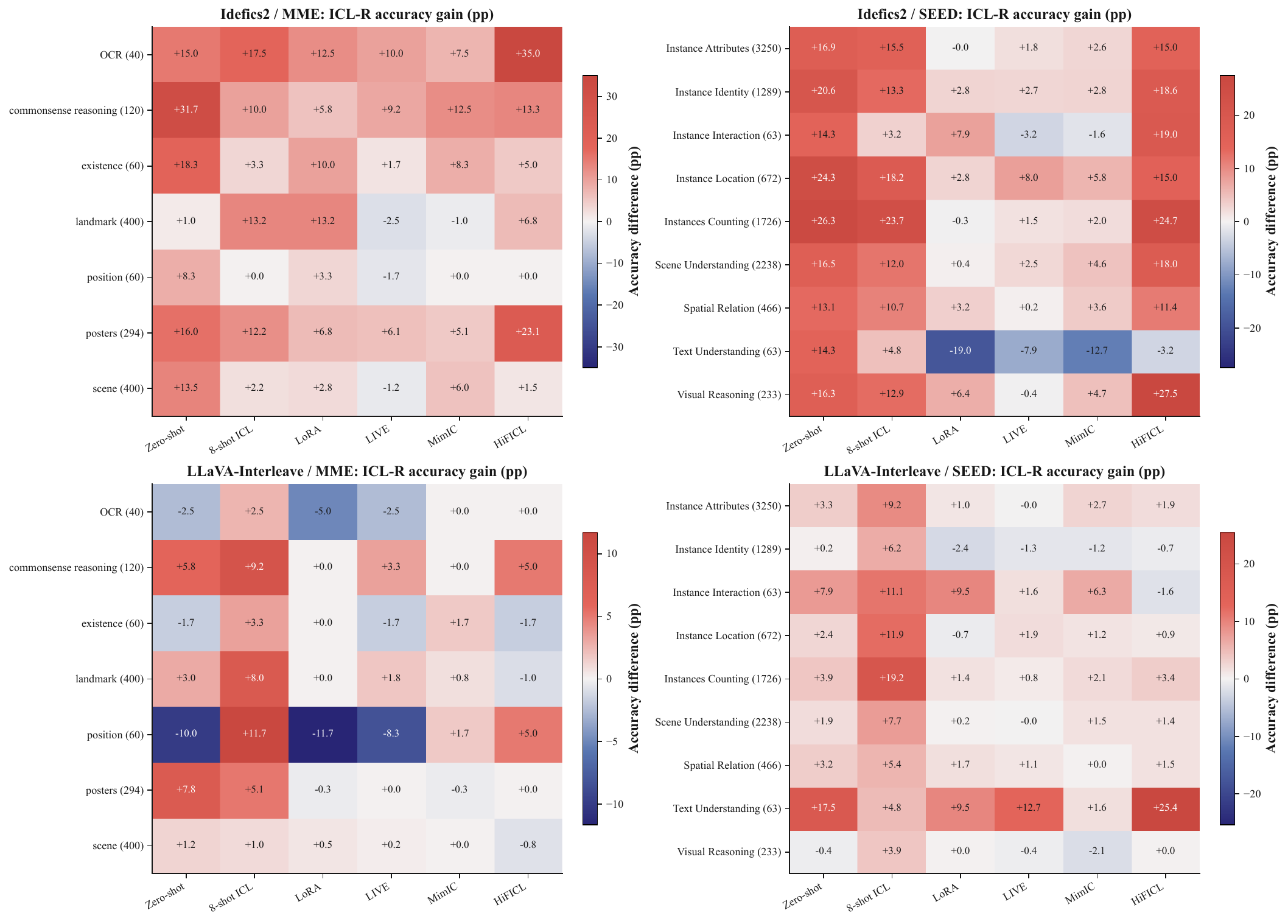}
    \caption{Category accuracy differences between ICL-R and each baseline, measured in percentage points. Rows correspond to Idefics2 (top) and LLaVA-Interleave (bottom); columns correspond to MME (left) and SEED-Bench (right). Category sizes are shown in parentheses. Positive values indicate improvements, while negative values indicate regressions.}
    \label{fig:error-correction-categories}
\end{figure}
\endgroup

\noindent\textbf{Corrected and newly introduced errors.}
To distinguish beneficial prediction changes from regressions, we count questions that each baseline answers incorrectly but ICL-R answers correctly, and questions for which the reverse holds.
Figure~\ref{fig:error-correction-transitions} expresses both counts as shares of all evaluated questions, so their difference equals the overall accuracy gain.
On Idefics2, corrected errors outnumber newly introduced errors for every comparator on both benchmarks.
For example, relative to HiFICL on MME, ICL-R corrects 223 errors while introducing 89 new errors, yielding a net gain of 9.75 percentage points.
On LLaVA-Interleave, corrections clearly exceed new errors relative to 8-shot ICL, whereas the two are more closely balanced relative to LoRA and LIVE.
Thus, the gains over explicit ICL reflect a favorable balance of corrections and regressions on both backbones, while this balance is less consistently favorable against strong trainable baselines on LLaVA-Interleave.

\begingroup
\setlength{\abovecaptionskip}{2pt}
\setlength{\belowcaptionskip}{0pt}
\begin{figure}[h]
    \centering
    \includegraphics[width=0.9\linewidth]{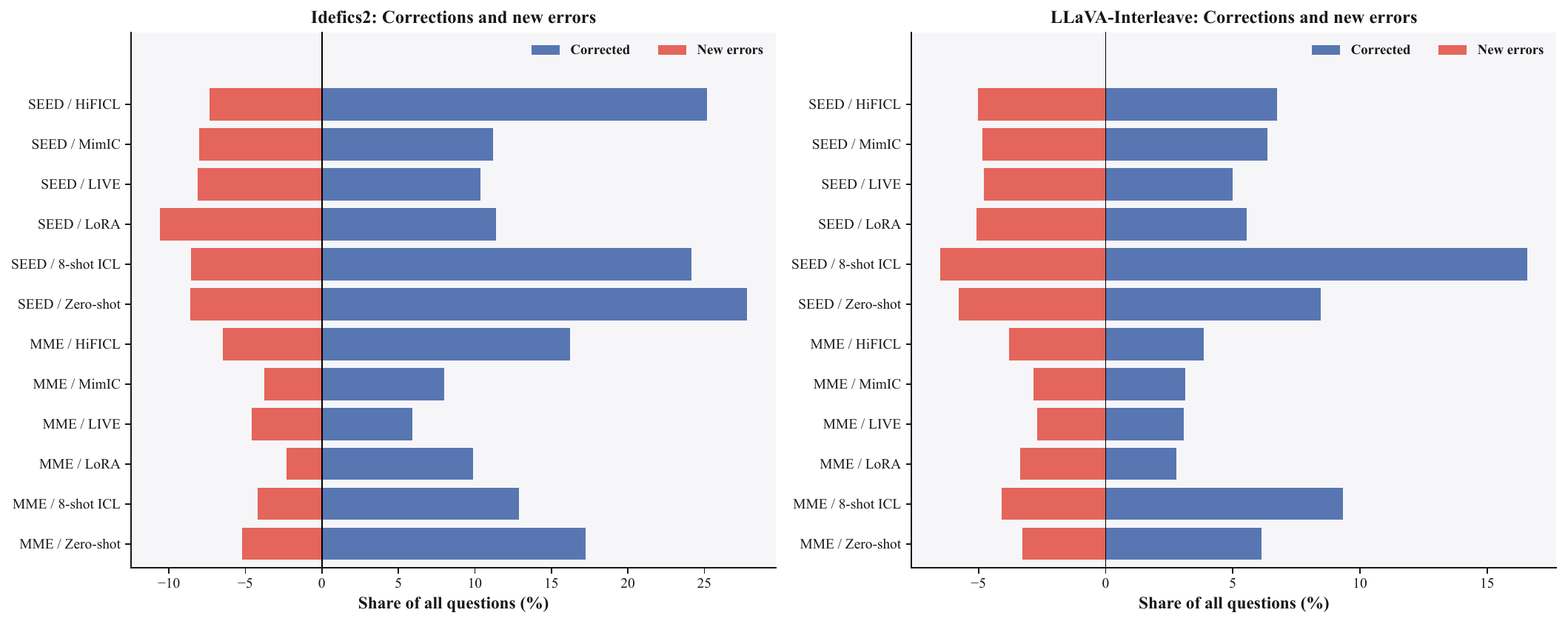}
    \caption{Corrected and newly introduced errors on Idefics2 (left) and LLaVA-Interleave (right). For each benchmark and baseline, bars show the shares of all questions corrected by ICL-R (right of zero) or newly answered incorrectly (left of zero). The difference between their magnitudes gives the net accuracy gain.}
    \label{fig:error-correction-transitions}
\end{figure}
\endgroup

\noindent\textbf{Paired overall gains and uncertainty.}
To assess the uncertainty of the aggregate improvements, we estimate 95\% confidence intervals for paired accuracy differences using 2,000 bootstrap resamples of images, retaining all questions associated with each sampled image.
Figure~\ref{fig:error-correction-paired-gains} reports the net gains and their intervals for both benchmarks and backbones.
Relative to HiFICL, ICL-R gains 9.75 points on MME and 17.83 points on SEED-Bench with Idefics2, and both confidence intervals exclude zero.
On LLaVA-Interleave, the gains over 8-shot ICL are 5.24 and 10.08 points, with confidence intervals of [3.35, 7.13] and [8.85, 10.60], respectively.
It also gains 1.72 points over HiFICL on SEED-Bench, whereas the smaller differences from LoRA and LIVE have confidence intervals that include zero.
These results provide evidence that ICL-R can improve accuracy across the evaluated backbones and benchmarks.

\begingroup
\setlength{\abovecaptionskip}{2pt}
\setlength{\belowcaptionskip}{0pt}
\begin{figure}[h]
    \centering
    \includegraphics[width=0.9\linewidth]{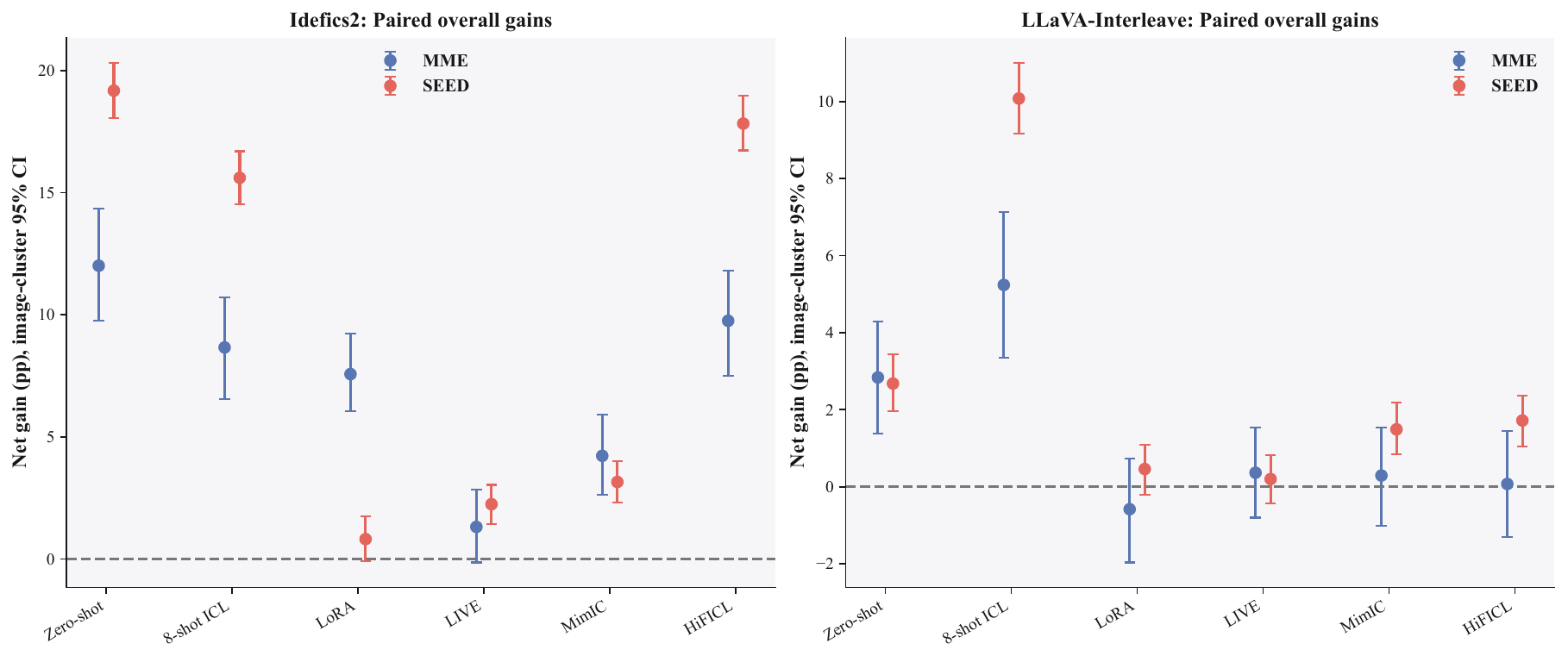}
    \caption{Paired overall accuracy gains of ICL-R over each baseline on Idefics2 (left) and LLaVA-Interleave (right). Markers show net gains in percentage points on MME and SEED-Bench; error bars show 95\% confidence intervals from 2,000 paired bootstrap resamples of images. The dashed line denotes no accuracy difference.}
    \label{fig:error-correction-paired-gains}
\end{figure}
\endgroup

To examine whether the same capabilities benefit across backbones, we pair the accuracy gains for each category and comparator and plot the gains on Idefics2 and LLaVA-Interleave as the two coordinates in Figure~\ref{fig:error-correction-cross-backbone}. 
This comparison includes all seven MME categories and nine SEED-Bench categories, retaining positive, negligible, and negative differences. 
Relative to HiFICL, ICL-R improves MME OCR accuracy by 35.00 percentage points on Idefics2, while leaving it unchanged on LLaVA-Interleave. 
For SEED-Bench text understanding, the corresponding changes are -3.17 and +25.40 points. 
These patterns show that the distribution and magnitude of the gains depend on the backbone. 
Category analysis thus complements aggregate accuracy by revealing both shared improvements and differences in where ICL-R provides the greatest benefit.

\begingroup
\setlength{\abovecaptionskip}{2pt}
\setlength{\belowcaptionskip}{0pt}
\begin{figure}[h]
    \centering
    \includegraphics[width=0.9\linewidth]{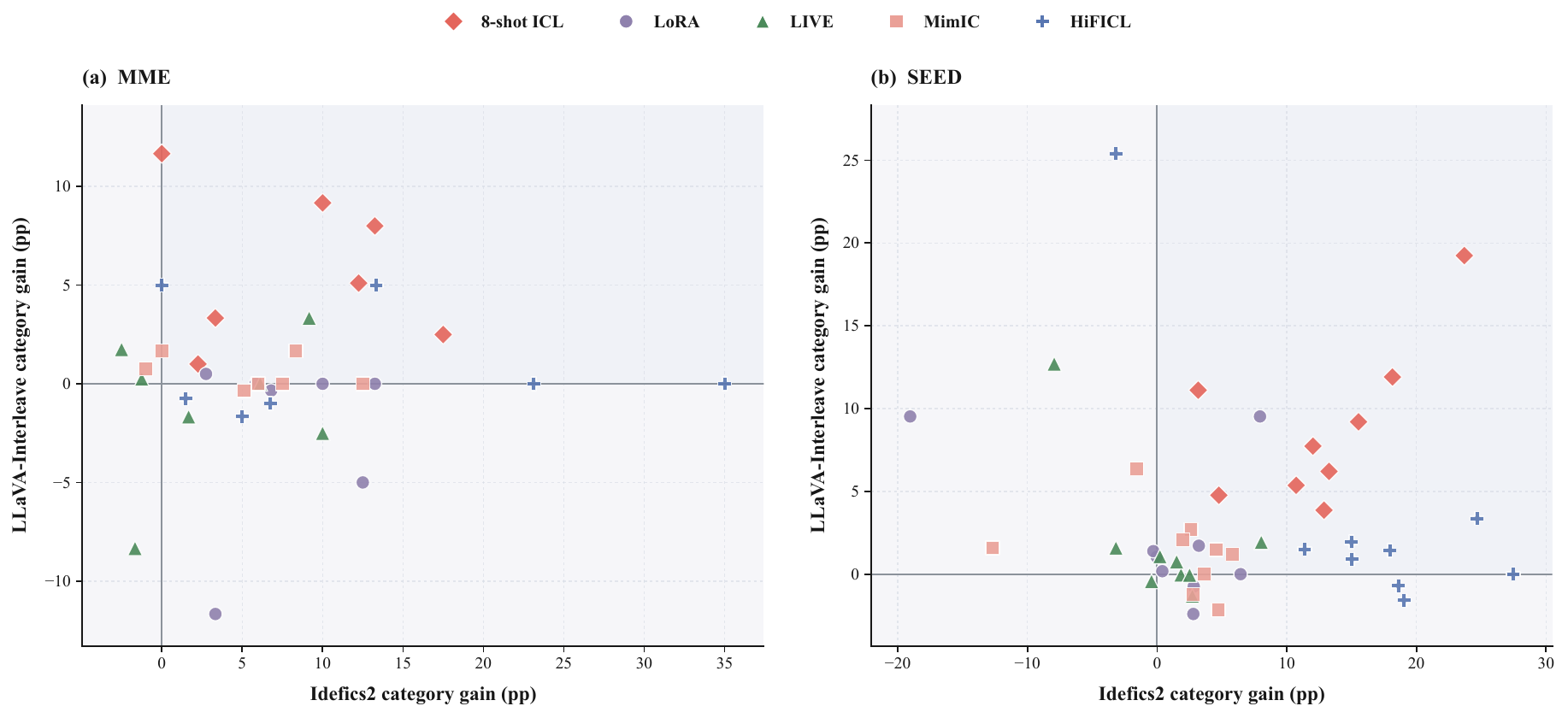}
    \caption{Category accuracy gains across backbones on MME (left) and SEED-Bench (right). Each point represents a category and comparator, with gains on Idefics2 and LLaVA-Interleave shown on the horizontal and vertical axes, respectively. Gains are measured in percentage points.}
    \label{fig:error-correction-cross-backbone}
\end{figure}
\endgroup

\subsection{Paired discrimination on MME}
We complement question-level accuracy with the existing MME Accuracy+ metric,
which requires both questions associated with an image to be answered correctly.
We retain the same 687 evaluation images and all seven methods on both backbones,
including invalid outputs as errors. ICL-R increases paired accuracy from
59.68\% to 74.53\% on Idefics2 and from 60.99\% to 72.05\% on LLaVA-Interleave,
relative to explicit 8-shot ICL. The corresponding paired gains are 14.85
percentage points (95\% CI: [11.21,18.49]) and 11.06 points
(95\% CI: [7.71,14.56]). Confidence intervals use 2,000 bootstrap resamples of
complete images. The within-image error distribution and transition matrices
show how changes in individual answers translate into more images with two
correct judgments. Advantages over trainable baselines are less uniform:
Idefics2's interval relative to LIVE includes zero, as do LLaVA-Interleave's
intervals relative to all four trainable baselines. 
\begin{figure}[H]
\centering
\includegraphics[width=0.9\linewidth]{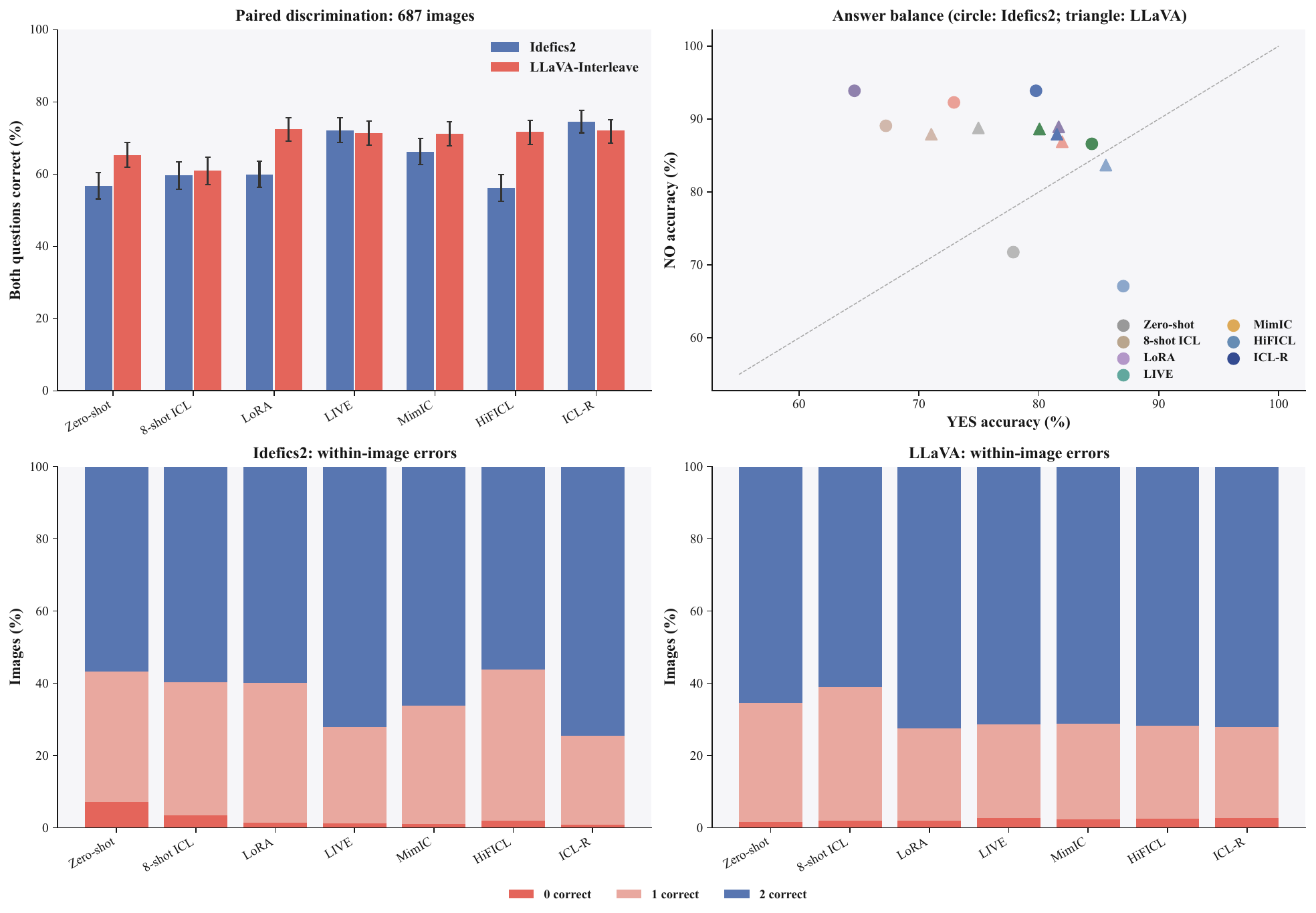}
\caption{Paired discrimination on the fixed MME evaluation subset. The panels
show both-questions-correct accuracy with image-bootstrap 95\% intervals,
YES/NO accuracy, and the fraction of images with zero, one or two correct
answers. All seven methods are retained for both backbones.}
\label{fig:exp5-mme-pairs}
\end{figure}

\begin{figure}[H]
\centering
\includegraphics[width=0.9\linewidth]{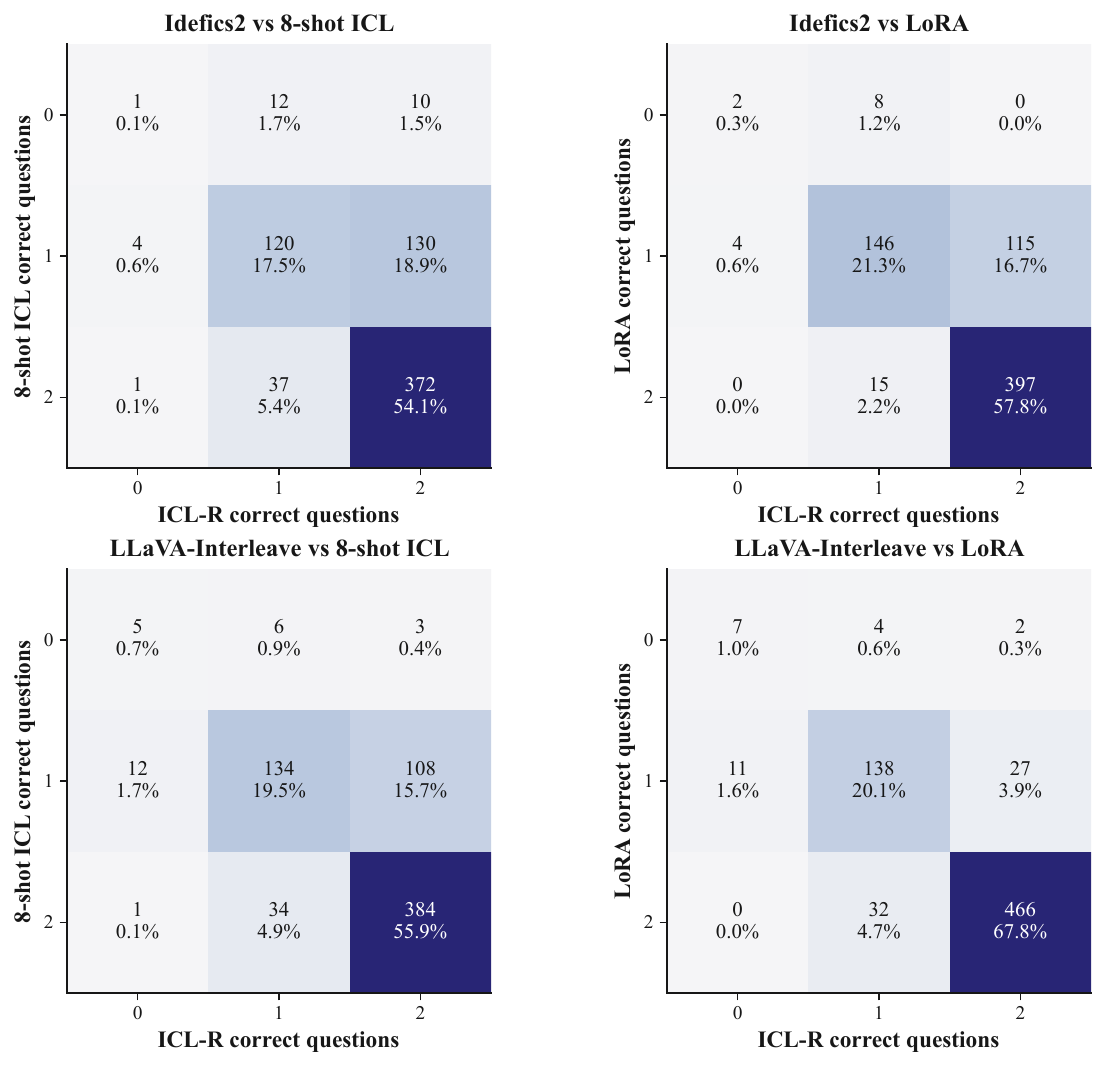}
\caption{Changes in the number of correctly answered questions per MME image.
Rows denote the comparator's correct-question count and columns denote ICL-R's
count. Cells show image counts and percentages of all 687 evaluation images.
Results are shown against explicit 8-shot ICL and LoRA on both backbones.}
\label{fig:exp5-mme-transitions}
\end{figure}

\subsection{Residual failures and qualitative limitations.}
We distinguish persistent errors from newly introduced errors on the complete
paired MME and SEED evaluation records. Task categories are taken from benchmark
annotations and are not interpreted as causal labels for internal model failure.
We additionally prepare a fixed stratified sample of 120 ICL-R errors, with
30 records for each backbone--dataset combination, and inspect ten diagnostic
examples selected by declared task and outcome strata. This qualitative review
includes text-associated color errors, a failed verification of the visible
words ``CASA GRECQUE'' and ``RESTAURANTS'', and incorrect spatial judgments,
alongside correct controls. Some examples also expose evaluation ambiguity:
two text-related controls concern stock-image watermarks, one spatial question
calls a visible fire extinguisher a fire hydrant, and a scene with multiple
cranes leaves the intended spatial referents unclear. We retain these records
in the quantitative evaluation and report their limitations rather than
attributing them to a particular calibration branch. The qualitative review
is not a double-annotated estimate of causal error-type prevalence.

\begin{figure}[H]
\centering
\includegraphics[width=\linewidth]{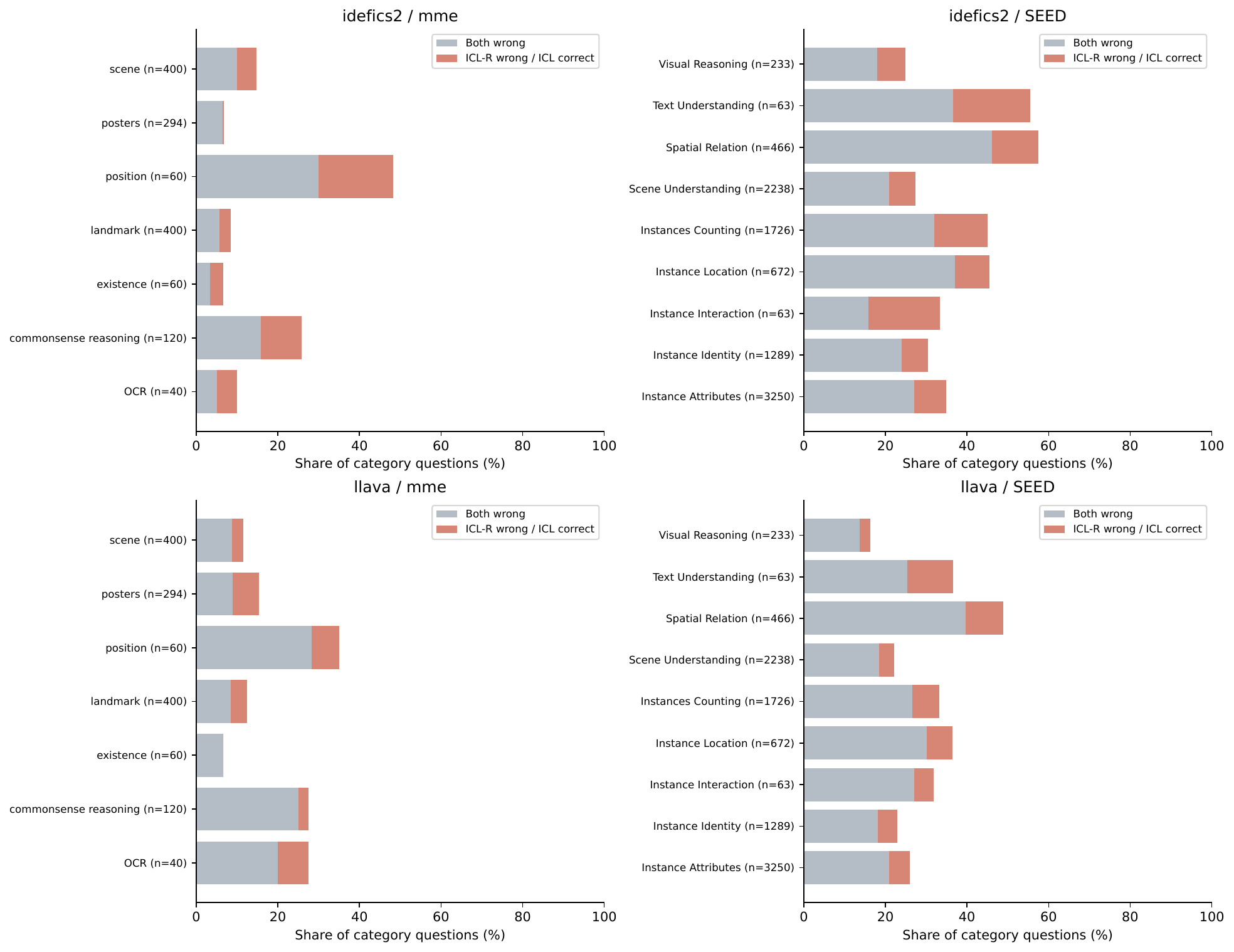}
\caption{Residual ICL-R errors by benchmark task category. Gray bars represent
questions answered incorrectly by both ICL-R and explicit 8-shot ICL; red bars
represent questions answered correctly by explicit ICL but incorrectly by
ICL-R. Percentages use all questions in each category, with category sample
sizes included. The categories describe task requirements rather than proven
internal failure mechanisms.}
\label{fig:exp1-residual-failures}
\end{figure}

\subsection{Qualitative Examples Analysis}\label{app:qualitative-examples}

We examine 56 curated examples to understand where ICL-R's gains arise and where its predictions remain unreliable. In the MME cases (Figure~\ref{fig:qual_mme}), ICL-R answers every question correctly, while at most two of the six baselines answer any individual question correctly. The examples span exact logo spelling, chart reading, object-related commonsense, and spatial relations, suggesting that the advantage is not confined to a single question type. The SEED-Bench cases in Figures~\ref{fig:qual_seed_unique_a} and~\ref{fig:qual_seed_unique_b} provide further examples where only ICL-R selects the ground-truth answer. Several require identifying what is absent or recognizing what cannot be inferred from the image: no cat is visible among the cushions, and a monochrome photograph does not reveal a guitar's actual color. Figure~\ref{fig:qual_seed_shared} provides an important complement: ICL-R also succeeds on cases solved by one or more baselines. The qualitative advantage therefore concerns which visual distinctions are preserved across cases.

The Flickr30k comparisons reveal differences that answer accuracy alone cannot express (Figures~\ref{fig:qual_flickr_vqav2} and~\ref{fig:qual_flickr_okvqa}). ICL-R identifies the yellow vehicle as a pedicab where several baselines call it a taxi, describes a conductor rather than a trumpet player, and retains details such as a tattooed arm or the number of medals. These are phrase-level comparisons: a baseline caption can contain both useful observations and an unsupported detail. The failure cases expose three remaining challenges. On VQAv2, ICL-R can bind a visible attribute to the wrong object or misread what the question requests, as when it gives the color of a suitcase for a purse or a time instead of the word printed on a clock. On OK-VQA, recognizing the pictured object does not always lead to the requested knowledge: the model answers ``dog'' when asked for its wild ancestor. On COCO, fluent captions can still introduce unsupported objects, equipment, or actions, including a dog absent from a vehicle and a tennis racket in place of a frisbee (Figure~\ref{fig:qual_coco_failures}). Together, these examples suggest that efficient query-only adaptation must preserve task-relevant visual evidence while also controlling how that evidence is used in an answer. The selected cases illustrate these behaviors; their prevalence and overall performance are assessed by the quantitative results.
\begingroup
\setlength{\abovecaptionskip}{2pt}
\setlength{\belowcaptionskip}{0pt}
\begin{figure}[H]
    \centering
    \includegraphics[width=0.98\linewidth]{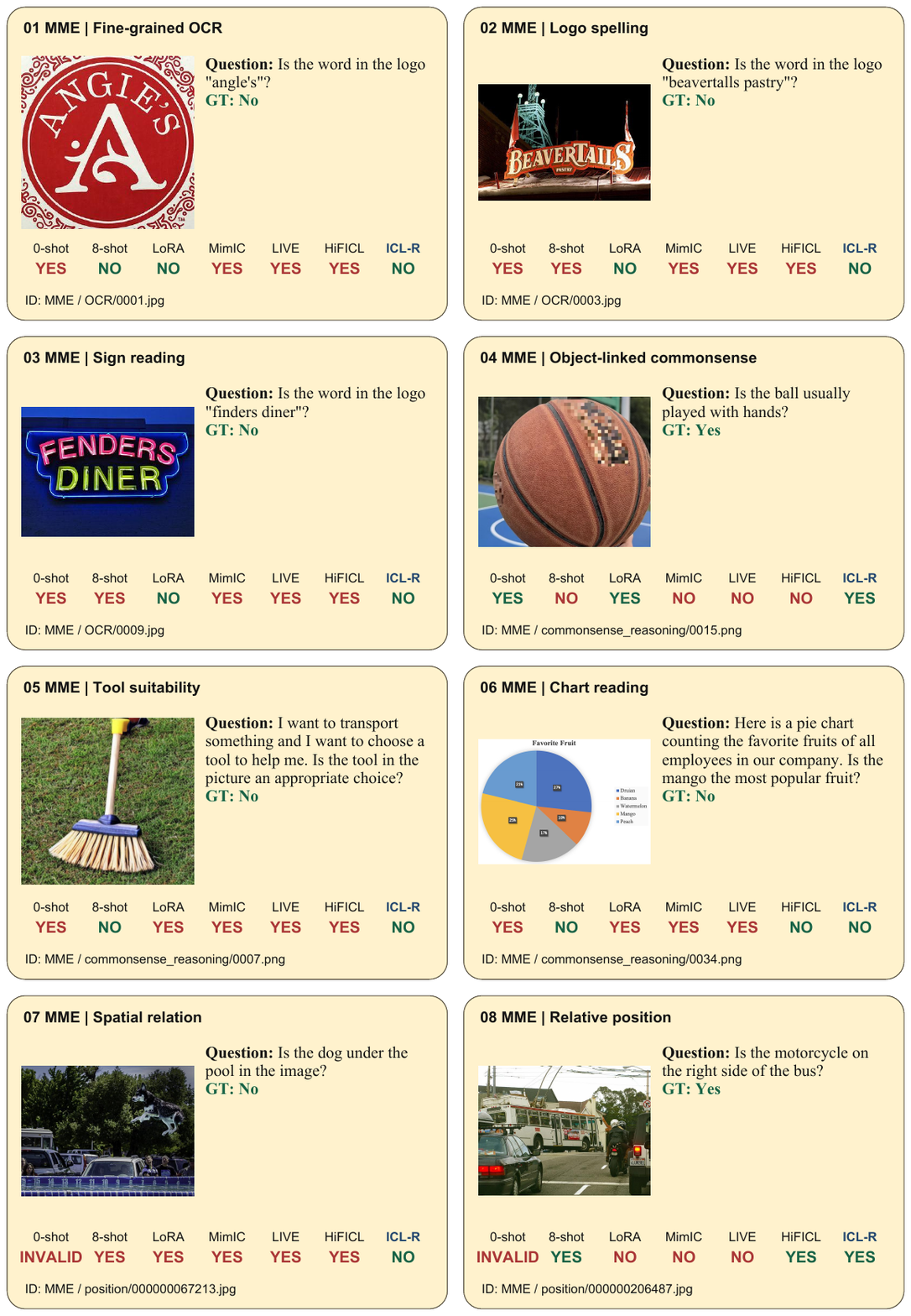}
    \vspace{-3pt}
    \caption{Selected MME examples using Idefics2. ICL-R answers all eight correctly, while no more than two of the six baselines answer any individual case correctly. The examples cover fine-grained OCR, commonsense, chart reading, and spatial relations. Green denotes correct answers; red denotes incorrect or invalid answers.}
    \label{fig:qual_mme}
\end{figure}
\endgroup

\begingroup
\setlength{\abovecaptionskip}{2pt}
\setlength{\belowcaptionskip}{0pt}
\begin{figure}[H]
    \centering
    \includegraphics[width=0.98\linewidth]{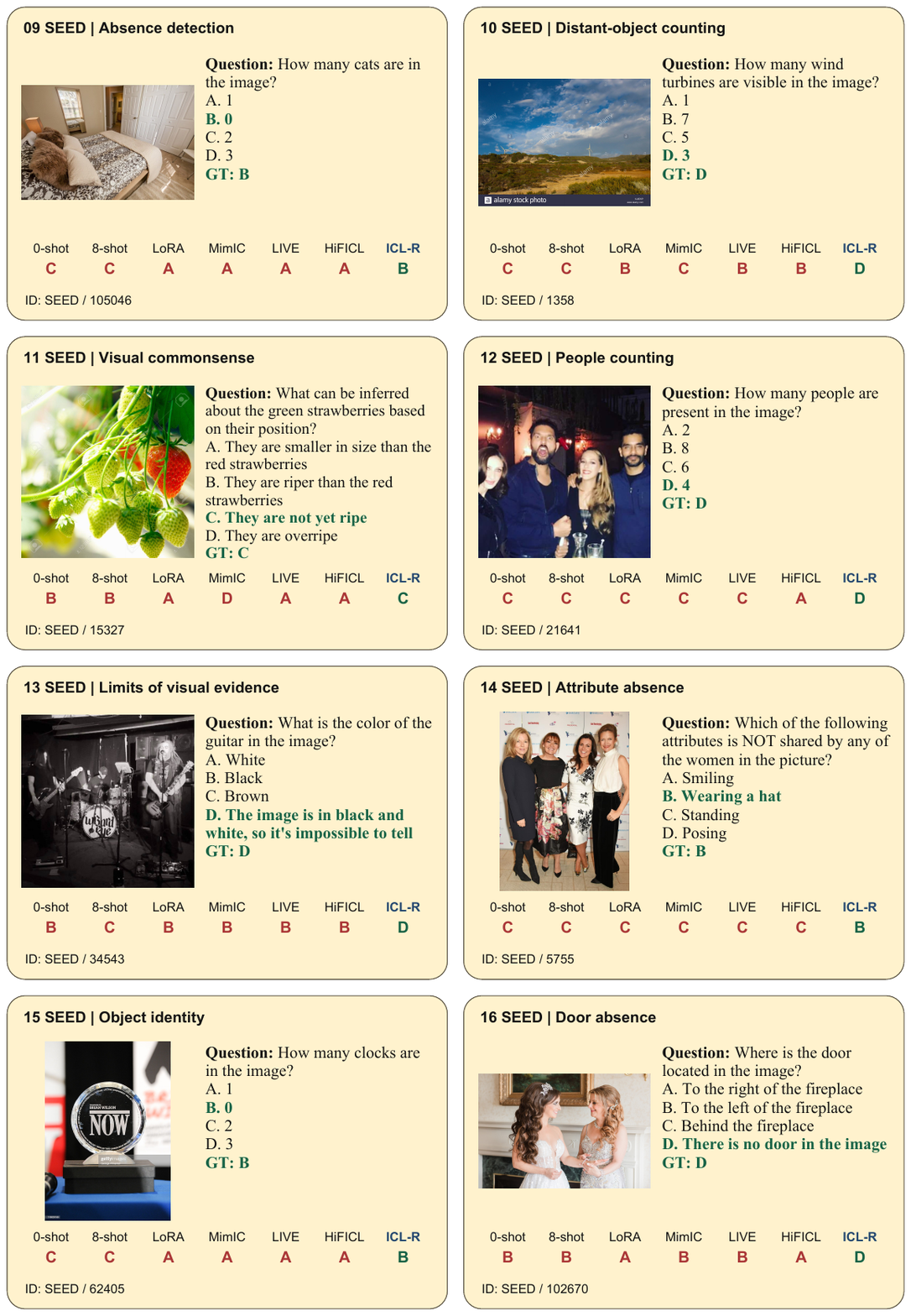}
    \vspace{-3pt}
    \caption{SEED-Bench examples answered correctly only by ICL-R among the seven Idefics2 methods. These cases involve object absence, counting, visual commonsense, and recognizing the limits of visual evidence. Green denotes correct answers and the ground-truth choice; red denotes incorrect baseline answers.}
    \label{fig:qual_seed_unique_a}
\end{figure}
\endgroup

\begingroup
\setlength{\abovecaptionskip}{2pt}
\setlength{\belowcaptionskip}{0pt}
\begin{figure}[H]
    \centering
    \includegraphics[width=0.98\linewidth]{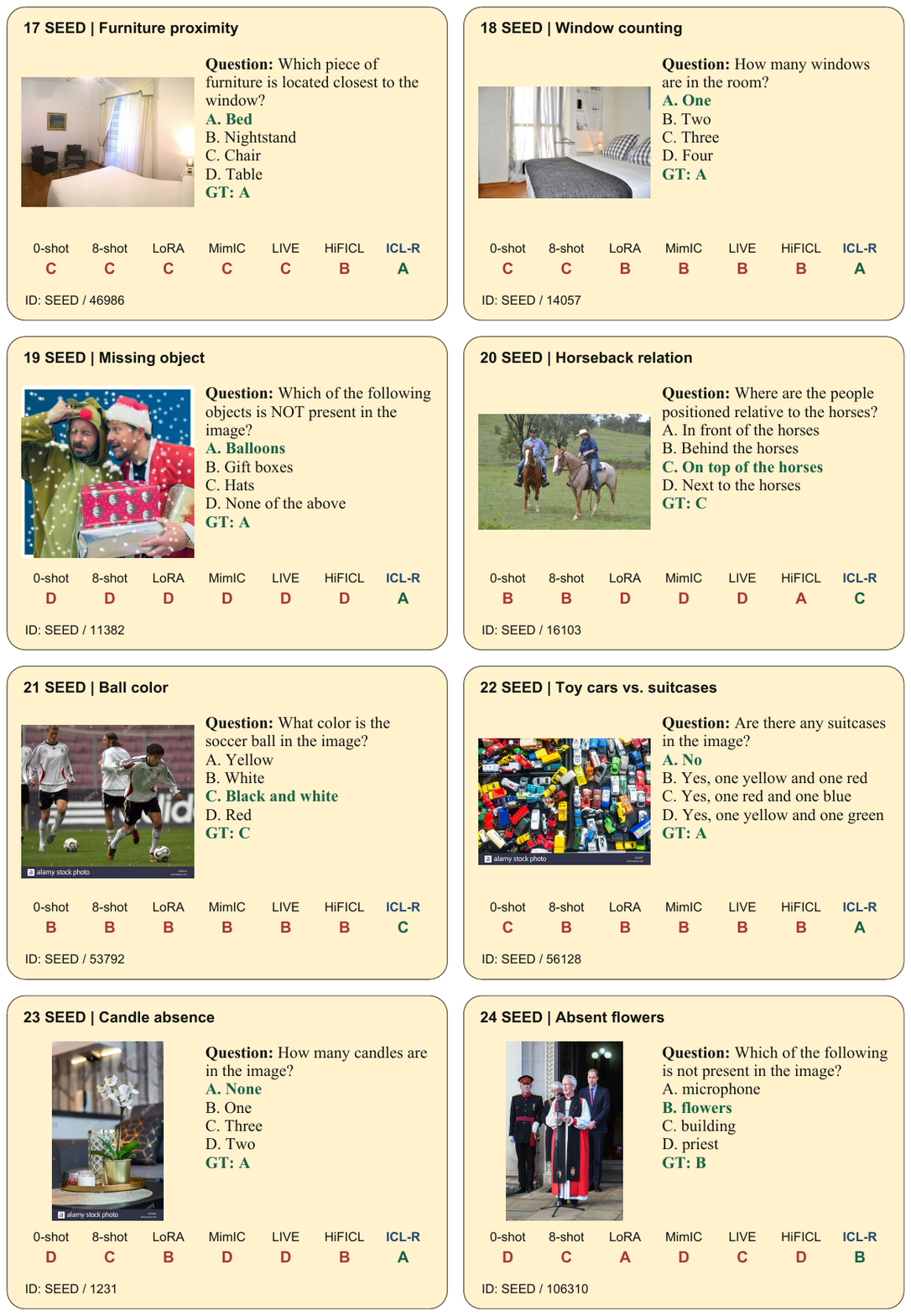}
    \vspace{-3pt}
    \caption{Additional SEED-Bench examples answered correctly only by ICL-R. The cases test object identity, spatial relations, counting, and discrimination between visually similar objects. Green denotes correct answers and the ground-truth choice; red denotes incorrect baseline answers.}
    \label{fig:qual_seed_unique_b}
\end{figure}
\endgroup

\begingroup
\setlength{\abovecaptionskip}{2pt}
\setlength{\belowcaptionskip}{0pt}
\begin{figure}[H]
    \centering
    \includegraphics[width=0.98\linewidth]{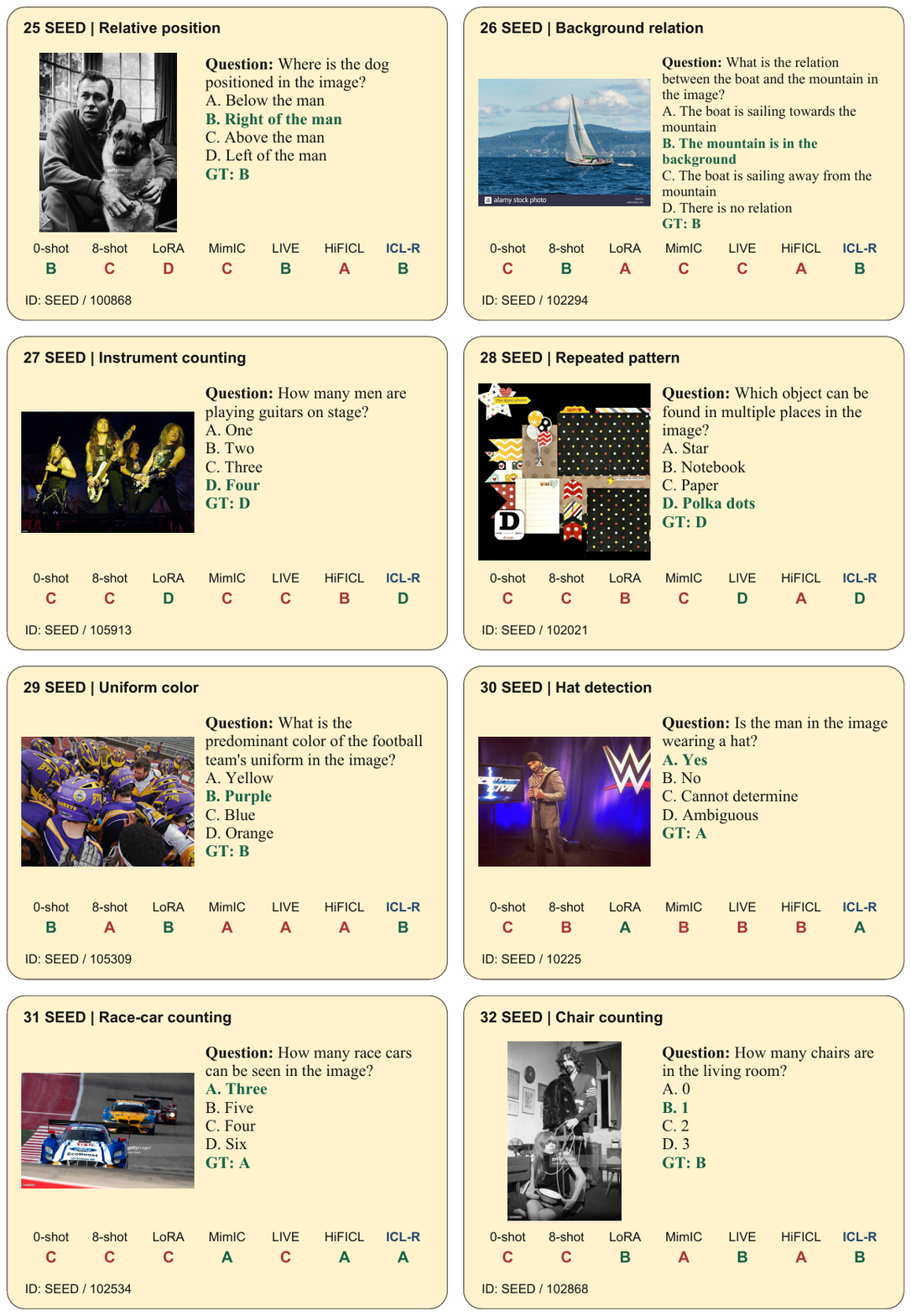}
    \vspace{-3pt}
    \caption{SEED-Bench examples where ICL-R and at least one baseline answer correctly. The comparisons cover spatial relations, repeated patterns, colors, and counting, showing which capabilities are shared and where predictions differ. Green denotes correct answers; red denotes incorrect answers.}
    \label{fig:qual_seed_shared}
\end{figure}
\endgroup

\begingroup
\setlength{\abovecaptionskip}{2pt}
\setlength{\belowcaptionskip}{0pt}
\begin{figure}[H]
    \centering
    \includegraphics[width=0.98\linewidth]{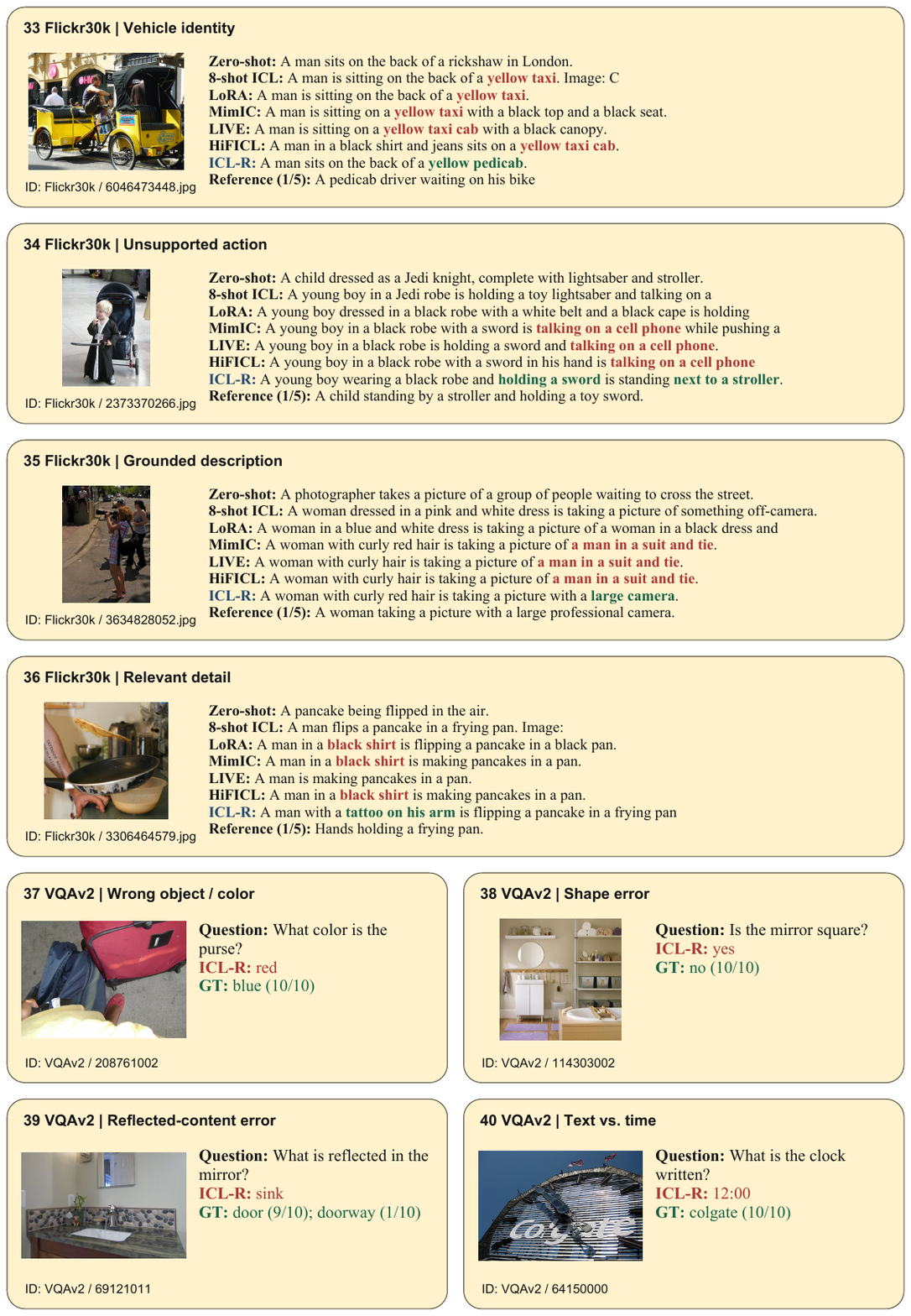}
    \vspace{-3pt}
    \caption{Flickr30k caption comparisons (top) and ICL-R failures on VQAv2 (bottom). The Flickr30k cases highlight grounded details in ICL-R's descriptions and unsupported details in some baseline captions. The VQAv2 cases show errors in object attributes, mirror interpretation, and reading the question. Green marks supported phrases or ground-truth answers; red marks incorrect or unsupported content.}
    \label{fig:qual_flickr_vqav2}
\end{figure}
\endgroup

\begingroup
\setlength{\abovecaptionskip}{2pt}
\setlength{\belowcaptionskip}{0pt}
\begin{figure}[H]
    \centering
    \includegraphics[width=0.98\linewidth]{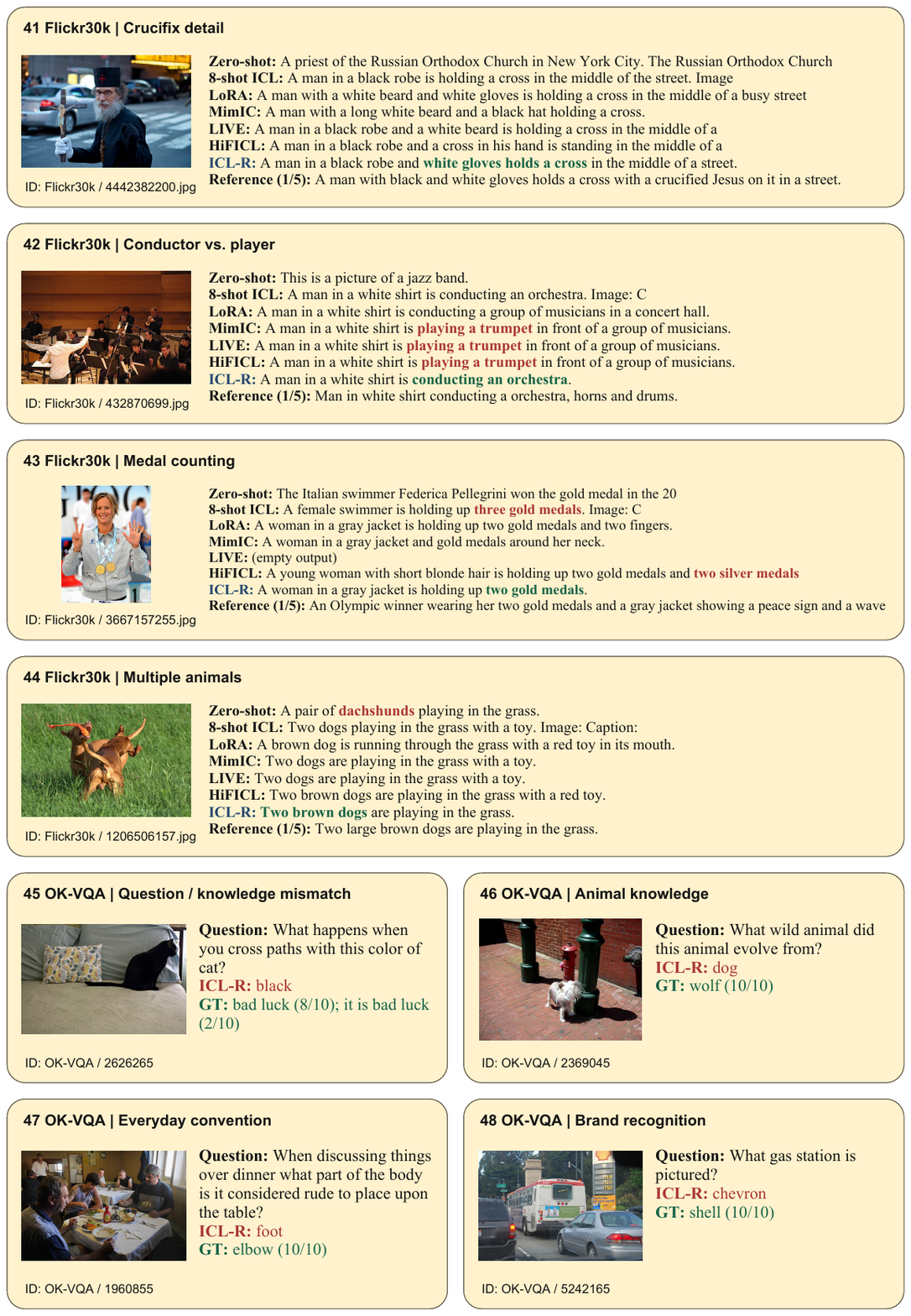}
    \vspace{-3pt}
    \caption{Further Flickr30k caption comparisons (top) and ICL-R failures on OK-VQA (bottom). ICL-R captures details such as a conductor's role and the number of medals, but the OK-VQA cases expose difficulties with external knowledge, everyday conventions, and brand recognition. Green marks supported phrases or ground-truth answers; red marks incorrect or unsupported content.}
    \label{fig:qual_flickr_okvqa}
\end{figure}
\endgroup

\begingroup
\setlength{\abovecaptionskip}{2pt}
\setlength{\belowcaptionskip}{0pt}
\begin{figure}[H]
    \centering
    \includegraphics[width=0.98\linewidth]{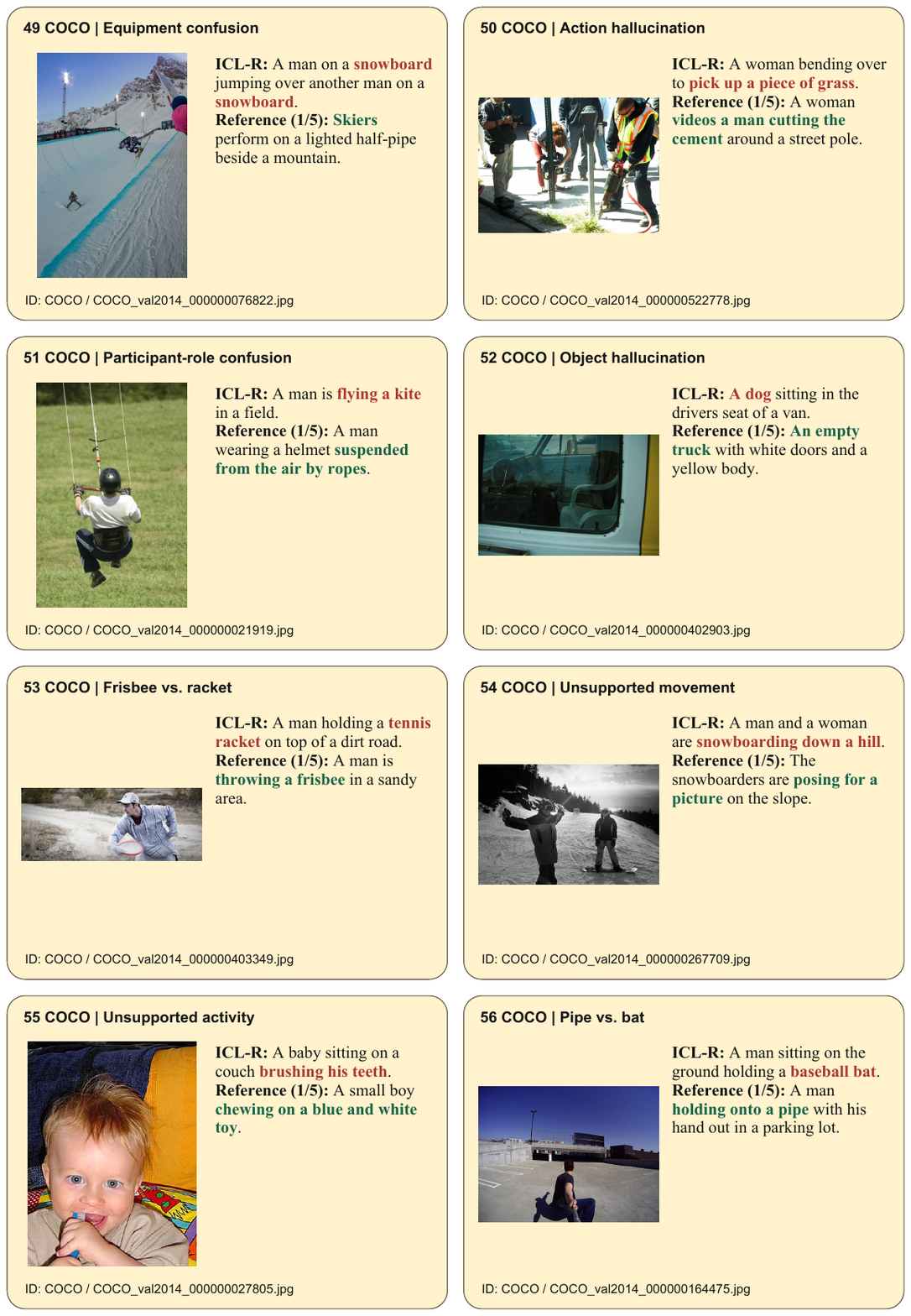}
    \vspace{-3pt}
    \caption{ICL-R failure cases on COCO captioning. The examples include object and equipment confusion, unsupported actions, and hallucinated objects. Red highlights the specific problematic phrase in the generated caption, while green highlights relevant evidence in one of the five reference captions shown for each image.}
    \label{fig:qual_coco_failures}
\end{figure}
\endgroup

\end{document}